%% file: main.tex
\documentclass[runningheads]{llncs}
\PassOptionsToPackage{table,dvipsnames}{xcolor}

\usepackage{eccv}

\usepackage{eccvabbrv}
\usepackage{hyphenat}
\usepackage{xurl}

\usepackage{graphicx}
\usepackage{booktabs}

\usepackage[accsupp]{axessibility}  % Improves PDF readability for those with disabilities.

\usepackage{multirow}
\usepackage{rotating}
\usepackage{caption}
\newcommand{\cmark}{\checkmark}
\usepackage{pifont}
\usepackage{makecell}
\usepackage{wrapfig}

\definecolor{topOne}{RGB}{161,217,155}
\definecolor{topTwo}{RGB}{204,235,197}
\definecolor{topThr}{RGB}{237,248,233}
\newcommand{\topA}[1]{\cellcolor{topOne}#1}
\newcommand{\topB}[1]{\cellcolor{topTwo}#1}
\newcommand{\topC}[1]{\cellcolor{topThr}#1}

\usepackage[hidelinks]{hyperref}
\hypersetup{
  pdftitle={Weather-Conditioned Depth Anything},
  pdfauthor={Zhaoming Xu, Chan-Wei Hu, Kuan-Ru Huang, Zihao Zhu, Renjie Li, Yang Zhou, Zhengzhong Tu}
}

\usepackage{orcidlink}

\usepackage{paralist}
\begin{document}

\title{Weather-Conditioned Depth Anything} 

\titlerunning{Weather-Conditioned Depth Anything}

\author{
Zhaoming Xu\inst{1}\orcidlink{0009-0008-1033-5978}
\and
Chan-Wei Hu\inst{1}\orcidlink{0009-0005-9066-5229}
\and
Kuan-Ru Huang\inst{1}\orcidlink{0009-0002-2398-6012}
\and
Zihao Zhu\inst{1}\orcidlink{0009-0001-8585-7518}
\and
Renjie Li\inst{1}
\and
Yang Zhou\inst{1}\orcidlink{0000-0001-5366-5389}
\and
Zhengzhong Tu\inst{1}\orcidlink{0000-0002-7594-2292}\thanks{Corresponding Author}
}

\authorrunning{Z. Xu et al.}
% \authorrunning{Zhaoming Xu\inst{1}\orcidlink{0009-0008-1033-5978}}

\institute{
Texas A\&M University, College Station, TX 77843, USA\\
\email{\{zhaoming0812,tzz\}@tamu.edu}
}
% \institute{Texas A\&M University, College Station TX 77843, USA
% \email {zhaoming0812, } \\
% \and
% Springer Hj jeidelberg, Tiergartenstr.~17, 69121 Heidelberg, Germany
% \email{\{zhaoming0812,tzz\}@tamu.edu}

\maketitle
% \begin{figure}[t]
    % \centering
    
\begin{center}
    \setlength{\tabcolsep}{1pt}
    \renewcommand{\arraystretch}{1.0}
    
    \resizebox{\textwidth}{!}{
        \begin{tabular}{ccccc}
             & Snow & Fog & Low-light & Rain \\[2pt]
            
            % \multirow{2}{*}{\rotatebox[origin=c]{90}{\textbf{Input}}}
            \multirow{1}{*}[11mm]{\rotatebox[origin=c]{90}{\scriptsize\textbf{Input}}}
            & \includegraphics[width=.23\linewidth,keepaspectratio]{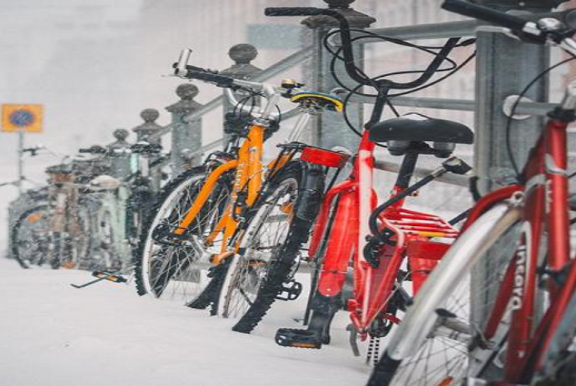}
            & \includegraphics[width=.23\linewidth,keepaspectratio]{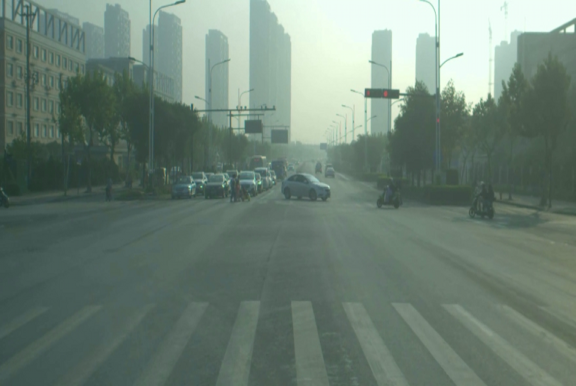}
            & \includegraphics[width=.23\linewidth,keepaspectratio]{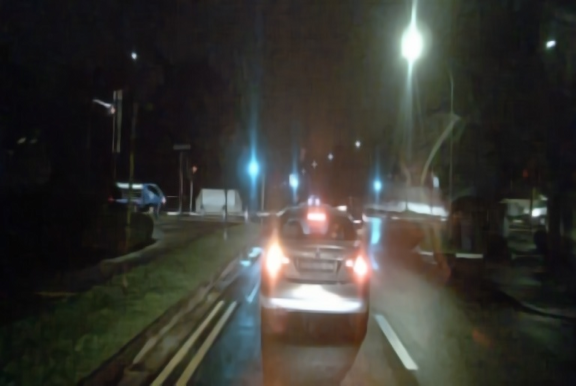}
            & \includegraphics[width=.23\linewidth,keepaspectratio]{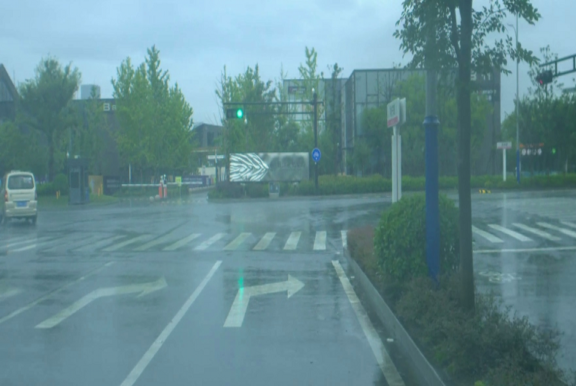} \\
            
            % \multirow{2}{*}{\rotatebox[origin=c]{90}{\textbf{DA v2}}}
            \multirow{1}{*}[11mm]{\rotatebox[origin=c]{90}{\scriptsize\textbf{DA v2}}}
            & \includegraphics[width=.23\linewidth,keepaspectratio]{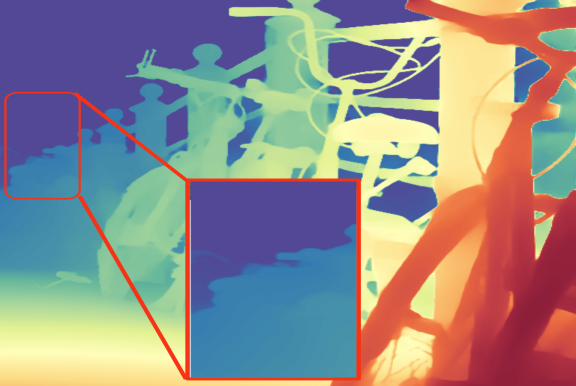}
            & \includegraphics[width=.23\linewidth,keepaspectratio]{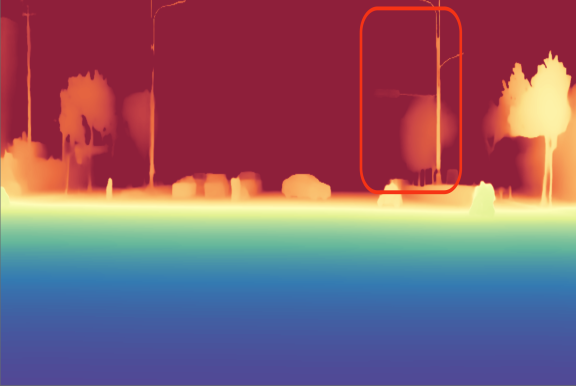}
            & \includegraphics[width=.23\linewidth,keepaspectratio]{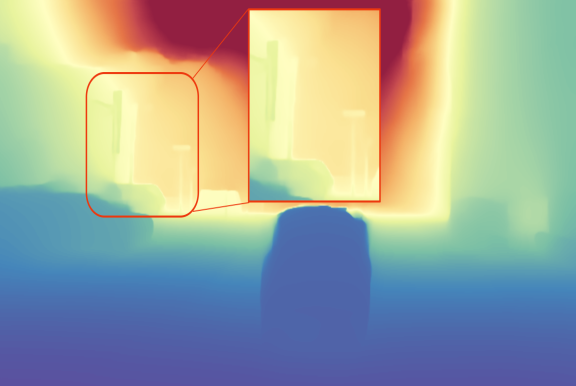}
            & \includegraphics[width=.23\linewidth,keepaspectratio]{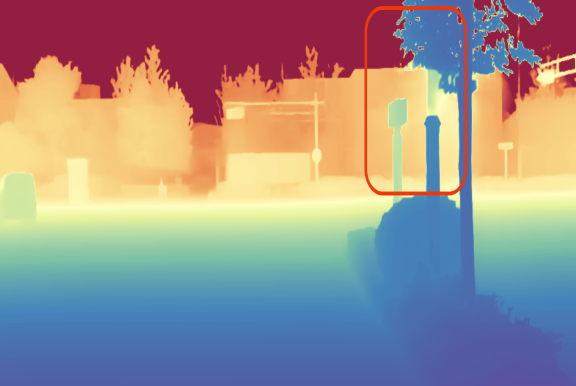} \\
            
            % \multirow{2}{*}{\rotatebox[origin=c]{90}{\textbf{DA-AC}}}
            \multirow{1}{*}[12mm]{\rotatebox[origin=c]{90}{\scriptsize\textbf{DA-AC}}}
            & \includegraphics[width=.23\linewidth,keepaspectratio]{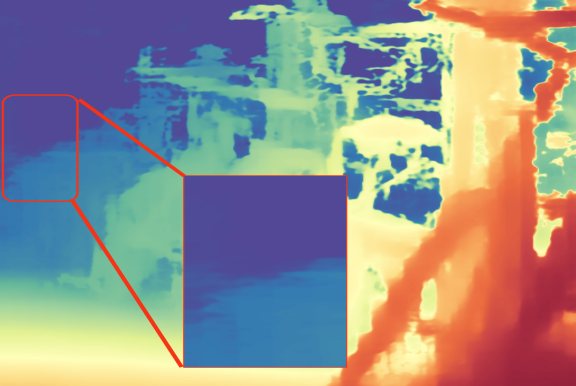}
            & \includegraphics[width=.23\linewidth,keepaspectratio]{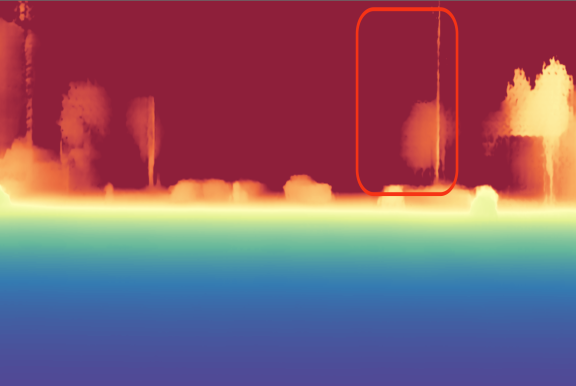}
            & \includegraphics[width=.23\linewidth,keepaspectratio]{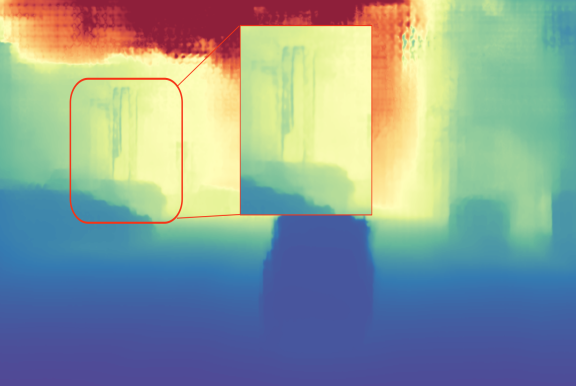}
            & \includegraphics[width=.23\linewidth,keepaspectratio]{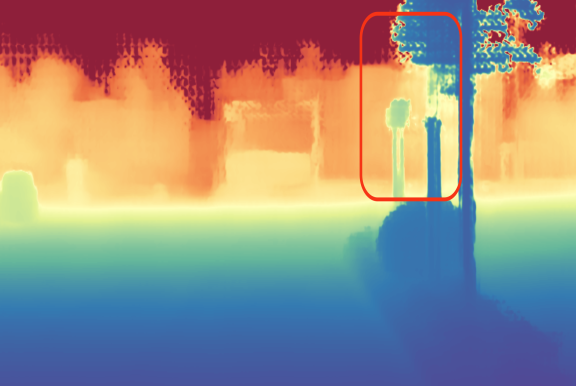} \\
            
            % \multirow{2}{*}{\rotatebox[origin=c]{90}{\textbf{DA-W (Ours)}}}
            \multirow{1}{*}[15.5mm]{\rotatebox[origin=c]{90}{\scriptsize\textbf{DA-W(Ours)}}}
            & \includegraphics[width=.23\linewidth,keepaspectratio]{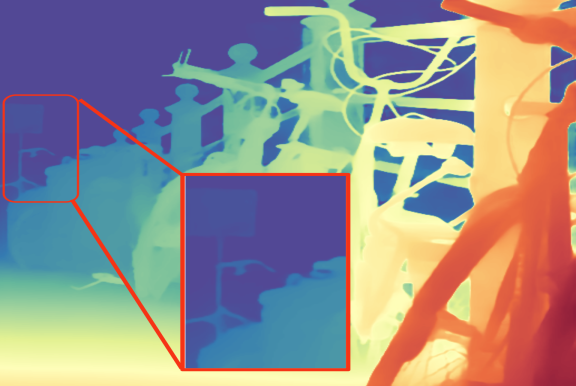}
            & \includegraphics[width=.23\linewidth,keepaspectratio]{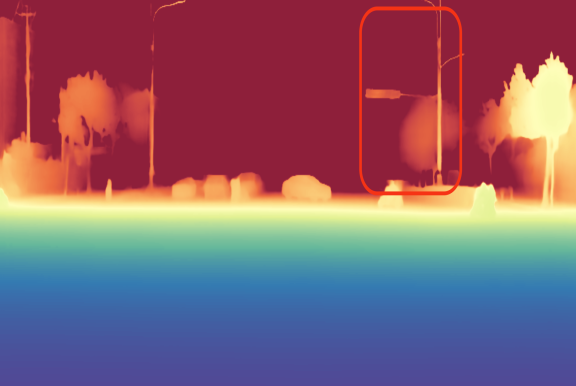}
            & \includegraphics[width=.23\linewidth,keepaspectratio]{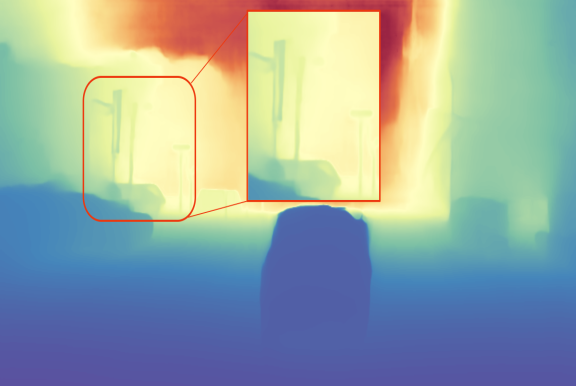}
            & \includegraphics[width=.23\linewidth,keepaspectratio]{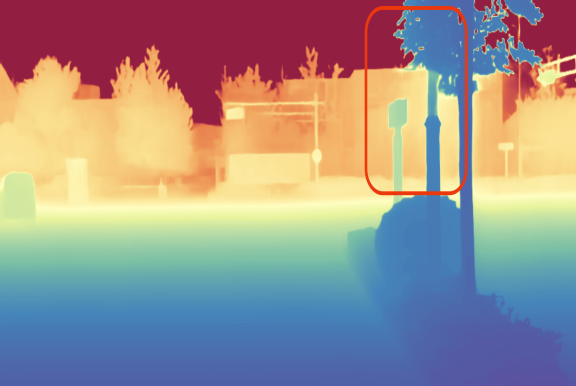}
        \end{tabular}
    }
    \vspace{-2mm}
    \captionof{figure}{\textbf{Depth estimation results under various real-weather degraded conditions.}
    Our DA-W generates accurate details and well-preserved structure under snow, fog, rain, and low-light conditions. Compared with Depth Anything v2~\cite{yang2024depth} and Depth Anything-AC~\cite{sun2025depth}, our method produces finer-grained and more robust predictions.}
    \label{fig:teaser}
\end{center}
    % \vspace{-4mm}
% \end{figure}

\begin{abstract}
Monocular depth estimation foundation models, such as the Depth Anything series, have achieved remarkable performance across diverse domains. However, they still suffer from critical failures under adverse weather conditions, such as fog, rain, snow, or at night.
To address this, we present Weather-Conditioned Depth Anything (DA-W), a framework that explicitly disentangles style from content for weather-robust depth estimation. 
Specifically, we introduce a \textit{Style Filter} trained on a curated mix of real and synthetic degradation datasets to extract content-independent, degradation-aware weather embeddings. 
This style embedding is then injected into the Depth Anything backbone using a parameter-efficient, zero-initialized adapter. 
Such a lightweight modulation allows a single unified model to robustly adapt to diverse conditions—including fog, rain, snow, and low-light—while avoiding catastrophic forgetting of its core generalization abilities in normal conditions. We train the adapter using a pseudo-label distillation and alignment strategy.
Our comprehensive experiments demonstrate that our proposed DA-W achieves state-of-the-art robust depth estimation, improving AbsRel by an average of 3.7\% on our curated weather benchmarks, while matching or slightly outperforming performance on standard clean benchmarks. Our project page is available at: \url{https://zhaoming-tamu.github.io/WCDA/}.
\end{abstract}

\section{Introduction}
\label{sec:intro}

Monocular depth estimation (MDE) has become a fundamental component in 3D perception~\cite{xue2020toward, ganj2024mobile, liu2025efficient} for cost-effective applications, such as autonomous driving, robotics, and AR/VR, where it can replace relying on LiDAR or stereo cameras. 
Driven by large-scale datasets and model training, recent depth foundation models~\cite{bochkovskii2024depth, wang2025moge} have demonstrated strong generalization in depth estimation across diverse environments.
However, their performance degrades dramatically under adverse weather conditions—fog, snow, rain, and nighttime—due to atmospheric scattering, occlusion, specular artifacts, and low-light noise, which fundamentally alter the appearance (See Figure. \ref{fig:teaser}).

\begin{wrapfigure}{R}{0.42\textwidth}
  \vspace{-6mm}
  \centering
  \includegraphics[width=\linewidth]{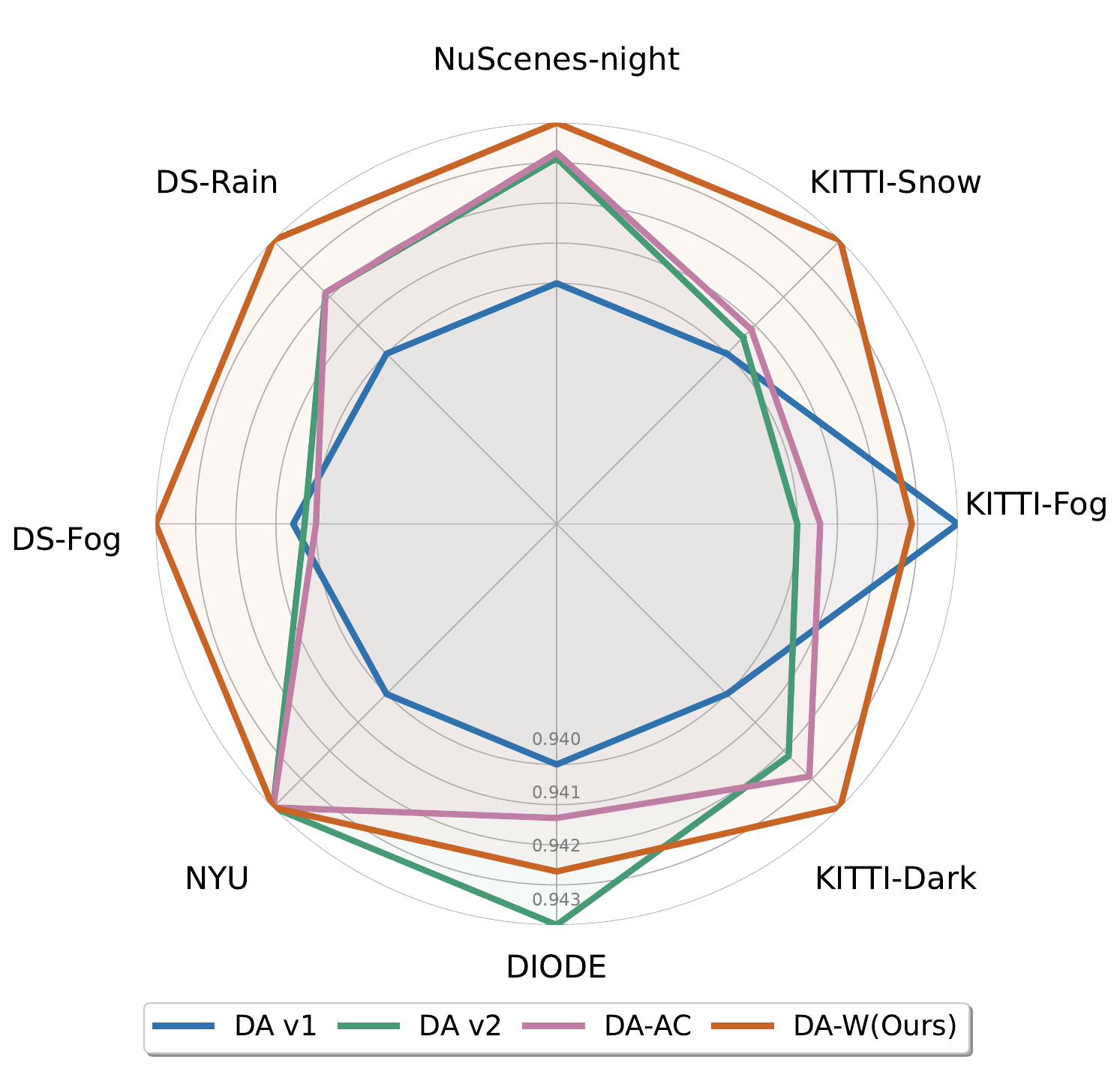}
  \vspace{-4mm}
  \caption{\textbf{Performance $\delta_1\uparrow$ comparison across three datasets}: real adverse weather, synthetic adverse weather, and clean weather. Values are normalized per dataset (Baseline = 0.6).}
  \label{fig:radar}
  \vspace{-6mm}
\end{wrapfigure}

This robustness gap persists even for state-of-the-art foundation models such as the Depth Anything ~\cite{depthanything, yang2024depth, depthanything3}. Broadly, there are two strategies to improve robustness under adverse weather conditions: \textbf{(1)} pre-process the image to remove weather effects before depth estimation, and \textbf{(2)} adapt the depth model to the weather conditions.

The first strategy has an inherent limitation: restoration methods are typically optimized for perceptual quality rather than geometric accuracy, and often distort the very cues critical for reliable depth estimation~\cite{danier2025depthcues, Shyam_2024_WACV, chen2024robustsam}.% (see Table~\ref{tab:real_low_quality}).

Some studies train the restoration for task-oriented objectives~\cite{Feijoo_2025_CVPR, li2024light}, yet they often overfit to a specific downstream model, making them non-generalizable to other estimators.
The second strategy is more principled but requires large-scale annotated depth data under real adverse conditions, which is prohibitively expensive to collect. As a practical workaround, existing methods predominantly rely on \textbf{synthetic} weather degradations applied to clean images~\cite{tosi2024diffusion, karvat2024adver, saunders2023self}.

\begin{figure}[t]
    \centering
    \begin{minipage}{0.49\linewidth}
        \centering
        \includegraphics[width=\linewidth]{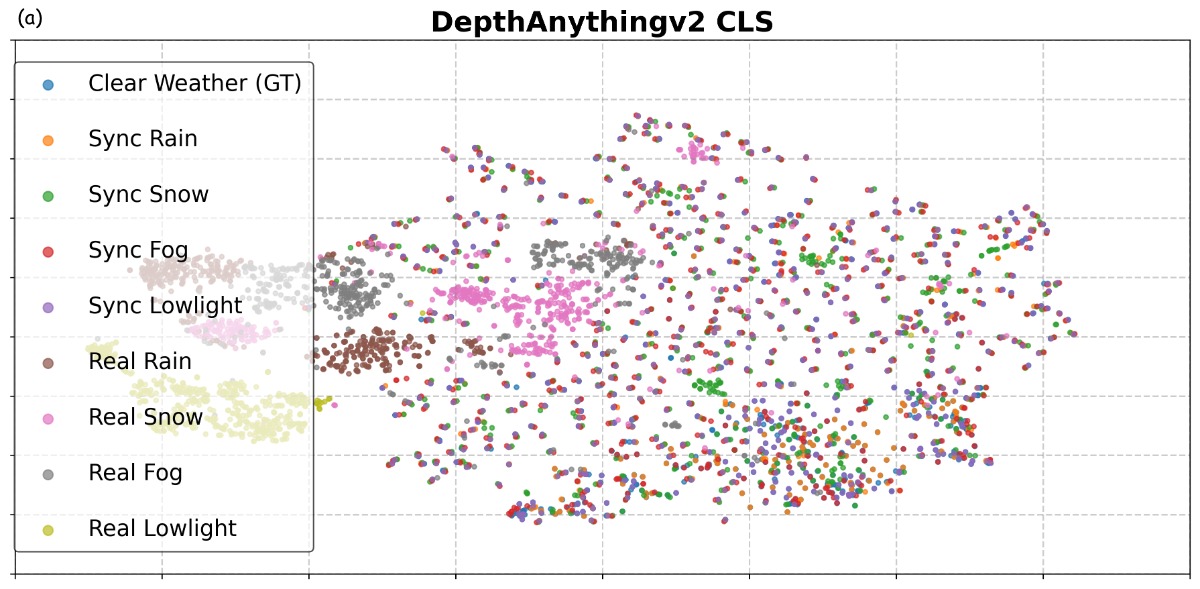}
        % \vspace{0.1em}
        \small (a) Depth Anything v2 CLS
    \end{minipage}\hfill
    \begin{minipage}{0.49\linewidth}
        \centering
        \includegraphics[width=\linewidth]{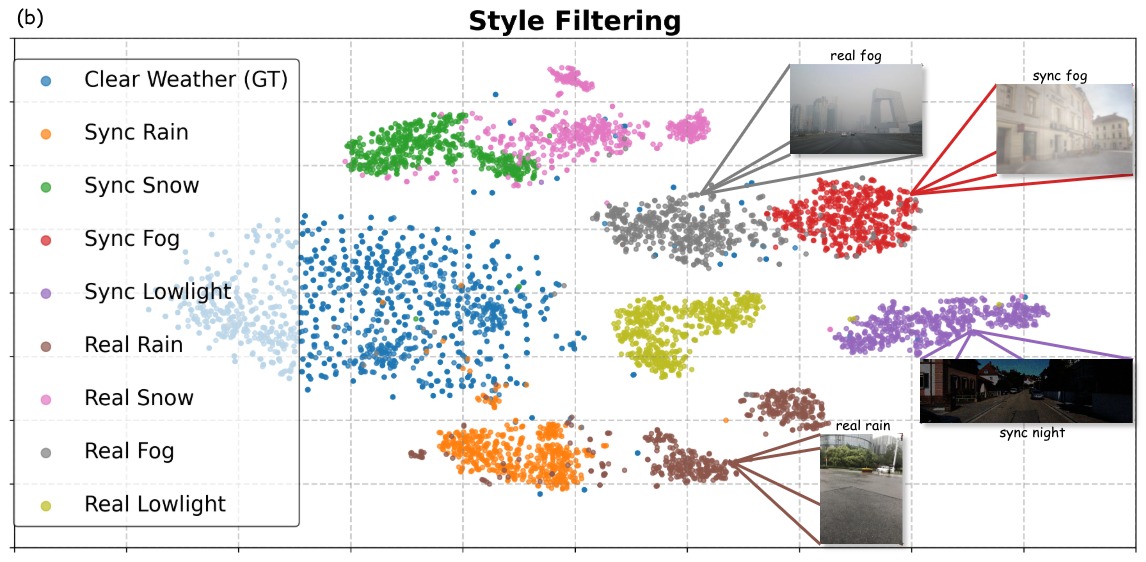}
        % \vspace{0.1em}
        \small (b) Style Filter
    \end{minipage}

    \caption{\textbf{Feature space comparison for style characterization.}
    Depth Anything v2~\cite{yang2024depth} \textit{CLS} tokens fail to separate weather conditions, while our \textit{style filter} features form distinct clusters.}
    \label{fig:tsne}
\end{figure}

Consequently, the key question is whether features learned from synthetic data generalize reliably to real-world scenarios.
To probe this generalization gap, we analyze how a foundation depth model represents appearance variations across domains.
We extract the \texttt{[CLS]} tokens in vision transformer (ViT) embeddings~\cite{tumanyan2022splicing} as the global appearance vector on the Depth Anything v2 encoder~\cite{yang2024depth} across diverse real and synthetic weather conditions, and visualize the feature embeddings via t\textendash SNE~\cite{JMLR:v9:vandermaaten08a} in Figure.~\ref{fig:tsne} (left).

% {\setlength{\textfloatsep}{2pt plus 1pt minus 1pt}%

As may be seen, the features are fundamentally entangled. Representations from different weather types (e.g., fog, rain, snow) are heavily mixed, rather than forming separable clusters.
The model conflates weather-specific style with scene geometry into a single, undifferentiated representation, instead of providing a pure weather code. This entanglement is the root cause of the model's brittleness, as it cannot distinguish appearance variations caused by weather from those of the scene itself, preventing the learning of generalizable weather patterns and fatally hindering synthetic-to-real adaptation.

Motivated by these insights, we propose \textbf{W}eather\hyp{}\textbf{C}onditioned \textbf{D}epth \textbf{A}ny-thing (\textbf{DepthAnything\hspace{0pt}-W} or \textbf{DA-W}), a simple yet robust adaptation strategy to robustify depth estimation under adverse weather conditions.
We first design a \textit{Style Filter} network using contrastive learning on a curated collection comprising 7 real-world adverse-weather datasets and 4 clean datasets, from which corresponding synthetic weather views are generated.
This module learns to align the style distributions of real and synthetic degradations, as shown in Figure~\ref{fig:tsne}(right), thereby producing a discriminative weather embedding. 
The embedding then conditions the Depth Anything foundation model using lightweight modulation techniques \textit{AdaLN-Zero}. 
This lightweight conditioning allows the model to adjust its predictions based on the estimated weather while keeping the encoder frozen, preventing catastrophic forgetting and preserving clean-scene performance.
Extensive experiments demonstrate that DA-W achieves state-of-the-art performance across both general and adverse-weather benchmarks Figure~\ref{fig:radar}.

Our contributions are summarized as follows: 
\begin{compactitem}
\item We propose DepthAnything-W, a novel weather-robust depth estimation foundation model that learns to adapt to diverse real weather conditions.
    \item We introduce a \textit{Style Filter} trained via contrastive loss on mixed datasets of real and synthetic data, learning domain-aligned weather embeddings that bridge the gap between synthetic and real data.
    \item We propose a parameter-efficient adaptation strategy that conditions the Depth Anything foundation model using the learned weather embedding via lightweight \textit{AdaLN-Zero} modulation.
    \item Our comprehensive experiments show that DA-W achieves state-of-the-art performance on diverse adverse-weather benchmarks without degrading generalization performance on clean data.
\end{compactitem}

\section{Related Work}

\subsection{Monocular Depth Estimation (MDE).}
Monocular depth estimation (MDE) predicts scene geometry from a single RGB image and serves as a foundational component for various 3D perception tasks in robotics~\cite{simon2023mononav, job2024radar}, augmented and virtual reality~\cite{bochkovskii2024depth}, and autonomous driving.
The Initial deep learning approaches were either supervised using datasets such as KITTI~\cite{geiger2013vision} and NYUv2~\cite{silberman2012indoor}, or self-supervised through stereo~\cite{godard2019digging, zhou2022learning} or video photometric consistency~\cite{watson2021temporal, zhou2017unsupervised, yin2018geonet}. However, both strategies exhibited domain bias and limited generalization capabilities.

Recent research has addressed domain bias in early supervised and self-supervised pipelines by using more diverse datasets and larger models. These strategies help improve cross-domain robustness and facilitate foundational depth estimation.
MiDaS~\cite{ranftl2020towards} was one of the early methods to improve generalization by training on multiple datasets. Subsequent research advances these efforts by pretraining on larger and more diverse labeled and unlabeled datasets, as demonstrated by the Depth Anything series ~\cite{depthanything, yang2024depth, depthanything3}. Additional methods, including ZoeDepth~\cite{bhat2023zoedepth}, UniDepth~\cite{piccinelli2024unidepth}, and Metric3D(v2)~\cite{hu2024metric3d}, target generalization across various camera systems and domains.
These strategies demonstrate strong performance on previously unseen tasks.

In addition to scaling discriminative models, another research direction investigates generative depth estimation, which leverages strong priors from large generative models.
These methods formulate depth prediction as sampling from a conditional distribution. The framework inherently supports uncertainty-aware inference and enables tasks such as depth inpainting.
For example, Marigold~\cite{ke2024repurposing} employs diffusion-based image generators for monocular depth estimation. Flow matching models like DepthFM~\cite{gui2024depthfm} formulates monocular depth estimation as distribution transport and enable rapid, high-quality sampling.

\subsection{Robust Depth Estimation.}

Although large-scale training improves average cross-domain performance, depth models remain brittle when input images violate standard assumptions of image formation, such as rain streaks, fog scattering, or nighttime noise. These conditions corrupt or remove geometric cues~\cite{gasperini2023robust, saunders2023self}.
Among these factors, adverse weather is particularly challenging because of scattering, occlusions, and low illumination~\cite{gasperini2023robust}.

To address this, prior work largely falls into two categories:
(1) \textit{Two-stage pipelines}, which restore degraded images using diffusion or unified restoration models like UniRestore\cite{chen2025unirestore} and MWFormer~\cite{10767188} before depth estimation, but often alter geometric cues and degrade accuracy~\cite{saunders2023self}; and 
(2) \textit{Weather-aware training}, which incorporates robustness directly into the depth estimation model.
WeatherDepth~\cite{wang2024weatherdepth} employs curriculum contrastive learning from clear to severe weather, and md4all~\cite{gasperini2023robust} enforces consistency between clean and degraded samples.
Foundation model extensions like  DepthAnything-AC~\cite{sun2025depth} and ACDepth~\cite{jiang2025always}, incorporate physics-based degradations and knowledge distillation to achieve strong zero-shot robustness.
However, although these methods enhance resilience, they often entangle weather appearance with scene geometry.
Our approach explicitly disentangles them using a \textit{Style Filter} that learns domain-invariant weather embeddings, thereby enabling robust depth estimation without sacrificing clean-image generalization.

\section{Proposed Method}

\subsection{Preliminary: Depth Anything Model}
\begin{figure*}[t]
\centering
\includegraphics[width=\textwidth,keepaspectratio]{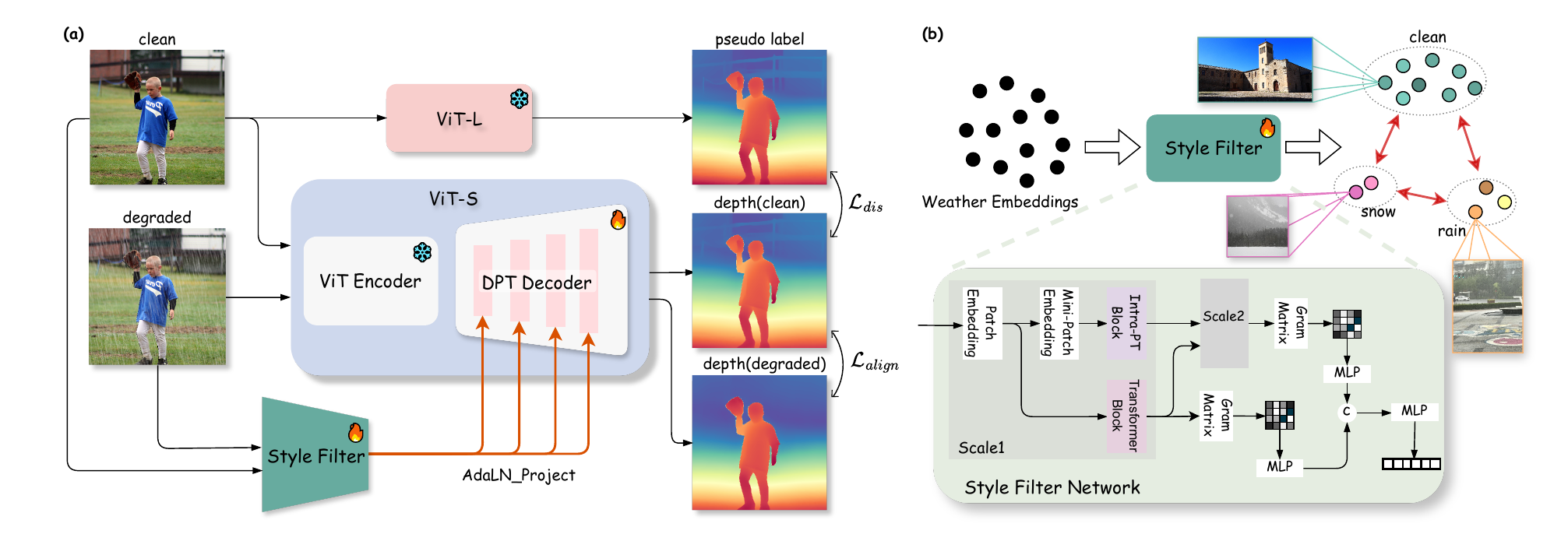}
\caption{\textbf{Overview of our proposed DepthAnything-W.} (a) The training pipeline: clean and degraded images pass through the Depth Anything backbone, with weather embeddings from the \textit{Style Filter} injected via \textit{AdaLN-Zero} at multiple decoder scales to produce weather-conditioned depth maps supervised by $\mathcal{L}_{dis}$ and $\mathcal{L}_{align}$. (b) The \textit{Style Filter} extracts weather embeddings from images via multi-scale Gram matrices, using contrastive learning to cluster by weather type while maintaining scene invariance and real-synthetic alignment. }
\label{fig:overview}
\end{figure*}

We provide a concise overview of the Depth Anything~\cite{yang2024depth} framework. As shown in Figure \ref{fig:overview}, it comprises two main components: the encoder employs a vision transformer (DINOv2) to extract hierarchical features, augmented with a multi-scale readout mechanism that produces pyramid feature representations, and the decoder adopts the DPT design~\cite{ranftl2021vision}, progressively fusing multi-scale features through refinement blocks with skip connections, followed by a lightweight prediction head for dense depth regression.  The decoder performs progressive upsampling with skip-wise fusion to recover fine structure. Depth Anything is trained on large mixed-supervision data with an affine-invariant objective, yielding zero-shot generalization across diverse scenes. In our method, we keep this backbone intact and inject weather conditioning only in the decoder to preserve clean-scene behavior while improving robustness.

\subsection{Weather Adaptation}

\subsubsection{Weather Embedding Formulation}
Given an RGB image $\mathit{I} \in \mathbb{R}^{H \times W \times 3}$ captured under a weather condition 
$c \in \mathcal{C}$ (e.g., fog, rain, snow, or low-light), our goal is to learn a \textit{Style Filter} 
$F_{\theta}(\cdot)$ that maps $\mathit{I}$ to a compact weather embedding $\mathbf{w} \in \mathbb{R}^{d}$:
\begin{equation}
    \mathbf{w} = F_{\theta}(\mathit{I}).
\end{equation}

The embedding $\mathbf{w}$ is designed to be both \textbf{weather-discriminative} and \textbf{domain-aligned}.  
It should encode the type of weather while remaining invariant to scene geometry and semantics, such that two distinct scenes under the same weather yield similar embeddings, whereas the same scene under different weather conditions produces distinct ones.  
In addition, for a fixed weather category $c$, embeddings derived from real and synthetic sources are encouraged to occupy the same region of the latent manifold, promoting cross-domain consistency.  
We qualitatively verify this property using t\textendash SNE~\cite{JMLR:v9:vandermaaten08a} visualizations, where successful alignment manifests as overlapping clusters for real and synthetic samples of the same weather type.

\subsubsection{Cross-domain Weather Alignment}
The goal of the \textit{Style Filter} is to cluster images affected by similar weather degradations while remaining invariant to content and domain. 
We adopt a contrastive loss that encourages embeddings for the same weather condition to be close in the latent space, while pushing embeddings from different weather categories apart. 
The loss is defined as:

\begin{equation}
\mathcal{L}_{con} = \sum_{(a,b)\in\mathcal{P}} \left[\mathbb{I}_{ab}[m - d_{ab}]_+ + (1-\mathbb{I}_{ab})d_{ab}\right],
\label{eq:contrastive}
\end{equation}

where $\mathcal{P}$ denotes all possible image pairs in a batch, 
$d_{ab} = d(\mathbf{w}_a, \mathbf{w}_b)$ is the cosine similarity between weather embeddings $\mathbf{w}_a$ and $\mathbf{w}_b$ extracted by \textit{Style Filter},
$m$ is a positive margin, 
$\mathbb{I}_{ab} = 1$ when images $(a,b)$ share the same weather type and $0$ otherwise, and 
$[\cdot]_+ = \max(0, \cdot)$.

To promote cross-domain alignment between real and synthetic data, 
we further treat real–synthetic pairs that share the same weather condition as positive samples in Eq.~\ref{eq:contrastive}.  
This modification differs from MWFormer~\cite{10767188}, which considers only intra-domain pairs.  
By explicitly mixing real and synthetic samples during contrastive training, 
the \textit{Style Filter} learns a unified embedding space in which images from both domains cluster according to solely weather type.  
Consequently, $\mathbf{w}$ becomes both \textbf{weather-discriminative} and \textbf{domain-aligned}, facilitating robust conditioning of the depth estimation backbone across real-world and synthetic degradations.

\subsubsection{Optional Designs}
Motivated by CLIP-style alignment, we also evaluated two variants. 
\textit{(i) Prompt-driven only ~\cite{fahes2022p}}  and \textit{(ii) Image–image alignment~\cite{fahes2024domain}}. 
Despite extensive tuning, real and synthetic samples of the same weather still formed separable clusters. We conjecture this is because CLIP features are dominated by semantics and invariant to low-level, physics-driven cues, making them insufficient to close the real–synthetic domain gap. This motivates our Style Filter, which learns weather-specific statistics via an explicit domain-alignment objective. Additional details and negative results are in the supplementary material.

\subsection{Weather-Conditioned Depth Anything}
In this section, we seek a concise way to inject the compact \emph{weather embedding} \textbf{w} into a depth foundation model without altering its core architecture. To achieve this, we propose a \emph{weather-conditioned modulation} tailored for DPT-based Depth Anything backbones. As shown in Figure \ref{fig:overview}, a 64-dimensional weather embedding $\mathbf{w}\in\mathbb{R}^{64}$ is fused at four multi-scale feature maps $\{F_i\!\in\!\mathbb{R}^{C_i\times H_i\times W_i}\}_{i=1}^{4}$ after their per-scale reductions to $C_i{=}64$, and once on the final fused feature before the output head (five sites in total). A lightweight MLP maps $\mathbf{w}$ to per-channel shift, scale, and a gate, denoted by $[\boldsymbol{\beta}_i(\mathbf{w}),\,\boldsymbol{\gamma}_i(\mathbf{w}),\,g_i(\mathbf{w})]$. The features are then updated via an \textit{AdaLN-Zero} style operation:
\begin{equation}
F_i \;\leftarrow\; F_i \;+\; g_i(\mathbf{w})\Big(\mathrm{LN}(F_i)\odot\big(1+\boldsymbol{\gamma}_i(\mathbf{w})\big)\;+\;\boldsymbol{\beta}_i(\mathbf{w})\Big),
\end{equation}
where $\mathrm{LN}$ is channel-wise LayerNorm without affine parameters and broadcasting over spatial dimensions $(H_i,W_i)$ is implicit. All modulation heads are \emph{zero-initialized}, ensuring the initial network output matches the base model. The conditioned path then learns weather-specific increments during training.

The proposed modulation introduces only a negligible parameter yet yields substantial gains in adverse-weather robustness. Moreover, we initialize the model from a pretrained Depth Anything~\cite{yang2024depth} and update only the decoder head and the small modulation heads, thereby preserving clean-domain behavior and mitigating catastrophic forgetting, effectively serving as a lightweight weather adapter for DPT-style Depth Anything backbones.

\subsection{Overall Loss}
Following Depth Anything~\cite{yang2024depth}, all supervision terms use the affine-invariant loss $\ell$ on disparity maps $y,\hat{y}\in\mathbb{R}^{H\times W}$:
\begin{equation}
\label{eq:ai_loss}
\ell(y,\hat{y})
=
\frac{1}{HW}\sum_{j=1}^{HW}
\left|
\frac{y_j - t(y)}{s(y)}
-
\frac{\hat{y}_j - t(\hat{y})}{s(\hat{y})}
\right|,
\end{equation}
where $t(\cdot)$ and $s(\cdot)$ denote per-image mean and mean absolute deviation, computed over valid pixels that are masked.

\paragraph{Data regime and objectives.}
We train with three data sources: (i) \emph{clean} images $\mathcal{D}$, (ii) \emph{synthetic weather} images $\mathcal{D}^s$ generated from $\mathcal{D}$, and (iii) \emph{real weather} images $\mathcal{D}^r$ without clean counterparts.

Let $\mathcal{T}_{\text{syn}}$ denote a family of geometry-preserving synthetic weather operators (fog/rain/snow/low-light). For each clean image $x \in \mathcal{D}$ and operator $\tau \in \mathcal{T}_{\text{syn}}$, we generate its synthetic counterpart $x^s = \tau(x) \in \mathcal{D}^s$, forming a paired sample $(x, x^s)$.

A frozen teacher $\mathcal{F}_t$ provides pseudo labels; the student $\mathcal{F}_\phi(\cdot;\mathbf{w})$ is conditioned by the \textit{Style Filter} embedding $\mathbf{w}(\cdot)$.
All discrepancies are measured on \emph{disparity} using the affine-invariant loss $\ell(\cdot,\cdot)$ in Eq.~\eqref{eq:ai_loss}.

\paragraph{Distillation loss.}
For clean images $x \in \mathcal{D}$, synthetic weather images $x^s \in \mathcal{D}^s$, and real weather images $x \in \mathcal{D}^r$, we align the student model (ViT-S) to the teacher (ViT-L). For synthetic images that share geometry with clean counterparts, we use the teacher's prediction on the corresponding clean image as the target:

\begin{equation}
\label{eq:distill_updated}
\mathcal{L}_{\mathrm{dis}}=
\begin{cases}
\ell\!\left(\mathcal{F}_\phi(x;\mathbf{w}(x)),\,\mathcal{F}_t(x)\right),
& x \in \mathcal{D}\cup\mathcal{D}^{r},\\
\ell\!\left(\mathcal{F}_\phi(x^{s};\mathbf{w}(x^{s})),\,\mathcal{F}_t(x)\right),
& x^{s}=\tau(x),\,x\in\mathcal{D}.
\end{cases}
\end{equation}

\paragraph{Alignment losses.}
For \emph{clean-synthetic} pairs $(x, x^s)$ where $x \in \mathcal{D}$ and $x^s \in \mathcal{D}^s$ share geometry, we enforce agreement between the two student predictions:
\begin{equation}
\label{eq:pair_align}
\mathcal{L}_{\mathrm{pair}} = \ell\big(\mathcal{F}_\phi(x;\mathbf{w}(x)), \mathcal{F}_\phi(x^s;\mathbf{w}(x^s))\big).
\end{equation}

For \emph{real weather} images $x \in \mathcal{D}^r$ without clean pairs, we apply a geometry-preserving appearance transform $\tau_{\text{aug}}$ (ColorJitter) and enforce augmentation consistency:
\begin{equation}
\label{eq:real_aug_align}
\mathcal{L}_{\mathrm{aug}} = \ell\big(\mathcal{F}_\phi(x;\mathbf{w}(x)), \mathcal{F}_\phi(\tau_{\text{aug}}(x);\mathbf{w}(\tau_{\text{aug}}(x)))\big).
\end{equation}

\begin{table*}[t]
  \caption{Zero-shot relative depth estimation on real adverse-weather benchmarks.
  ``--'' denotes unavailable results (not reported or not supported by the released model, e.g., MWFormer on NuScenes-night and RobotCar-night).}
  \label{tab:real_low_quality}
  \centering
  \small
  \setlength{\tabcolsep}{3pt}
  \renewcommand{\arraystretch}{0.95}
  \resizebox{\textwidth}{!}{%
  \begin{tabular}{l c cc cc cc cc cc}
    \toprule
    \multirow{2}{*}{Method} & \multirow{2}{*}{Encoder} &
    \multicolumn{2}{c}{NuScenes-night} &
    \multicolumn{2}{c}{RobotCar-night} &
    \multicolumn{2}{c}{DS-rain} &
    \multicolumn{2}{c}{DS-cloud} &
    \multicolumn{2}{c}{DS-fog} \\
    \cmidrule(lr){3-4}
    \cmidrule(lr){5-6}
    \cmidrule(lr){7-8}
    \cmidrule(lr){9-10}
    \cmidrule(lr){11-12}
    & & AbsRel$\downarrow$ & $\delta_1\uparrow$ & AbsRel$\downarrow$ & $\delta_1\uparrow$ & AbsRel$\downarrow$ & $\delta_1\uparrow$ & AbsRel$\downarrow$ & $\delta_1\uparrow$ & AbsRel$\downarrow$ & $\delta_1\uparrow$ \\
    \midrule
    DynaDepth~\cite{zhang2022towards} & ResNet
      & 0.381 & 0.394 & 0.512 & 0.294 & 0.239 & 0.606 & 0.172 & 0.608 & 0.144 & \topA{{0.901}} \\
    EC-Depth~\cite{song2023ec} & ViT-S
      & 0.243 & 0.623 & \topC{0.228} & \topC{0.552} & 0.155 & 0.766 & 0.158 & 0.767 & 0.109 & 0.861 \\
    STEPS~\cite{zheng2023steps} & ResNet
      & 0.252 & 0.588 & 0.350 & 0.367 & 0.301 & 0.480 & 0.252 & 0.588 & 0.216 & 0.641 \\
    robustdepth~\cite{saunders2023self} & MonoViT
      & 0.260 & 0.597 & 0.311 & 0.521 & 0.167 & 0.755 & 0.168 & 0.775 & 0.105 & 0.882 \\
    weather-depth~\cite{wang2024weatherdepth} & ViT-S
      & -- & -- & -- & -- & 0.158 & 0.764 & 0.160 & 0.767 & 0.105 & 0.879 \\
    Syn2Real~\cite{yan2025synthetic} & ViT-S
      & -- & -- & -- & -- & 0.171 & 0.729 & -- & -- & 0.128 & 0.845 \\
    DepthPro~\cite{bochkovskii2024depth} & ViT-S
      & 0.218 & 0.669 & 0.237 & 0.534 & \topB{{0.124}} & \topB{{0.841}} & 0.158 & 0.779 & \topC{{0.102}} & \topC{0.892} \\
    DepthAnything v1~\cite{depthanything} & ViT-S
      & 0.232 & 0.679 & 0.239 & 0.518 & 0.133 & 0.819 & \topB{0.150} & \topA{{0.801}} & \topA{{0.098}} & 0.891 \\
    DepthAnything v2~\cite{yang2024depth} & ViT-S
      & \topC{0.200} & \topC{0.725} & 0.239 & 0.518 & \topC{0.125} & \topC{0.840} & \topC{0.151} & \topB{0.798} & {0.103} & 0.890 \\
    DepthAnything v3~\cite{depthanything3} & ViT-S
      & 0.275 & 0.586 & \topA{0.169} & \topA{0.755} & 0.134 & 0.822 & 0.154 & 0.785 & 0.105 & 0.887 \\
    DepthAnything-AC~\cite{sun2025depth} & ViT-S
      & \topB{{0.198}} & \topB{{0.727}} & \topB{0.227} & \topB{0.555} & \topC{0.125} & \topC{0.840} & \topA{{0.149}} & \topA{{0.801}} & {0.103} & 0.889 \\
    MWFormer~\cite{10767188}+DA v2~\cite{yang2024depth} & ViT-S
      & -- & -- & -- & -- & 0.127 & 0.833 & \topC{0.151} & \topC{0.795} & 0.104 & 0.884 \\
    DarkIR~\cite{Feijoo_2025_CVPR}+DA v2~\cite{yang2024depth} & ViT-S
      & 0.204 & 0.710 & {0.232} & {0.542} & -- & -- & -- & -- & -- & -- \\
    \midrule
    \textbf{DepthAnything-W} & ViT-S
      & \topA{{0.194}} & \topA{{0.737}} & 0.239 & 0.513 & \topA{{0.123}} & \topA{{0.842}} & \topC{{0.151}} & \topC{{0.795}} & \topB{{0.101}} & \topB{{0.896}} \\
    \bottomrule
  \end{tabular}%
  }
\end{table*}

%%%%%%%%%%%%%%%%%%%%%%%%%%%%
%%%%%%%%%%%%%%%

\paragraph{Total objective.}
The final loss combines teacher guidance with scene-pair alignment (\emph{clean-synthetic}) and augmentation consistency (\emph{real weather}):
\begin{equation}
\label{eq:total_updated}
\mathcal{L}
\;=\;
\mathcal{L}_{\mathrm{dis}}
\;+\;
\mathcal{L}_{\mathrm{pair}}
\;+\;
\mathcal{L}_{\mathrm{aug}},
\end{equation}

\section{Implementation}

\begin{table*}[t]
  \caption{Zero-shot relative depth estimation on KITTI-C (synthetic corruptions). \emph{``--'' denotes unavailable results.}}
  \label{tab:kittic_synthetic}
  \centering
  \small
  \setlength{\tabcolsep}{4pt}
  \renewcommand{\arraystretch}{0.95}
  \resizebox{\textwidth}{!}{%
  \begin{tabular}{l c cc cc cc cc}
    \toprule
    \multirow{2}{*}{Method} & \multirow{2}{*}{Encoder} &
    \multicolumn{2}{c}{Dark} &
    \multicolumn{2}{c}{Snow} &
    \multicolumn{2}{c}{Fog} &
    \multicolumn{2}{c}{Motion} \\
    \cmidrule(lr){3-4}
    \cmidrule(lr){5-6}
    \cmidrule(lr){7-8}
    \cmidrule(lr){9-10}
    & & AbsRel$\downarrow$ & $\delta_1\uparrow$
      & AbsRel$\downarrow$ & $\delta_1\uparrow$
      & AbsRel$\downarrow$ & $\delta_1\uparrow$
      & AbsRel$\downarrow$ & $\delta_1\uparrow$ \\
    \midrule
    DepthPro~\cite{bochkovskii2024depth} & ViT\textendash S
      & 0.145 & 0.793
      & 0.197 & 0.685
      & \topA{0.087} & \topA{0.926}
      & 0.170 & 0.746 \\
    DepthAnything v1~\cite{depthanything} & ViT\textendash S
      & \topB{0.129} & 0.826
      & \topC{0.113} & 0.870
      & \topB{0.092} & \topB{0.912}
      & 0.133 & \topC{0.822} \\
    DepthAnything v2~\cite{yang2024depth} & ViT\textendash S
      & \topC{0.130} & \topC{0.832}
      & 0.115 & \topC{0.872}
      & 0.097 & 0.905
      & \topC{0.127} & \topB{0.840} \\
      DepthAnything v3~\cite{depthanything3} & ViT\textendash S
      & 0.180 & 0.737
      & 0.182 & 0.715
      & 0.119 & 0.858
      & 0.196 & 0.698 \\
    DepthAnything-AC~\cite{sun2025depth} & ViT\textendash S
      & \topC{0.130} & \topB{0.834}
      & 0.114 & \topB{0.873}
      & 0.096 & 0.906
      & \topB{0.126} & \topA{0.841} \\
    MWFormer~\cite{10767188} + DA v2~\cite{yang2024depth} & ViT\textendash S
      & -- & --
      & \topB{0.109} & \topA{0.884}
      & 0.099 & 0.899
      & -- & -- \\
    DarkIR~\cite{Feijoo_2025_CVPR} + DA v2~\cite{yang2024depth} & ViT\textendash S
      & 0.148 & 0.803
      & -- & --
      & -- & --
      & -- & --\\
    \midrule
    \textbf{DepthAnything-W} & ViT\textendash S
      & \topA{0.126} & \topA{0.837}
      & \topA{0.107} & \topA{0.884}
      & \topC{0.093} & \topC{0.910}
      & \topA{0.125} & \topA{0.841} \\
    \bottomrule
  \end{tabular}%
  }
\end{table*}

\subsection{Dataset Curation}
To learn a weather-embedding invariant across domains, we curate 7 real-world adverse-weather datasets and 4 synthetic weather datasets, along with their corresponding synthetic weather views.
The adverse-weather collection comprises ACDC~\cite{sakaridis2025acdc}, RTTS~\cite{fang2024real, 8451944}, Snow100K~\cite{liu2018desnownet}, Muses~\cite{brodermann2024muses}, RID and RIS~\cite{8954466}, and NightCity~\cite{Tan_2021_TIP_NightCity}, totaling approximately 15K images covering fog, rain, snow, and low-light conditions.
The clean datasets include COCO~\cite{lin2014microsoft}, MegaDepth~\cite{li2018megadepth}, SA-1B~\cite{kirillov2023segment}, and HRWSI~\cite{gaidon2016virtual, xian2020structure}, providing approximately 200K high-quality images.

During \textit{Style Filter} training (Stage I), we uniformly subsample 20K images from the clean datasets to balance the scale with real-world adverse weather data, ensuring effective cross-domain learning without category imbalance.
Following the synthetic degradation recipe in RobustSAM~\cite{chen2024robustsam}, we generate corresponding synthetic weather corruptions for each clean image, creating paired \emph{clean-synthetic} views with shared underlying geometry. 

% This real-synthetic pairing enables the \textit{Style Filter} to learn degradation-aware embeddings that remain content-agnostic and domain-invariant.
During depth adaptation training (Stage II), we use the entire dataset, comprising all real adverse-weather images and synthetic corruptions. For depth supervision, we use validity masks when provided by the datasets; otherwise, we generate sky masks using Lang-SAM~\cite{langsam2023} and exclude these regions to avoid ill-posed supervision for unbounded depth. 
Detailed dataset statistics are provided in the supplementary material.

\begin{figure*}[t!]
  \centering
  \setlength{\tabcolsep}{1pt}
  \renewcommand{\arraystretch}{1.0}
  \newcommand{\imgwidth}{.245\textwidth}
  \resizebox{\textwidth}{!}{
  % \begin{tabular}{@{}m{0.4cm}m{3.9cm}m{3.9cm}m{3.9cm}m{3.9cm}@{}}
  \begin{tabular}{@{}c c c c c@{}}
    & \centering\textbf{Input} & \centering\textbf{DA v2~\cite{yang2024depth}} & \centering\textbf{DA-AC~\cite{sun2025depth}} & \centering\arraybackslash\textbf{DA-W(Ours)} \\
    % \hline
    % \rotatebox[origin=c]{90}{\textbf{Fog}} &
    \raisebox{3em}{\rotatebox[origin=c]{90}{\textbf{Fog}}} &
    \includegraphics[width=\imgwidth]{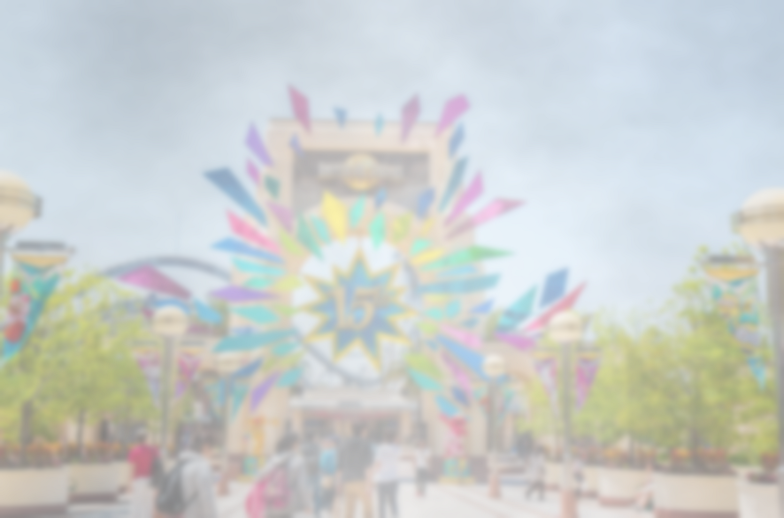} &
    \includegraphics[width=\imgwidth]{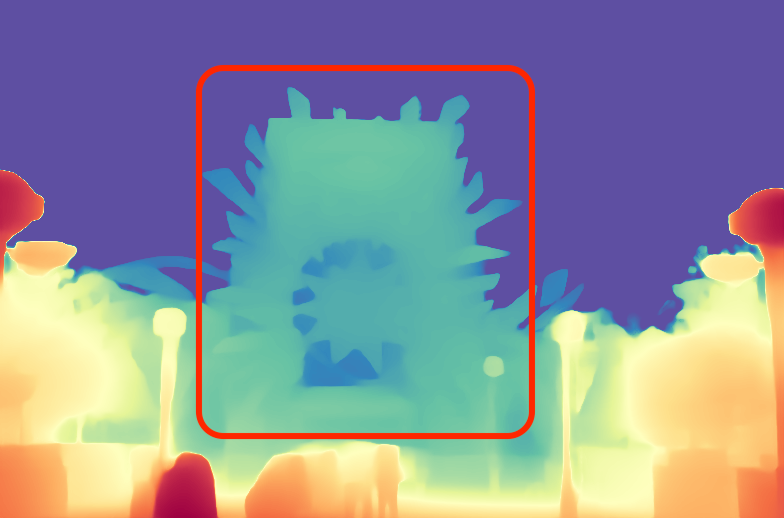} &
    \includegraphics[width=\imgwidth]{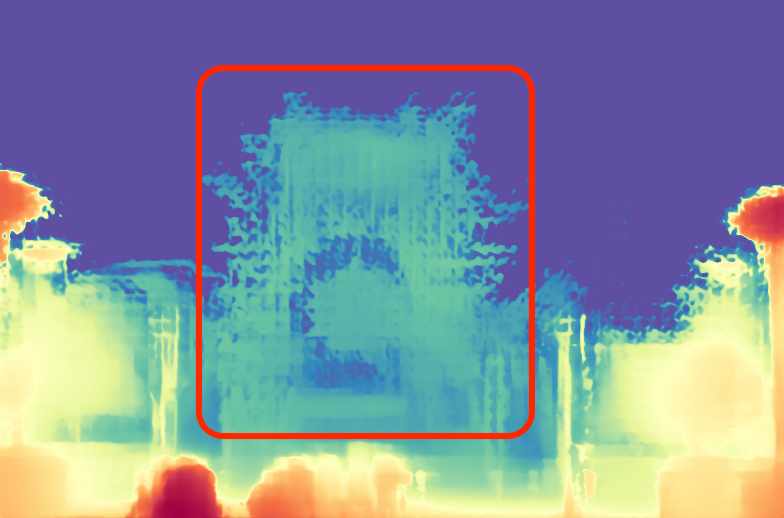} &
    \includegraphics[width=\imgwidth]{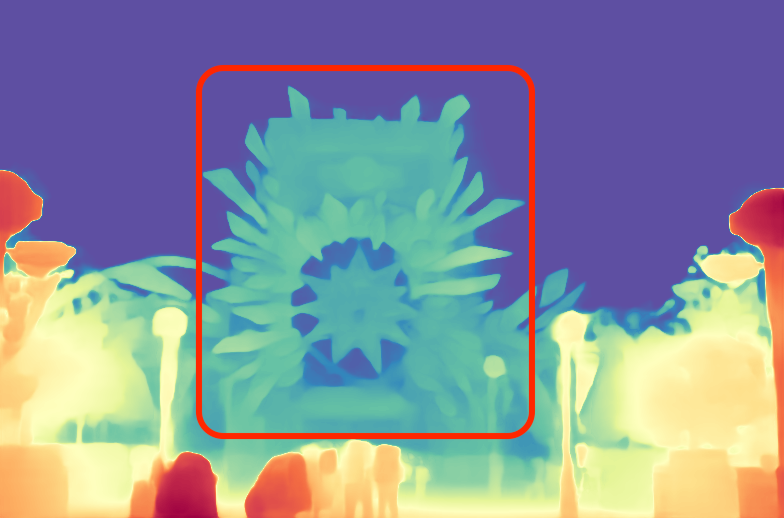} \\
    
    % \rotatebox[origin=c]{90}{\textbf{Lowlight}} &
    % \raisebox{1.6em}{\rotatebox[origin=c]{90}{\textbf{Lowlight}}} &
    % \includegraphics[width=\imgwidth]{images/supp/dark/dark_img.png} &
    % \includegraphics[width=\imgwidth]{images/supp/dark/da2_dark.png} &
    % \includegraphics[width=\imgwidth]{images/supp/dark/daac_dark.png} &
    % \includegraphics[width=\imgwidth]{images/supp/dark/daw_dark.png} \\
    
    % \raisebox{1.2em}{\rotatebox[origin=c]{90}{\textbf{Rain}}} &
    % \includegraphics[width=\imgwidth]{images/supp/rain/57_img.png} &
    % \includegraphics[width=\imgwidth]{images/supp/rain/da_57_pred copy.png} &
    % \includegraphics[width=\imgwidth]{images/supp/rain/ac_57_pred copy.png} &
    % \includegraphics[width=\imgwidth]{images/supp/rain/57_pred copy.png} \\
    
    \raisebox{3em}{\rotatebox[origin=c]{90}{\textbf{Snow}}} &
    \includegraphics[width=\imgwidth]{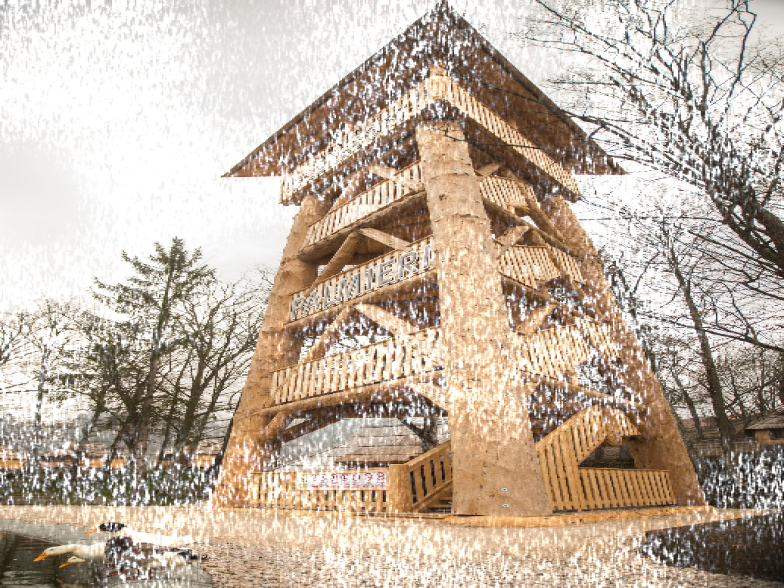} &
    \includegraphics[width=\imgwidth]{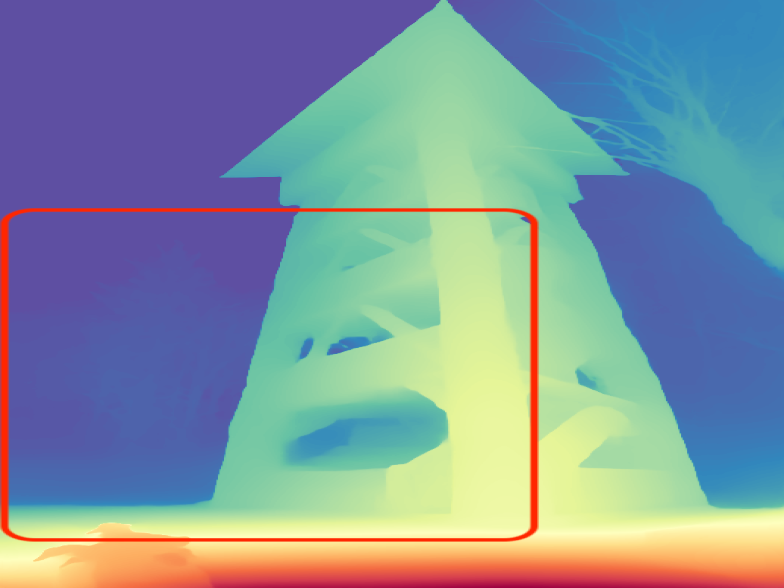} &
    \includegraphics[width=\imgwidth]{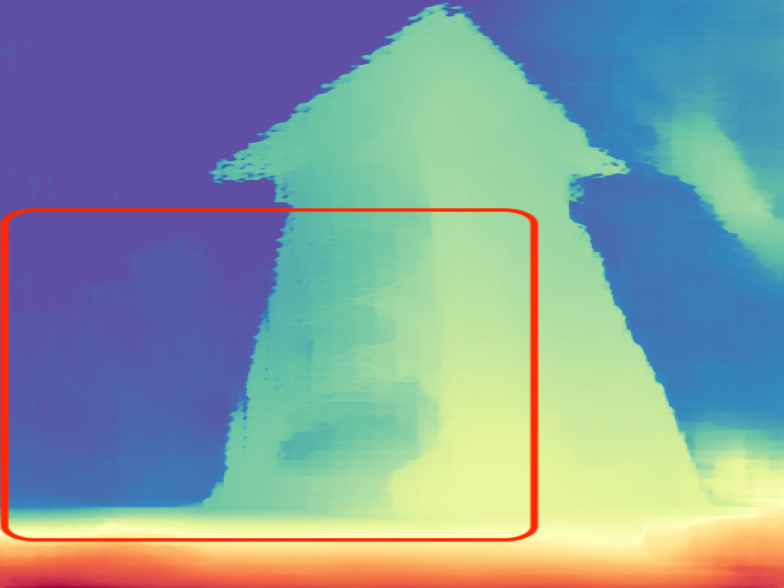} &
    \includegraphics[width=\imgwidth]{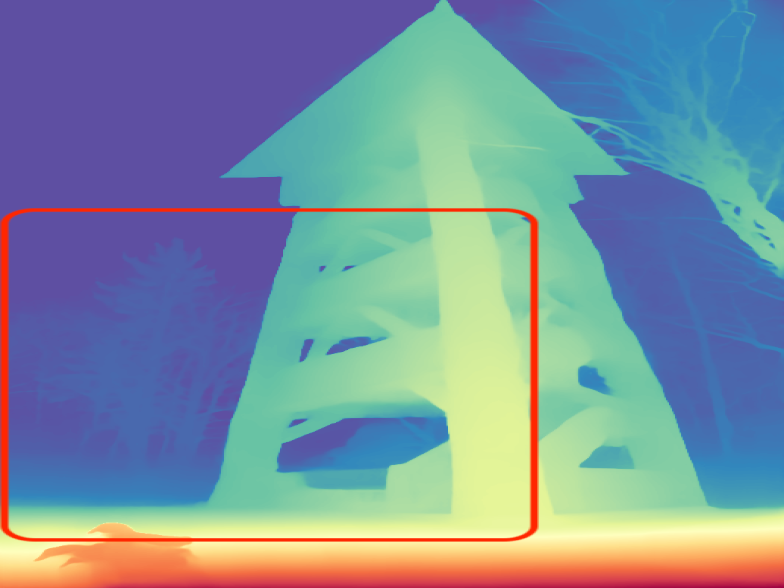} \\
  \end{tabular}
  }
    \vspace{-4mm}
  \caption{\textbf{Qualitative comparison on various synthetic weather degradations.} Our method (DA-W) produces more accurate and stable depth predictions under challenging synthetic weather conditions. The red boxes highlight regions where DA-W maintains clearer structural details and smoother depth transitions, while other methods suffer from noisy artifacts and depth discontinuities.}
  \label{fig:qualitative_ori}
\end{figure*}

\subsection{Training Details}
We adopt a parameter-efficient two-stage training protocol. 
\textbf{Stage I:} We train the \textit{Style Filter} on the combined real and synthetic weather data to learn a compact 64-D weather embedding. The training procedure is described in MWFormer~\cite{10767188}.
\textbf{Stage II:} We optimize only the DPT decoder head and the \textit{AdaLN-Zero} modulation layers, which inject the weather embedding at five decoder stages. We train for 20 epochs using AdamW with a learning rate of 5e-6, weight decay of 0.01, and a batch size of 16. All experiments are conducted on a single NVIDIA A100 GPU. 

\subsection{Evaluation Protocol}
We follow the evaluation protocol of DepthAnything-AC~\cite{sun2025depth}, using standard relative depth metrics (AbsRel and $\delta_1$) across two benchmark categories.
\textbf{Adverse-weather benchmarks:} We evaluate on real world adverse weather datasets NuScenes-night~\cite{caesar2020nuscenes}, RobotCar-night~\cite{maddern20171}, DrivingStereo-rain/cloud/fog~\cite{yan2025synthetic, yang2019drivingstereo}, and the synthetic weather corruption datasets KITTI-C (Dark, Snow, Fog, Motion)~\cite{kong2023robodepth}.
\textbf{Mix-weather benchmarks:} To address the practical setting where multiple weather conditions may coexist (e.g., rainy nights or snowy nights), we introduce a \emph{mix-weather} benchmark composed of both synthetic and real-world subsets. 
% For the \emph{synthetic} subset, we adopt the KITTI Eigen split and apply the corruption composition strategy of RoboDepth~\cite{kong2023robodepth}, evaluating combinations of weather corruptions. For the \emph{real-world} subset, we use the \textit{night} and \textit{snow} sequences from Boreas~\cite{burnett_ijrr23} for qualitative visualization, focusing on mixed-condition failure modes and cross-condition generalization.
\textbf{Clean-domain benchmarks:} To verify that weather adaptation does not degrade general performance, we report results on KITTI~\cite{geiger2013vision}, NYU-Depth v2~\cite{silberman2012indoor}, Sintel~\cite{butler2012naturalistic}, ETH3D~\cite{schops2017multi}, and DIODE~\cite{vasiljevic2019diode}.

\section{Experiment}

\begin{table*}[t]
  \caption{Relative depth estimation results on general benchmarks. Our method preserves DepthAnything's performance on clean scenes. Best/second best are \textbf{bold}/\underline{underlined}.}
  \label{tab:general_benchmark_results}
  \centering
  \small
  \setlength{\tabcolsep}{4pt}
  \renewcommand{\arraystretch}{0.95}
  \resizebox{\textwidth}{!}{%
  \begin{tabular}{l c cc cc cc cc cc}
    \toprule
    \multirow{2}{*}{Method} & \multirow{2}{*}{Encoder} &
    \multicolumn{2}{c}{KITTI~\cite{geiger2013vision}} &
    \multicolumn{2}{c}{NYU-D~\cite{silberman2012indoor}} &
    \multicolumn{2}{c}{Sintel~\cite{butler2012naturalistic}} &
    \multicolumn{2}{c}{ETH3D~\cite{schops2017multi}} &
    \multicolumn{2}{c}{DIODE~\cite{vasiljevic2019diode}} \\
    \cmidrule(lr){3-4}
    \cmidrule(lr){5-6}
    \cmidrule(lr){7-8}
    \cmidrule(lr){9-10}
    \cmidrule(lr){11-12}
    & & AbsRel$\downarrow$ & $\delta_1\uparrow$
      & AbsRel$\downarrow$ & $\delta_1\uparrow$
      & AbsRel$\downarrow$ & $\delta_1\uparrow$
      & AbsRel$\downarrow$ & $\delta_1\uparrow$
      & AbsRel$\downarrow$ & $\delta_1\uparrow$ \\
    \midrule
    DynaDepth~\cite{zhang2022towards} & ResNet
      & 0.085 & 0.916 & 0.212 & 0.630 & 0.458 & 0.391 & 0.192 & 0.703 & 0.180 & 0.737 \\
    EC-Depth~\cite{song2023ec} & ViT-S
      & \underline{0.075} & 0.931 & 0.220 & 0.606 & 0.447 & 0.405 & 0.181 & 0.779 & 0.172 & 0.760 \\
    STEPS~\cite{zheng2023steps} & ResNet
      & 0.204 & 0.683 & 0.220 & 0.606 & 0.458 & 0.375 & 0.197 & 0.699 & 0.214 & 0.652 \\
    robustdepth~\cite{saunders2023self} & MonoViT
      & 0.076 & 0.929 & 0.174 & 0.718 & 0.411 & 0.436 & 0.185 & 0.776 & 0.182 & 0.741 \\
    weather-depth~\cite{wang2024weatherdepth} & ViT-S
      & \underline{0.075} & 0.932 & 0.167 & 0.733 & 0.419 & 0.443 & 0.181 & 0.771 & 0.170 & 0.760 \\
    DepthPro~\cite{bochkovskii2024depth} & ViT-S
      & \textbf{0.068} & \textbf{0.947} & \underline{0.052} & \underline{0.972} & 0.240 & 0.682 & \textbf{0.057} & 0.963 & \textbf{0.060} & \textbf{0.952} \\
    DepthAnything v1~\cite{depthanything} & ViT-S
      & 0.082 & 0.936 & 0.053 & \underline{0.972} & \textbf{0.211} & \textbf{0.736} & \underline{0.064} & 0.966 & 0.078 & 0.940 \\
    DepthAnything v2~\cite{yang2024depth} & ViT-S
      & 0.083 & 0.934 & \textbf{0.051} & \textbf{0.973} & 0.240 & 0.712 & \underline{0.064} & \textbf{0.969} & 0.073 & \underline{0.943} \\
    DepthAnything-AC~\cite{sun2025depth} & ViT-S
      & 0.083 & 0.934 & \textbf{0.051} & \textbf{0.973} & 0.235 & 0.715 & {0.065} & \underline{0.968} & 0.073 & 0.941 \\
    \midrule
    \textbf{DepthAnything-W} & {ViT-S}
      & 0.080 & \underline{0.937} & \textbf{0.051} & \textbf{0.973} & \underline{0.232} & \underline{0.716} & \underline{0.064} & \underline{0.968} & \underline{0.071} & 0.942 \\
    \bottomrule
  \end{tabular}%
  }
\end{table*}

\subsubsection{Comparison on real adverse weather.}
Table~\ref{tab:real_low_quality} presents a summary of zero-shot relative depth estimation performance across five real-world adverse-weather tracks.
\textbf{DepthAnything-W} demonstrates the highest overall robustness within the DepthAnything family  on \textit{NuScenes-night} and \textit{DS-rain}, outperforming both DepthAnything v2 and DepthAnything-AC in terms of \emph{both} AbsRel and $\delta_1$.
On \textit{DS-cloud} and \textit{DS-fog}, our method remains competitive but does not achieve the top score. These conditions, characterized by reduced global contrast and partially suppressed fine-grained textures, may still require specialized approaches.
The most challenging setting for our model is \textit{RobotCar-night}, where methods such as DepthAnything v3 and DepthPro perform substantially better. This performance gap likely reflects domain characteristics, including camera response, exposure dynamics, and scene statistics, that are underrepresented in the training mixture.
Finally, restoration-based two-stage pipelines (e.g., MWFormer + DepthAnything v2) can be competitive on some DrivingStereo subsets (e.g., \textit{DS-fog}: AbsRel $0.104$), but they require additional preprocessing.
Overall, these results support our hypothesis that explicit weather embeddings with lightweight decoder-only AdaLN conditioning provide a practical alternative for improving adverse-weather robustness while preserving strong generalization.

\subsubsection{Synthetic corruptions (KITTI-C).}
Table~\ref{tab:kittic_synthetic} evaluates robustness under four synthetic corruptions. 
\textbf{DepthAnything-W} achieves the best performance on \textit{Dark}, \textit{Snow}, and \textit{Motion}, and remains competitive on \textit{Fog}. 
Compared with DepthAnything v2, our weather-conditioned model consistently improves both metrics across all settings.
The largest gain appears on \textit{Snow}, suggesting that domain-aligned weather embeddings help preserve cues related to scattering and partial occlusions that are easily suppressed by generic representations.
On \textit{Fog}, we observe a smaller margin and slightly trail methods that explicitly target this corruption, indicating that dense fog remains challenging when appearance cues are heavily washed out.
Notably, our conditioning approach is comparable to restoration-based or two-stage pipelines while avoiding additional preprocessing and the potential for error accumulation, offering a simpler and more efficient path to corruption robustness.
Figure~\ref{fig:qualitative_ori} qualitatively illustrates these improvements, with additional results provided in the Appendix.

\begin{table}[t]
    \caption{Mix-weather benchmark on KITTI Eigen split with composed corruptions (RoboDepth~\cite{kong2023robodepth}). 
    We report $\delta_1 \uparrow$ and AbsRel $\downarrow$. Best/second best are \textbf{bold}/\underline{underlined}.}
    \label{tab:mix_weather}
    \centering
    \resizebox{\linewidth}{!}{%
    \begin{tabular}{l cc cc cc cc cc}
        \toprule
            \multirow{2}{*}{Corruption} 
            & \multicolumn{2}{c}{DA v1~\cite{depthanything}} 
            & \multicolumn{2}{c}{DA v2~\cite{yang2024depth}} 
            & \multicolumn{2}{c}{DA v3~\cite{depthanything3}} 
            & \multicolumn{2}{c}{DA-AC~\cite{sun2025depth}} 
            & \multicolumn{2}{c}{\textbf{DA-W (Ours)}} \\
            \cmidrule(lr){2-3}\cmidrule(lr){4-5}\cmidrule(lr){6-7}\cmidrule(lr){8-9}\cmidrule(lr){10-11}
            & $\delta_1\uparrow$ & AbsRel$\downarrow$
            & $\delta_1\uparrow$ & AbsRel$\downarrow$
            & $\delta_1\uparrow$ & AbsRel$\downarrow$
            & $\delta_1\uparrow$ & AbsRel$\downarrow$
            & $\delta_1\uparrow$ & AbsRel$\downarrow$ \\
            \midrule
            Clean               & \textbf{0.937} & \underline{0.082} & \underline{0.934} & 0.083 & 0.912 & 0.094 & 0.934 & 0.083 & \textbf{0.937} & \textbf{0.080} \\
            Fog+Rain            & 0.708 & 0.192 & 0.726 & 0.185 & 0.636 & 0.225 & \underline{0.731} & \underline{0.181} & \textbf{0.750} & \textbf{0.173} \\
            Fog+Snow            & 0.681 & 0.205 & 0.712 & 0.192 & 0.481 & 0.303 & \underline{0.731} & \underline{0.179} & \textbf{0.734} & \textbf{0.177} \\
            Fog+Night           & 0.715 & 0.185 & 0.738 & 0.178 & 0.640 & 0.233 & \underline{0.742} & \underline{0.175} & \textbf{0.747} & \textbf{0.171} \\
            Rain+Snow           & 0.671 & 0.209 & \underline{0.754} & 0.171 & 0.627 & 0.228 & \underline{0.754} & \underline{0.167} & \textbf{0.771} & \textbf{0.158} \\
            Rain+Night          & 0.658 & 0.217 & \underline{0.680} & 0.210 & 0.657 & 0.224 & \underline{0.680} & \underline{0.207} & \textbf{0.714} & \textbf{0.192} \\
            Snow+Night          & 0.724 & 0.182 & 0.742 & 0.174 & 0.639 & 0.230 & \textbf{0.756} & \underline{0.166} & \underline{0.754} & \textbf{0.165} \\
            Fog+Rain+Snow       & 0.562 & 0.269 & 0.596 & 0.251 & 0.451 & 0.322 & \underline{0.634} & \underline{0.230} & \textbf{0.670} & \textbf{0.214} \\
            Fog+Rain+Night      & 0.661 & 0.222 & \underline{0.668} & \underline{0.218} & 0.546 & 0.283 & 0.638 & 0.232 & \textbf{0.673} & \textbf{0.217} \\
            Fog+Snow+Night      & 0.529 & 0.287 & 0.577 & \underline{0.262} & 0.496 & 0.307 & \underline{0.617} & \textbf{0.237} & \textbf{0.623} & \textbf{0.237} \\
            Rain+Snow+Night     & 0.616 & 0.238 & 0.645 & 0.225 & 0.611 & 0.250 & \underline{0.679} & \underline{0.207} & \textbf{0.696} & \textbf{0.197} \\
            Fog+Rain+Snow+Night & 0.545 & 0.278 & 0.562 & 0.268 & 0.501 & 0.309 & \underline{0.608} & \underline{0.245} & \textbf{0.630} & \textbf{0.236} \\
        \bottomrule
    \end{tabular}%
}
\end{table}

\subsubsection{General benchmarks (clean scenes).}
To assess catastrophic forgetting, we evaluate on standard clean datasets (Table~\ref{tab:general_benchmark_results}).
\textbf{DepthAnything-W} maintains the backbone’s performance, with small but consistent gains on \textit{KITTI} and \textit{Sintel}, parity on \textit{ETH3D}.
This indicates that decoder-side AdaLN with zero-initialized heads preserves clean-scene behavior and improves robustness under adverse weather conditions.

\subsubsection{Mix-weather benchmark (coexisting conditions).}
Real driving often exhibits co-occurring degradations, such as rainy nights or snowy nights, which are not covered by single-factor benchmarks.
To evaluate this setting, we construct a \emph{mix-weather} benchmark with both synthetic and real-world subsets.
For the \emph{synthetic} subset, we follow RoboDepth~\cite{kong2023robodepth} to compose weather corruptions on the KITTI Eigen split, testing 11 mixed corruption combinations across 5 severity levels.
As summarized in Table~\ref{tab:mix_weather}, \textbf{DepthAnything-W} consistently improves both $\delta_1$ and AbsRel over DepthAnything v1/v2/v3 and DepthAnything-AC across most mixed-condition tracks, indicating that explicit weather embeddings and decoder-only AdaLN conditioning remain effective even when multiple condition factors overlap.
For the \emph{real-world} subset, we qualitatively evaluate on Boreas~\cite{burnett_ijrr23} \textit{night}+\textit{snow} sequences, where our predictions exhibit fewer structural artifacts and more stable geometry under simultaneous low-illumination and snowfall (see Figure.~\ref{fig:qualitative_mix}).
These results indicate that adaptation to individual weather conditions can transfer to \emph{coexisting} degradations: our conditioning design remains effective on mixed corruptions without requiring restoration-stage preprocessing.

\subsection{Ablation Study}
Table~\ref{tab:ablation_main} shows three design choices: fine-tuning strategy (full or decoder-only), data composition (with or without real-world degradations), and AdaLN-based weather conditioning. Although no single configuration dominates all metrics, the complete decoder-only variant with real-world degradations and weather injection offers the best overall trade-off.

\paragraph{Fine-tuning strategy.} Transitioning from full fine-tuning to decoder-only tuning gives a better overall trade-off across adverse and clean benchmarks. Specifically, decoder-only tuning results in higher $\delta_1$ on \textit{NuScenes-night} (0.741 versus 0.727) and enhances clean-scene generalization on \textit{NYU-D} and \textit{KITTI}. These findings indicate that the pretrained encoder is sufficiently general and that freezing it helps prevent representation drift.
\paragraph{Data composition.} Incorporating real-world degradations during training primarily benefits real adverse-weather domains, as demonstrated by clear gains on \textit{RobotCar-night} and a modest improvement on \textit{DS-rain}, while clean benchmarks remain largely unaffected. A minor trade-off is observed on \textit{NuScenes-night}, indicating a residual domain mismatch between the real-night data characteristics and this benchmark.
\paragraph{Weather conditioning.} Enabling AdaLN-based weather injection further enhances nighttime robustness, particularly on \textit{RobotCar-night} and to a lesser extent on \textit{NuScenes-night }, while maintaining strong performance on clean benchmarks (\textit{KITTI}: 0.080/0.937). For \textit{DrivingStereo} corruptions, the effect is mixed, as performance is slightly reduced on \textit{DS-fog}, suggesting that not all synthetic corruptions benefit equally from explicit weather conditioning.
\paragraph{Loss and restoration ablations.} Table~\ref{tab:loss_ablation} isolates the objective terms. Removing $L_{\rm dis}$ causes the largest degradation, increasing Global AbsRel from 0.117 to 0.352, which confirms that teacher distillation anchors the adapted model to geometry. $L_{\rm pair}$ provides a small but consistent gain, while $L_{\rm aug}$ shows no measurable independent gain under this protocol and is best interpreted as a real-weather consistency regularizer. The RGB restoration baseline disables Style Filter/AdaLN conditioning and adds a lightweight degraded-to-clean RGB head trained with masked L1 loss on synthetic pairs, weighted by $\lambda_{\rm rgb}=0.1$. It is competitive but does not improve the Global score over DA-W, suggesting that weather-conditioned modulation is not merely image restoration. Full fine-tuning improves the real-night average but is not better globally, supporting decoder-only adaptation as the better robustness-generalization trade-off.

\begin{table*}[t]
  \caption{\textbf{Ablation across training and weather-conditioning choices.}
  Metrics are AbsRel$\downarrow$ / $\delta_1\uparrow$.
  Avg. rank$\downarrow$ is computed over all 14 reported metrics with ties averaged.
  Bold and underlined denote the best and second-best values for each metric, respectively.}
  \label{tab:ablation_main}
  \centering
  \scriptsize
  \setlength{\tabcolsep}{2.2pt}
  \renewcommand{\arraystretch}{1.0}
  \resizebox{\textwidth}{!}{%
  \begin{tabular}{@{}cccccccccccc@{}}
    \toprule
    \multicolumn{4}{c}{Config (\cmark\ = enabled)} &
    \multicolumn{7}{c}{Zero-shot (AbsRel$\downarrow$ / $\delta_1\uparrow$)} &
    \multicolumn{1}{c}{Overall} \\
    \cmidrule(lr){1-4}\cmidrule(lr){5-11}\cmidrule(l){12-12}
    \makecell{Encoder\\FT} &
    \makecell{Decoder\\FT} &
    \makecell{Real\\Deg.} &
    \makecell{Weather\\Inject.} &
    NS-N & RC-N & DS-Rain & DS-Fog & NYU-D & KITTI & KITTI-Snow &
    \makecell{Avg.\\rank$\downarrow$} \\
    \midrule
    \cmark & \cmark & -- & --
    & \underline{0.194}/0.727
    & 0.245/0.493
    & \underline{0.122}/\textbf{0.842}
    & \textbf{0.098}/0.898
    & 0.056/0.968
    & 0.084/0.932
    & \underline{0.109}/\textbf{0.888}
    & 2.89 \\

    -- & \cmark & -- & --
    & \textbf{0.190}/\textbf{0.741}
    & 0.248/0.495
    & 0.123/0.840
    & 0.102/\textbf{0.902}
    & \textbf{0.050}/\textbf{0.974}
    & \underline{0.081}/\underline{0.936}
    & 0.113/0.877
    & 2.68 \\

    -- & \cmark & \cmark & --
    & 0.195/0.731
    & \underline{0.243}/\underline{0.507}
    & \textbf{0.121}/\textbf{0.842}
    & \underline{0.100}/\underline{0.901}
    & \textbf{0.050}/\textbf{0.974}
    & \underline{0.081}/\underline{0.936}
    & 0.112/0.878
    & \underline{2.29} \\

    -- & \cmark & \cmark & \cmark
    & \underline{0.194}/\underline{0.737}
    & \textbf{0.239}/\textbf{0.513}
    & {0.123}/\textbf{0.842}
    & {0.101}/{0.896}
    & \underline{0.051}/\underline{0.973}
    & \textbf{0.080}/\textbf{0.937}
    & \textbf{0.107}/\underline{0.884}
    & \textbf{2.14} \\
    \bottomrule
  \end{tabular}%
  }
\end{table*}

\begin{figure*}[t!]
  \centering
  \setlength{\tabcolsep}{1pt}
  \renewcommand{\arraystretch}{1.0}
  \resizebox{\textwidth}{!}{
  \begin{tabular}{@{}c c c c c c@{}}
    & \textbf{Input} & \textbf{DA v1~\cite{depthanything}} & \textbf{DA v2~\cite{yang2024depth}} & \textbf{DA-AC~\cite{sun2025depth}} & \textbf{Ours} \\

    \raisebox{3em}{\rotatebox[origin=c]{90}{\textbf{}}} &
    \includegraphics[width=.19\textwidth,height=.115\textwidth]{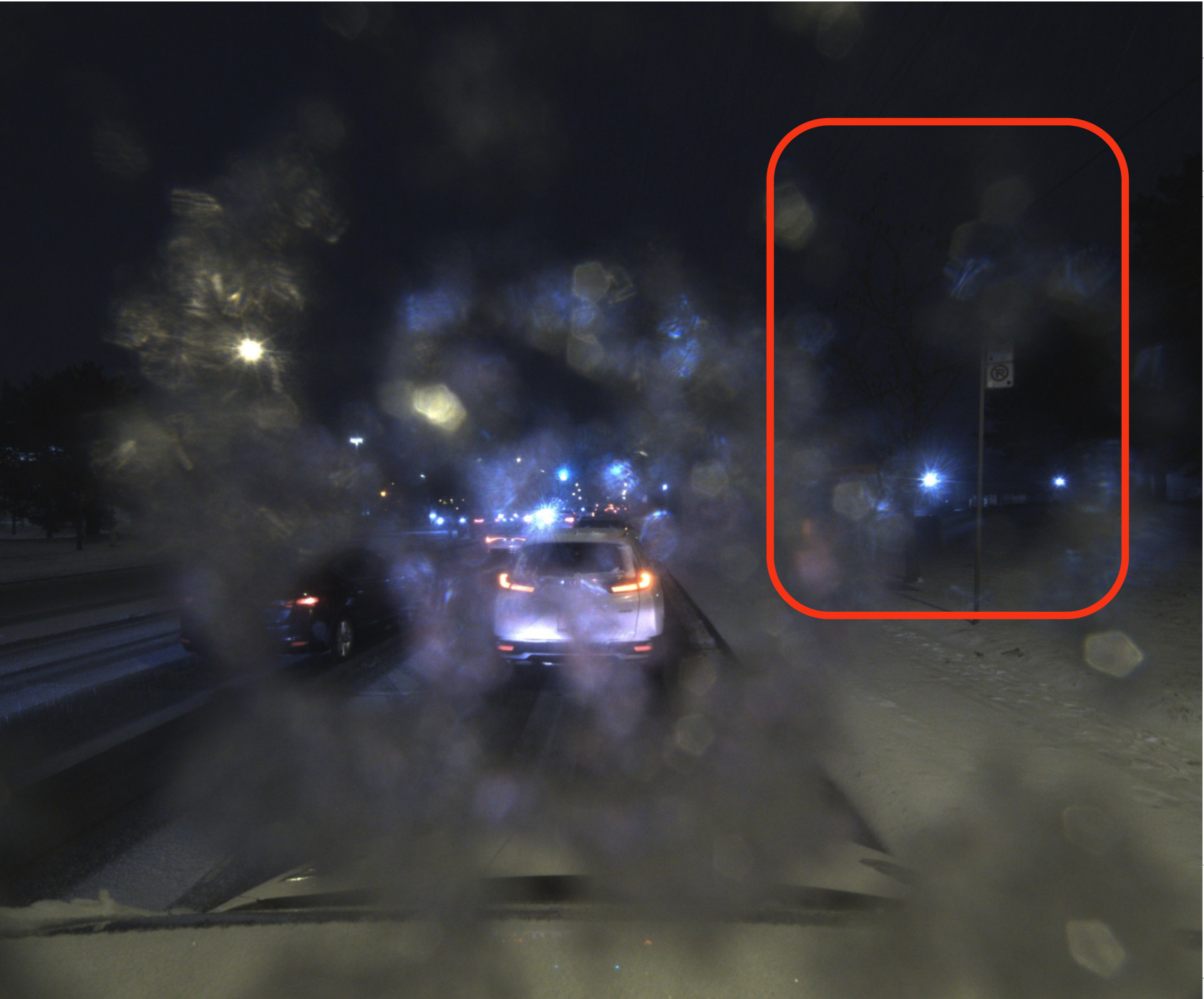} &
    \includegraphics[width=.19\textwidth,height=.115\textwidth]{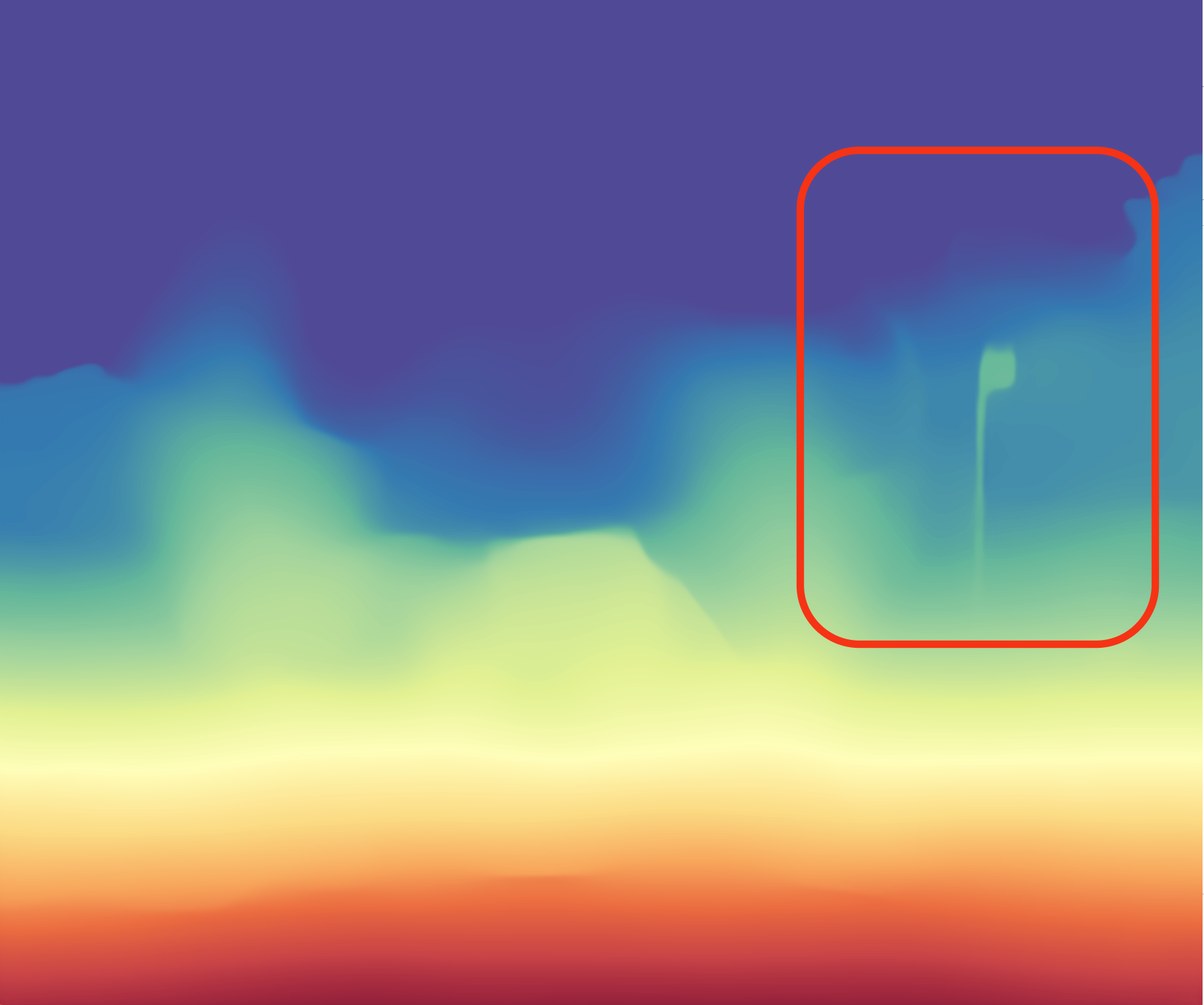} &
    \includegraphics[width=.19\textwidth,height=.115\textwidth]{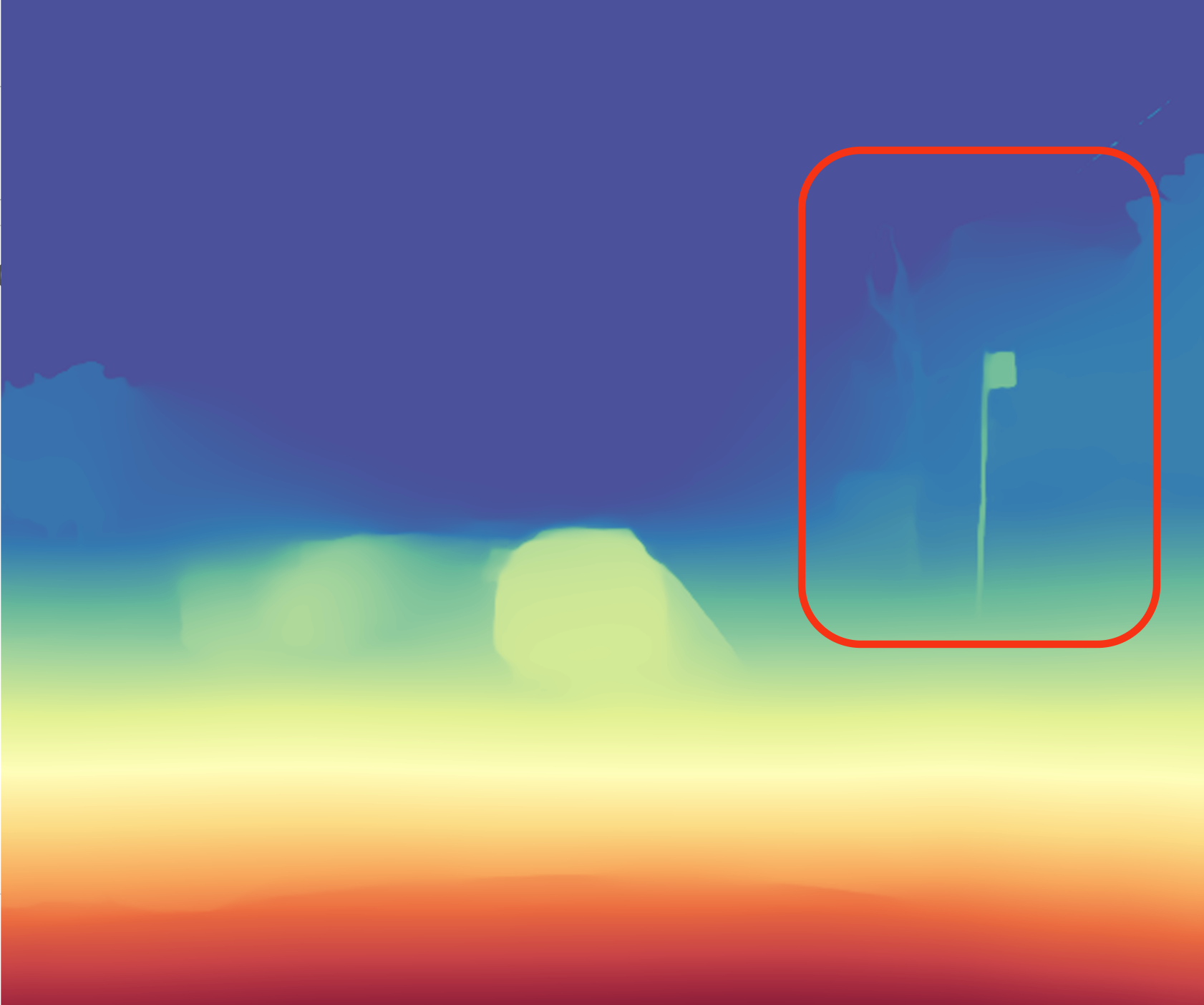} &
    \includegraphics[width=.19\textwidth,height=.115\textwidth]{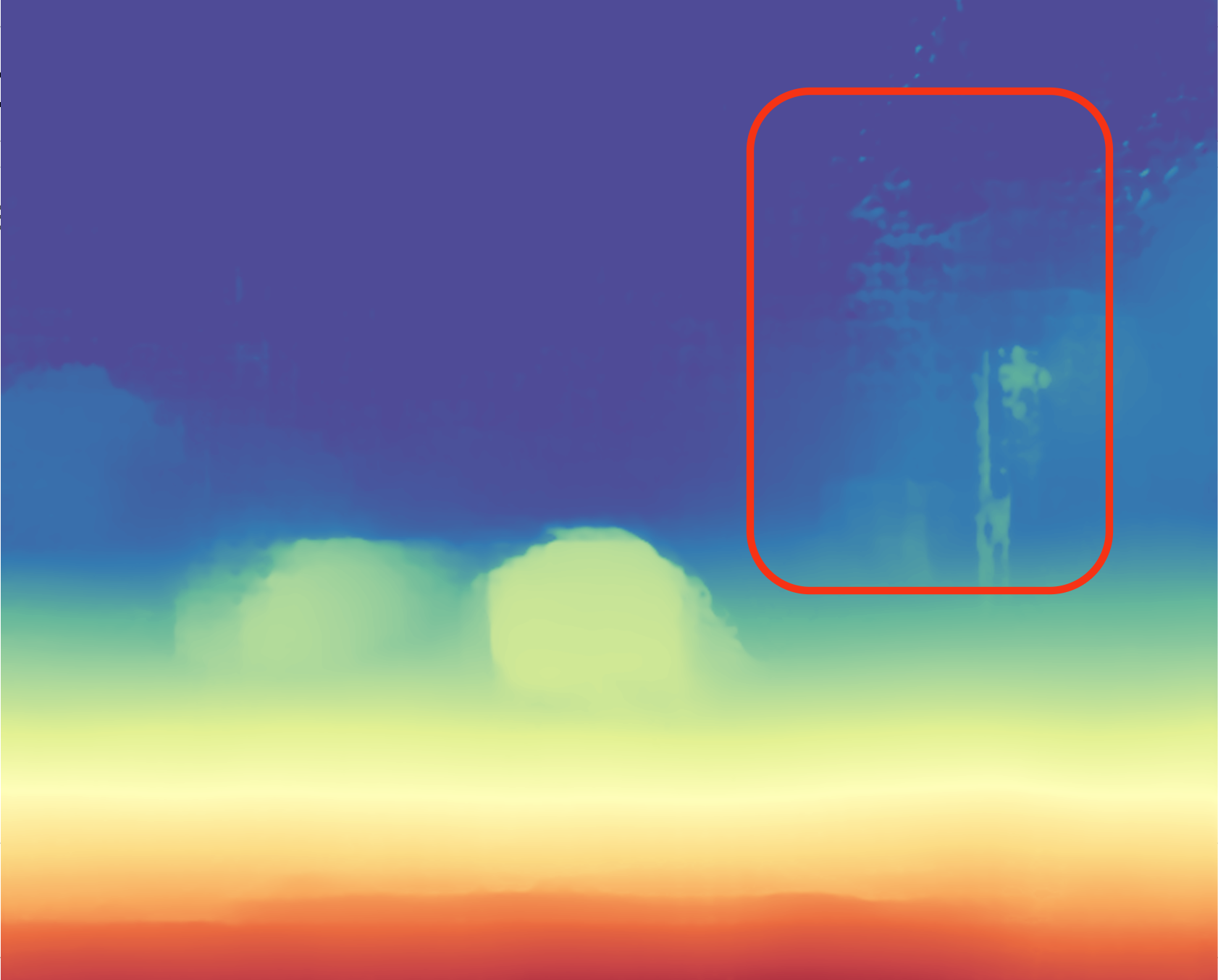} &
    \includegraphics[width=.19\textwidth,height=.115\textwidth]{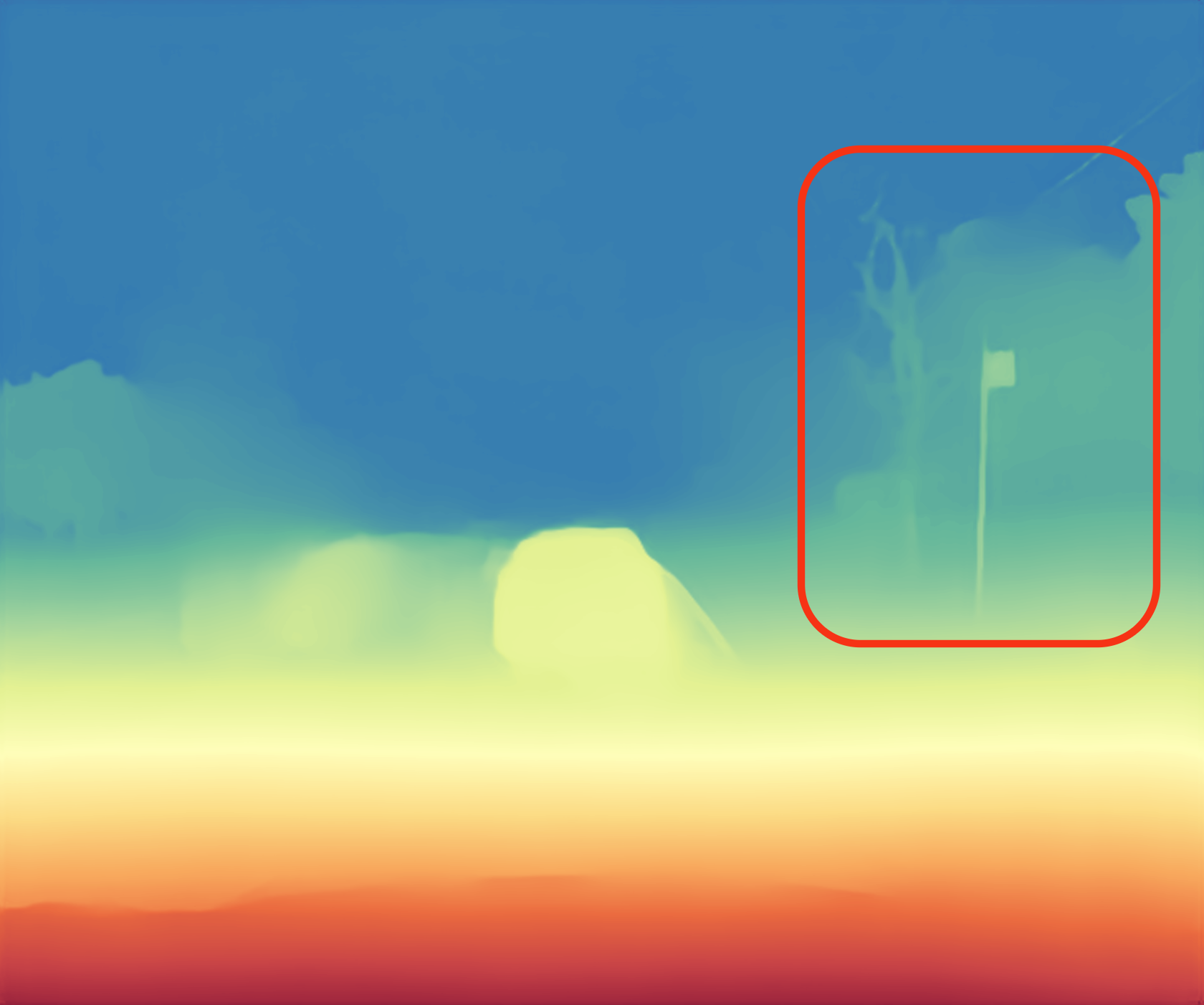} \\

    \raisebox{3em}{\rotatebox[origin=c]{90}{\textbf{}}} &
    \includegraphics[width=.19\textwidth,height=.115\textwidth]{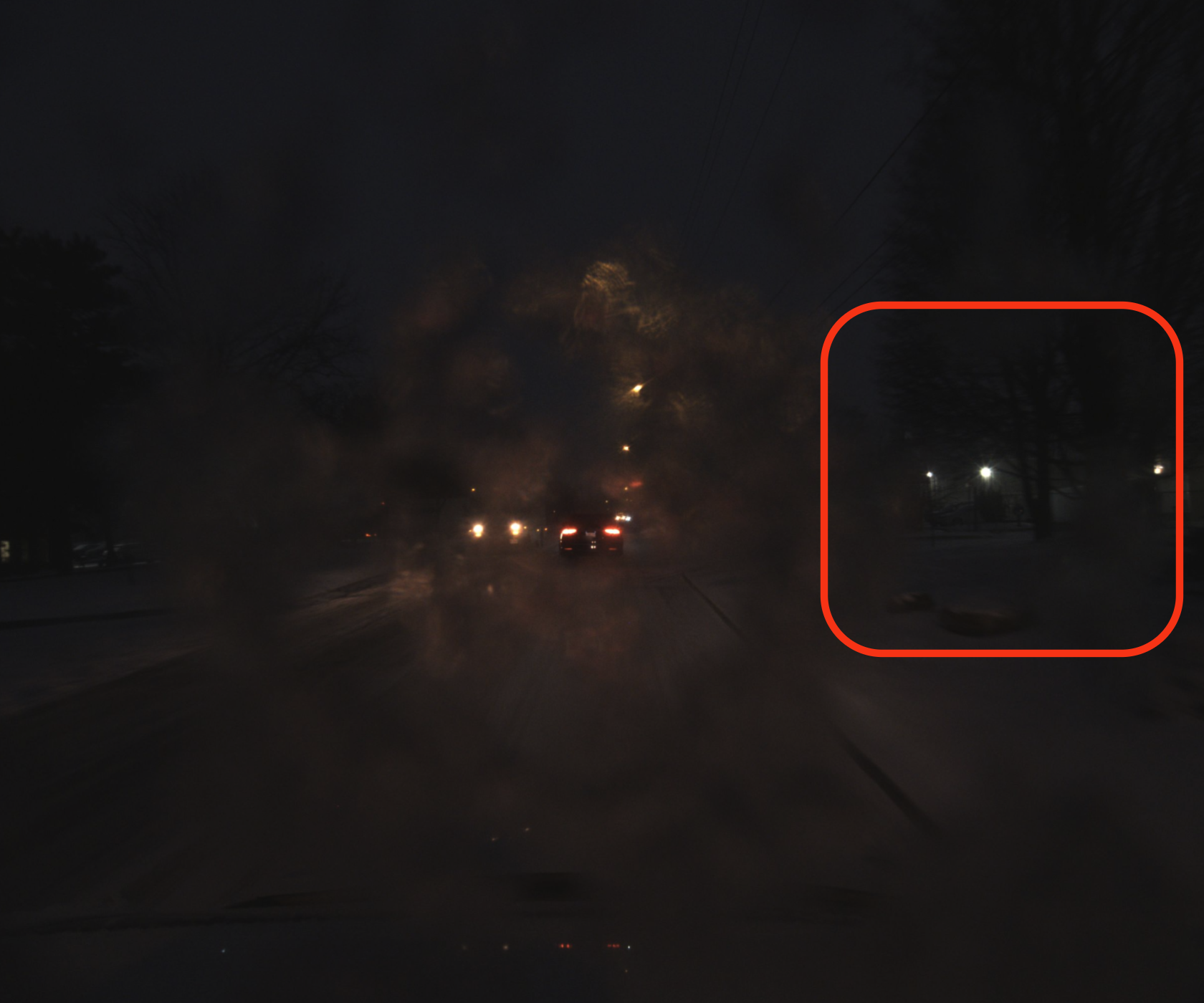} &
    \includegraphics[width=.19\textwidth,height=.115\textwidth]{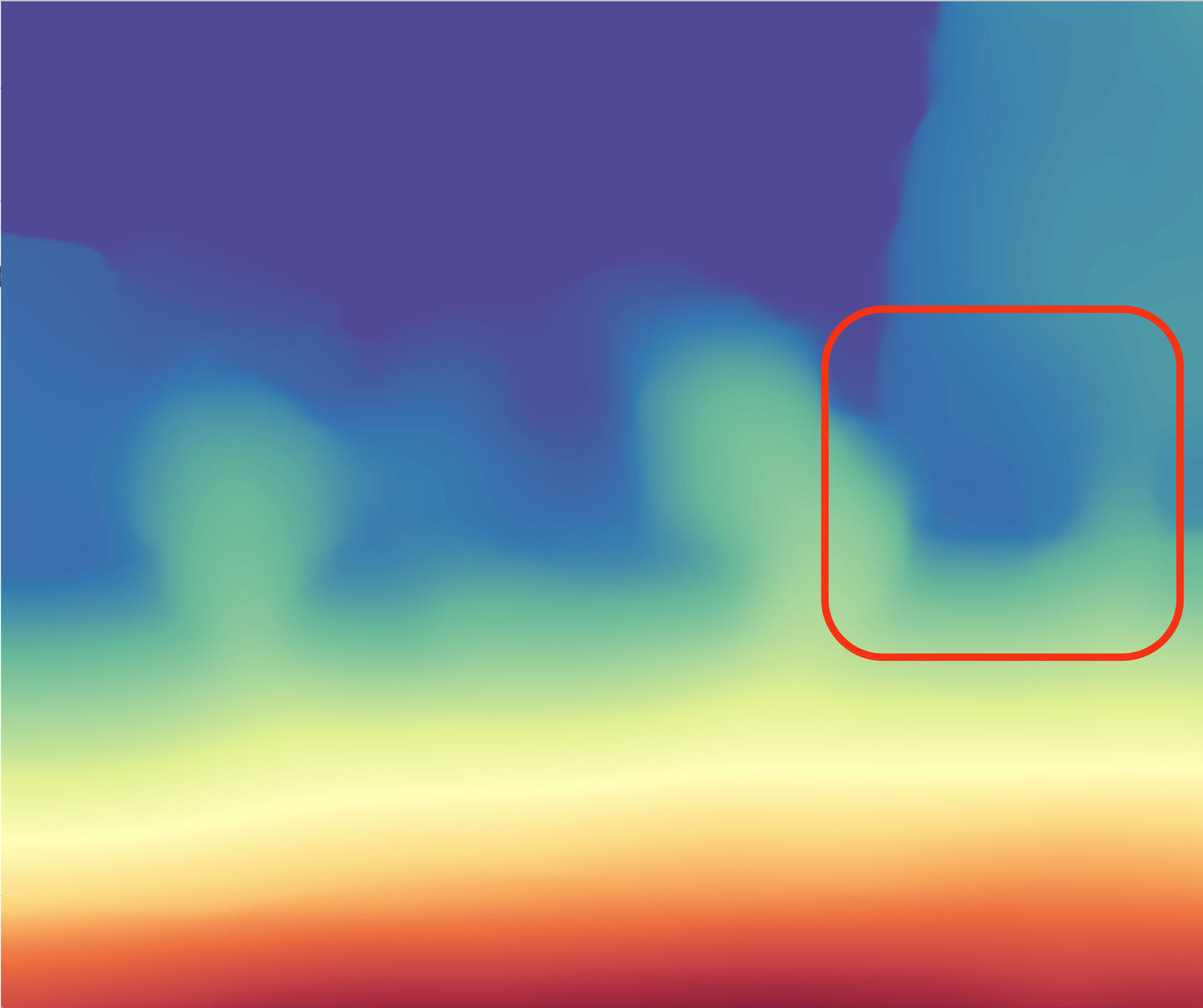} &
    \includegraphics[width=.19\textwidth,height=.115\textwidth]{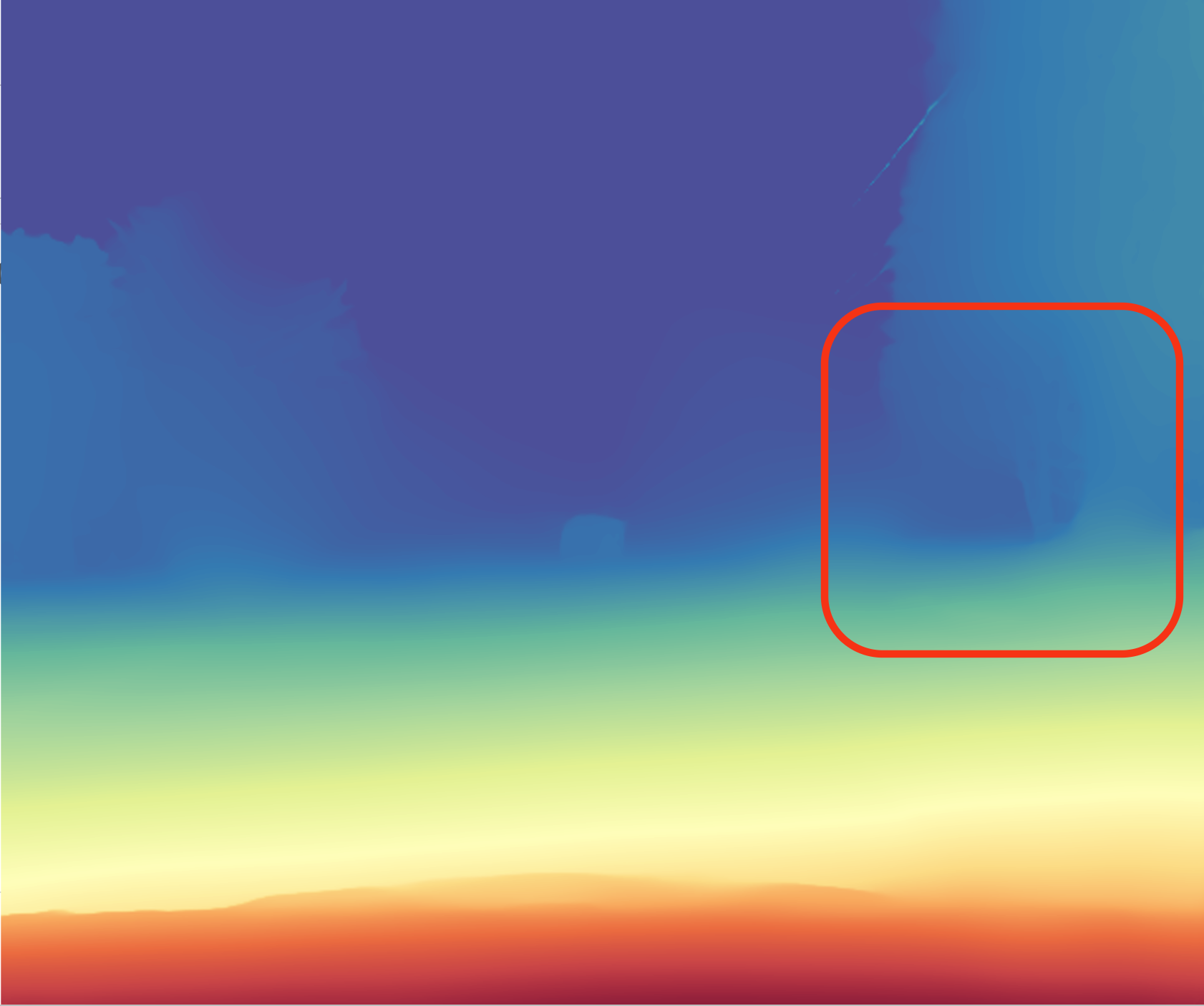} &
    \includegraphics[width=.19\textwidth,height=.115\textwidth]{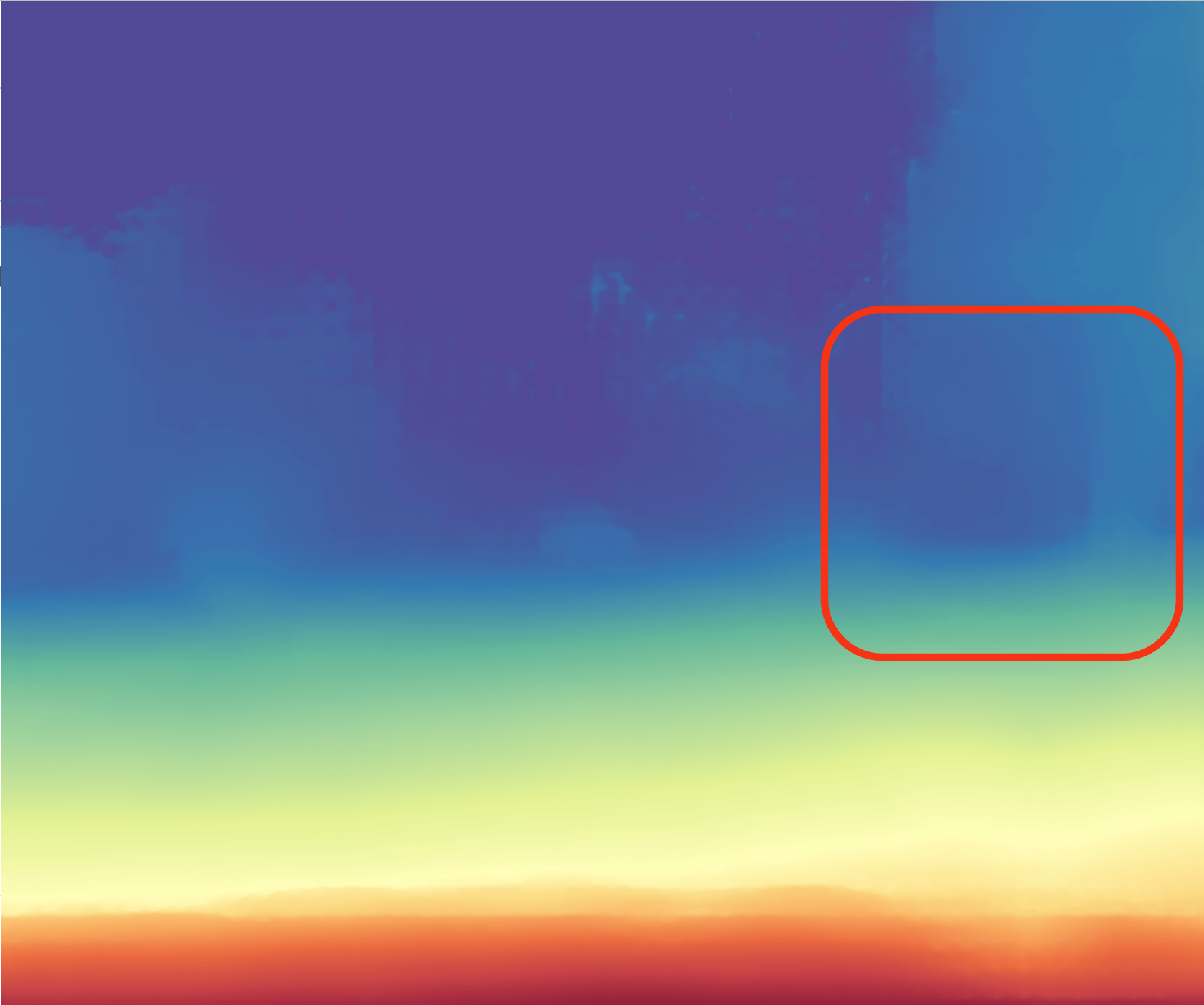} &
    \includegraphics[width=.19\textwidth,height=.115\textwidth]{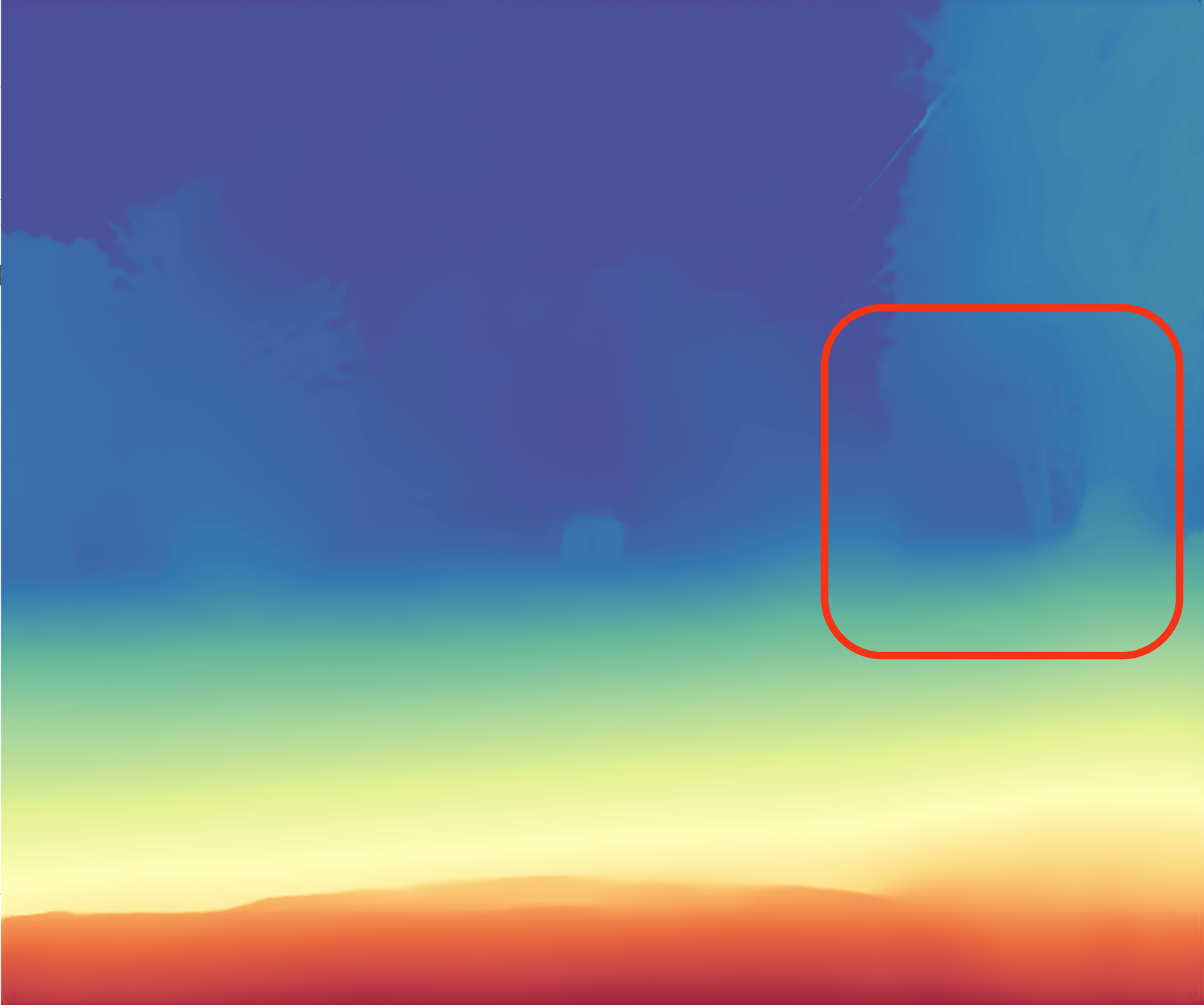} \\
  \end{tabular}
  }
  \vspace{-4mm}
  \caption{\textbf{Qualitative comparison on mixed degradations.}
  Each row shows real-world images of snow at night.}
  \label{fig:qualitative_mix}
\end{figure*}

\begin{table}[t]
\caption{\textbf{Controlled loss ablations. Results are AbsRel$\downarrow$.} All rows use the same 20\% clean/synthetic subset, full real-weather data, seed, and evaluation protocol unless noted. Real night averages RC-N and NS-N; DrivingStereo averages DS-Rain and DS-Fog; KITTI-C averages Fog, Snow, and Dark. Global is the pooled per-image mean over evaluated weather sets.}
\label{tab:loss_ablation}
\centering
\scriptsize
\setlength{\tabcolsep}{3.2pt}
\renewcommand{\arraystretch}{0.90}
\resizebox{\textwidth}{!}{%
\begin{tabular}{@{}lcccc@{}}
\toprule
Variant & Real night & DrivingStereo & KITTI-C & Global \\
\midrule
Full objective & 0.217 & 0.108 & 0.110 & 0.117 \\
w/o $L_{\rm dis}$ & 0.535 & 0.409 & 0.337 & 0.352 \\
w/o $L_{\rm pair}$ & 0.222 & 0.109 & 0.114 & 0.120 \\
w/o $L_{\rm aug}$ & 0.215 & 0.109 & 0.109 & 0.116 \\
$L_{\rm dis}$ only & 0.223 & 0.110 & 0.111 & 0.118 \\
RGB restore ($\lambda_{\rm rgb}=0.1$) & 0.214 & 0.106 & 0.111 & 0.118 \\
Full FT + Real Deg. + Weather Inject. & 0.209 & 0.108 & 0.113 & 0.119 \\
\bottomrule
\end{tabular}%
}
\end{table}

\subsection{Analysis and Discussion}
\begin{figure*}[t]
    \centering
    {%
    \setlength{\tabcolsep}{0pt}
    \begin{tabular}{@{}>{\centering\arraybackslash}p{.25\textwidth}
                    >{\centering\arraybackslash}p{.25\textwidth}
                    >{\centering\arraybackslash}p{.25\textwidth}
                    >{\centering\arraybackslash}p{.25\textwidth}@{}}
        {\scriptsize Input} &
        {\scriptsize Feat. (w/o Cond.)} &
        {\scriptsize Feat. (w/ Cond.)} &
        {\scriptsize $|\Delta$ Feat.$|$} \\
    \end{tabular}
    }

    \vspace{-8pt}

    \includegraphics[height=2cm,width=\textwidth]{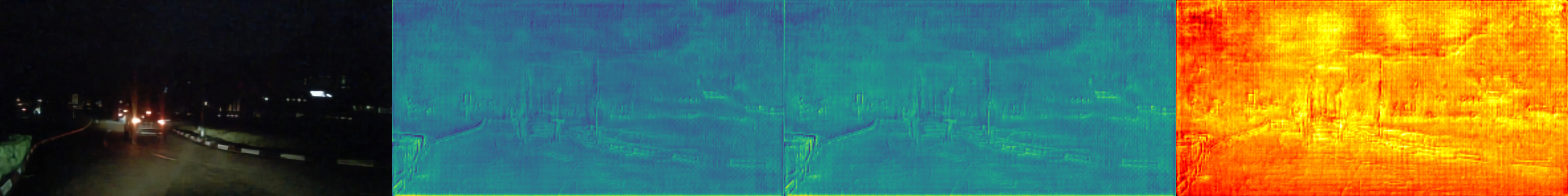}\\
    \includegraphics[height=2cm,width=\textwidth]{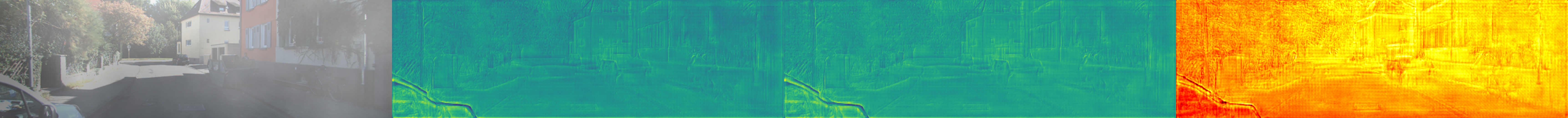}

    \caption{\textbf{Visualization of decoder feature modulation by weather-aware conditioning.}
    From left to right: input image, decoder features before modulation, decoder features after modulation, and the absolute difference map.}
    \label{fig:feature_map}
\end{figure*}

\paragraph{Feature modulation.}
To validate the effectiveness of our weather conditioning, we analyze the decoder features before and after AdaLN modulation. Figure~\ref{fig:feature_map} visualizes the intermediate features for nighttime and foggy scenarios with different maps.
Specifically, the spatial distribution of differences correlates with degradation characteristics: in nighttime scenes, darker regions exhibit markedly stronger modulation than illuminated areas, whereas in foggy scenes, texture-rich regions on the right side show more pronounced differences than the homogeneous regions on the left. This spatially varying behavior indicates that our weather-aware conditioning adaptively focuses on regions most affected by degradation, without requiring explicit spatial supervision. The feature-level differences propagate to depth predictions, confirming that our learned weather representations tangibly impact the final outputs.

\paragraph{Limitations.}
Our controlled adaptation experiments focus on DA-v2-S. Extending the same weather-conditioned interface to larger Depth Anything variants and other depth foundation models requires matched adaptation heads and optimization settings. Real nighttime benchmarks such as RobotCar also include camera response, exposure dynamics, motion blur, and scene-distribution shifts beyond weather appearance alone, which remain challenging for the current weather-conditioned modulation.

\section{Conclusion}

This paper introduces Weather-Conditioned Depth Anything (\textbf{DepthAnything-W}), a model designed for robust monocular depth estimation across diverse adverse weather conditions. The model's effectiveness stems from its key components, notably the \textit{Style Filter}, which learns domain-aligned weather embeddings, and the \textit{AdaLN-Zero}-based decoder, which injects these embeddings into a frozen Depth Anything backbone. To evaluate the proposed architecture, a mixed real-and-synthetic weather dataset is curated, and training incorporates distillation, pairwise alignment, and augmentation consistency to address the synthetic-to-real domain gap while maintaining generalization to clean domains. Extensive experiments on both real and synthetic weather benchmarks demonstrate that our method consistently outperforms strong depth foundation baselines and restoration-based pipelines, without sacrificing performance on standard clean datasets. Its performance sets a new standard for weather robustness in zero-shot depth estimation, offering a promising direction for future research on condition-aware depth foundation models.

\clearpage
\appendix
% \section*{Supplementary Material}
\input{supp_content}
\par\vfill\par
% Now we have reached the maximum length of an ECCV \ECCVyear{} submission (excluding references and acknowledgements).
% References should start immediately after the main text, but can continue past p.\ 14 if needed. 
% \clearpage  % TODO FINAL: This \clearpage needs to be removed from both review and camera-ready versions.

\section*{Acknowledgements}
This work was supported in part by the GPU hardware provided to the Texas A\&M University through the NVIDIA Academia Grant Program.

\bibliographystyle{splncs04}
\bibliography{main}
\end{document}

%% file: supp_content.tex
\section{Implementation Details}
\subsection{Implementation of \textit{\textbf{Style Filter}} Network}

Weather degradations can be conceptualized as weather-specific image styles that are inherently decoupled from scene content and domain. To operationalize this intuition, we adopt a \textit{Style Filter} Network that learns weather-discriminative embeddings via Gram matrix statistics and contrastive learning. The \textit{Style Filter} employs a two-scale patch-based transformer encoder with embedding dimensions [64, 128], attention heads [1, 2], MLP ratios [2, 2], and depths [2, 2], utilizing 4×4 patch tokenization and stochastic depth regularization (drop path rate 0.1). At each encoder scale, we compute the channel Gram matrix $G = FF^T$ over flattened feature maps to capture second-order correlations, which characterize appearance statistics independent of spatial layout. Exploiting the symmetric property of Gram matrices, we extract only the upper-triangular elements, yielding 2,080-D and 8,256-D style descriptors for scales 1 and 2, respectively. These compact representations are processed through dedicated refinement convolutional and residual modules, which transform the raw statistics into weather-specific features, each producing a 64-D intermediate embedding. The two scale-specific embeddings are concatenated and projected via a linear layer with LayerNorm to obtain the final 64-D weather embedding $\mathbf{w}$. The entire Style Filter is trained using a contrastive loss that clusters images by weather type while maintaining invariance to scene content and cross-domain (real-synthetic) alignment, thereby enabling robust weather conditioning for the downstream depth estimation model.

\begin{figure*}[t]
    \centering
    \includegraphics[width=\linewidth]{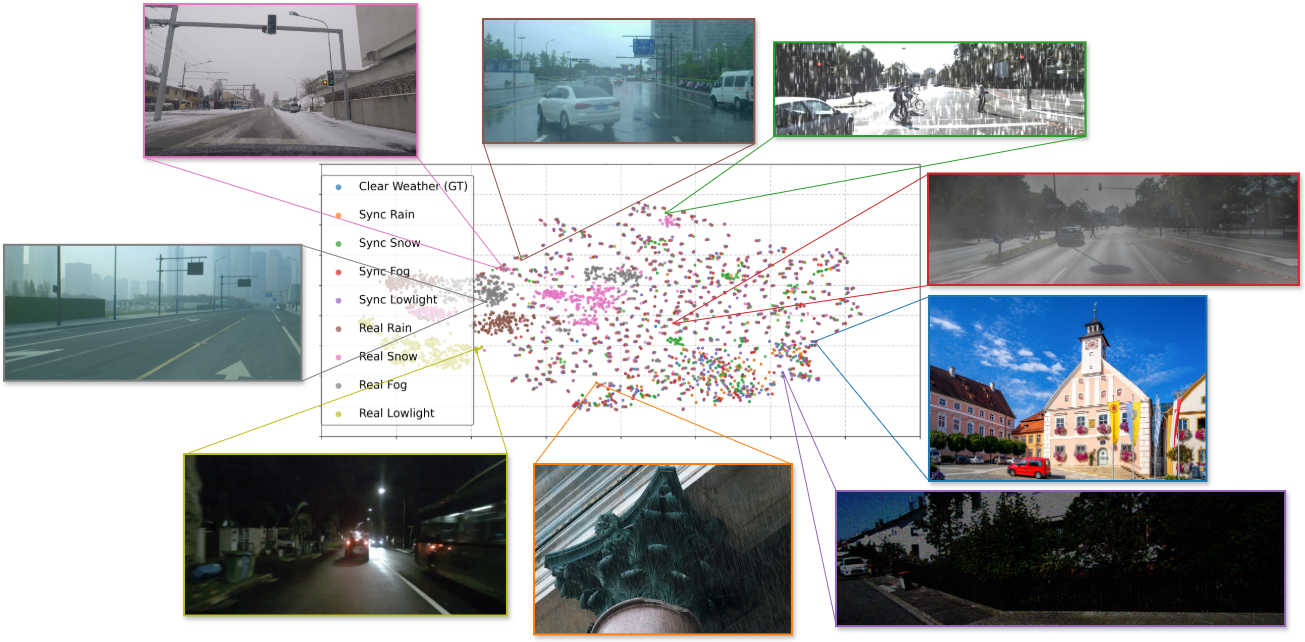} \\
    \caption{\textbf{t-SNE Visualization of real and synthetic dataset distributions.}}
    \label{fig:supp_vis}
\end{figure*}

To train our \textit{Style Filter} Network, we curated 7 real-world adverse weather degradation datasets and 4 synthetic weather datasets with corresponding synthetic weather degradations. Since the clean datasets (COCO~\cite{lin2014microsoft}, MegaDepth~\cite{li2018megadepth}, SA-1B~\cite{kirillov2023segment}, and HRWSI~\cite{xian2020structure}) have been extensively documented in prior Depth Anything works~\cite{depthanything, yang2024depth, depthanything3}, we focus our description on the real-world adverse weather datasets that are critical for learning domain-aligned weather embeddings. Figure~\ref{fig:supp_vis} visualizes the feature distributions of real and synthetic datasets under the pre-trained Depth Anything v2~\cite{yang2024depth} backbone.

\subsection{Dataset Curation}

\paragraph{ACDC Dataset.}
The Adverse Conditions Dataset with Correspondences (ACDC) ~\cite{sakaridis2025acdc} is a large-scale benchmark for semantic perception in driving scenes under challenging weather and illumination conditions. It comprises 8,012 images, of which 4,006 depict adverse scenarios such as fog, nighttime, rain, and snow. For each adverse-condition image, a corresponding image of the same scene captured under normal conditions is included, facilitating direct comparative analysis. All images are collected across multiple European cities using high-resolution RGB sensors and are accompanied by pixel-level panoptic semantic annotations for training and evaluation of robust perception systems. Following the dataset's official protocol, we use only images from the designated training split to train our \textit{Style Filter} Network.
\paragraph{Muses Dataset.}
 The Multi-Sensor Semantic Perception Dataset for Driving Under Uncertainty (MUSES)~\cite{brodermann2024muses} is a large-scale autonomous driving dataset containing 2,500 annotated scenes captured across diverse European locations under varying weather and illumination conditions, including fog, rain, snow, nighttime, and clear daytime scenarios. MUSES provides synchronized multimodal sensor recordings and high-quality annotations for multiple perception tasks. Notably, MUSES is the first dataset to provide uncertainty-aware annotations with varying degradation severities across different weather types, capturing the natural variability of real-world adverse conditions. To ensure sufficient coverage of snow degradation, which is relatively underrepresented in other datasets, we use the snow sequences from the MUSES training split to train our \textit{Style Filter} Network.
\paragraph{NightCity Dataset.}
The NightCity dataset~\cite{Tan_2021_TIP_NightCity} is a large-scale nighttime scene parsing dataset specifically designed to address perception challenges in low-light urban environments. The dataset contains over 4,000 high-resolution nighttime images captured across diverse cities under various illumination conditions, ranging from street-lit urban roads to poorly illuminated residential areas.  Unlike daytime datasets or synthetically darkened images, NightCity captures authentic nighttime characteristics including non-uniform illumination from artificial light sources, strong contrast variations, and low signal-to-noise ratios that fundamentally alter scene appearance. The dataset's diverse illumination patterns enable our model to handle varying degrees of low-light severity.
\paragraph{RID and RIS Dataset.}
The RIS (Rain in Surveillance) and RID (Rain in Driving) datasets are two real-world datasets of rainy images~\cite {8954466}. These datasets are part of the larger Multi-Purpose Image Deraining (MPID) benchmark and are specifically designed for task-driven evaluation of image deraining algorithms. The RID dataset comprises 2,496 real rainy images extracted from high-resolution driving videos captured by car-mounted cameras under rainy conditions. The RIS dataset contains 2,048 real rainy images extracted from 154 surveillance cameras under daytime rainy conditions.
\paragraph{RTTS Dataset.}
The Real-world Task-driven Testing Set (RTTS)~\cite{8451944} is a real-world dehazing benchmark comprising 4,322 hazy images crawled from the web, primarily covering outdoor traffic and driving scenes. The images exhibit varying fog and haze densities, with authentic atmospheric scattering effects and depth-dependent visibility degradation that differ substantially from those in synthetic fog simulations. The dataset includes object detection annotations with bounding boxes for five traffic object categories (car, person, bus, truck, and motorcycle), enabling task-driven evaluation of dehazing algorithms.
\paragraph{Snow100K Dataset.}
The Snow100K dataset~\cite{liu2018desnownet} is a large-scale snow removal benchmark comprising 100,000 synthetic snowy images paired with clean ground truth, along with 1,329 realistic snowy photographs captured under real-world conditions. The synthetic subset is organized into three categories (Snow100K-S, Snow100K-M, and Snow100K-L) based on snow particle size, thereby simulating diverse snowfall conditions. However, to maintain consistency with our goal of learning from authentic weather degradations, we exclusively use the 1,329 realistic snowy photographs from Snow100K to train our \textit{Style Filter} Network, excluding the synthetic samples.

\section{More Qualitative Results}
\subsection{Qualitative results under different degradation}

\begin{figure*}[t!]
  \centering
  \setlength{\tabcolsep}{1pt} 
  \resizebox{\textwidth}{!}{
  \begin{tabular}{ccccc}
    \textbf{Input} & \textbf{DA v1} & \textbf{DA v2} & \textbf{DA-AC} & \textbf{DA-W(ours)}\\

    \includegraphics[width=.20\textwidth, height=2.1cm]{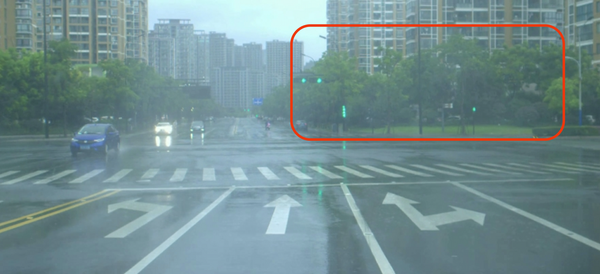} & 
    \includegraphics[width=.20\textwidth, height=2.1cm]{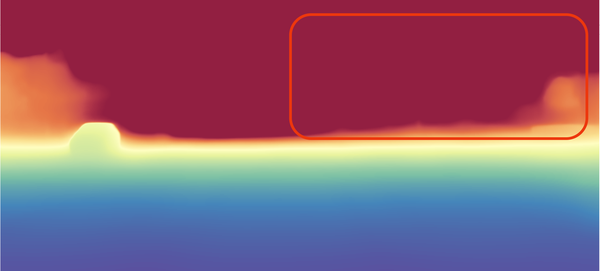} & 
    \includegraphics[width=.20\textwidth, height=2.1cm]{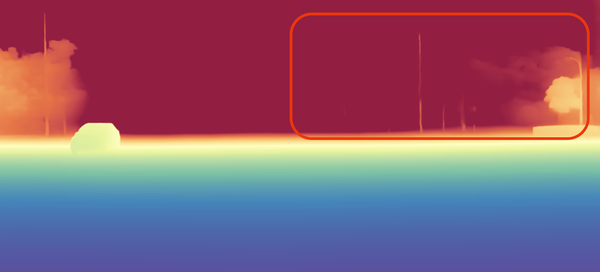} & 
    \includegraphics[width=.20\textwidth, height=2.1cm]{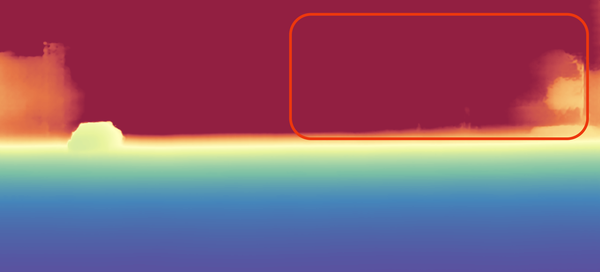} &
    \includegraphics[width=.20\textwidth, height=2.1cm]{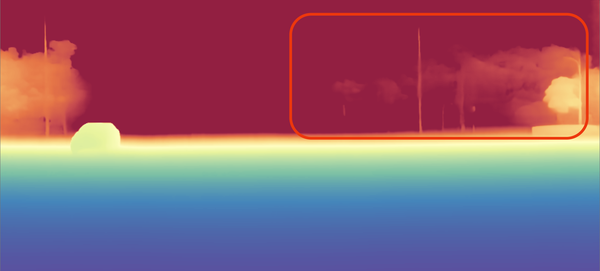} \\
        \multicolumn{5}{c}{\vspace{-14pt}} \\
    
    \includegraphics[width=.20\textwidth, height=2.1cm]{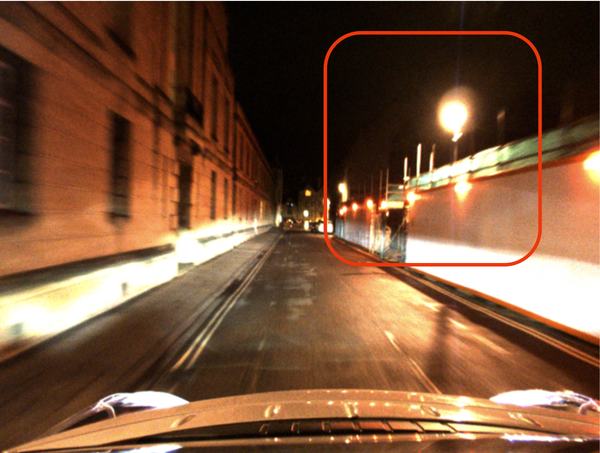} &
    \includegraphics[width=.20\textwidth, height=2.1cm]{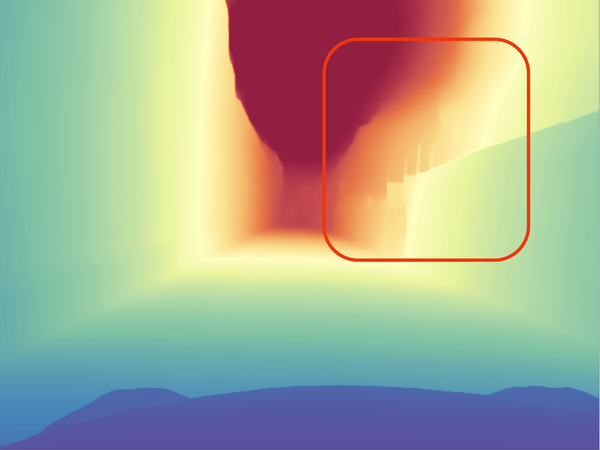} &
    \includegraphics[width=.20\textwidth, height=2.1cm]{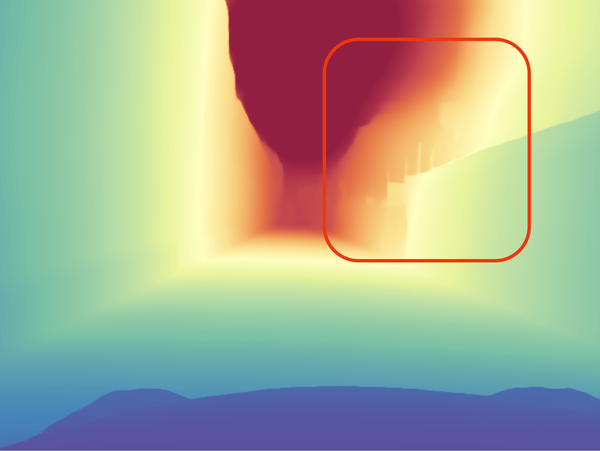} & 
    \includegraphics[width=.20\textwidth, height=2.1cm]{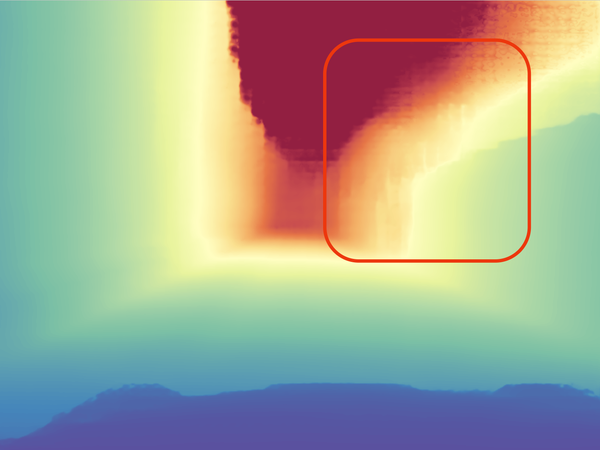} &
    \includegraphics[width=.20\textwidth, height=2.1cm]{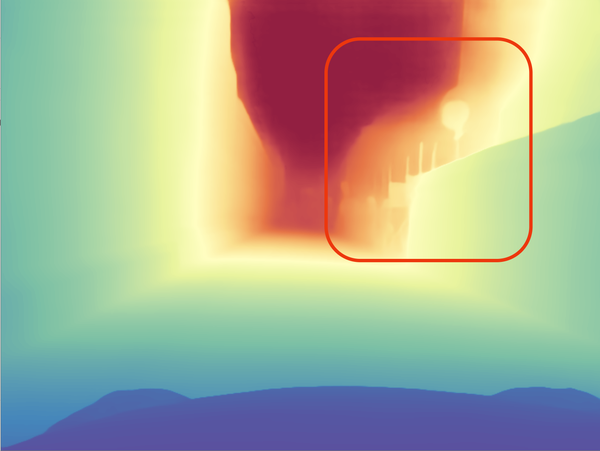} \\
        \multicolumn{5}{c}{\vspace{-14pt}} \\

    \includegraphics[width=.20\textwidth, height=2.1cm]{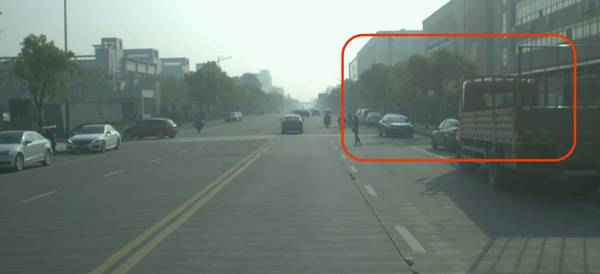} &
    \includegraphics[width=.20\textwidth, height=2.1cm]{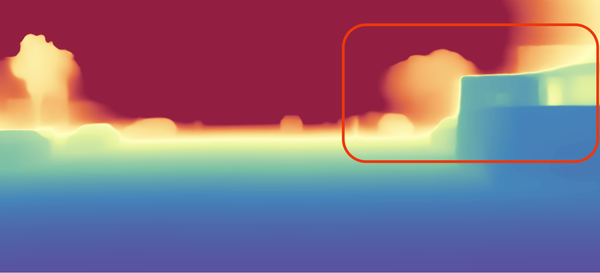} &
    \includegraphics[width=.20\textwidth, height=2.1cm]{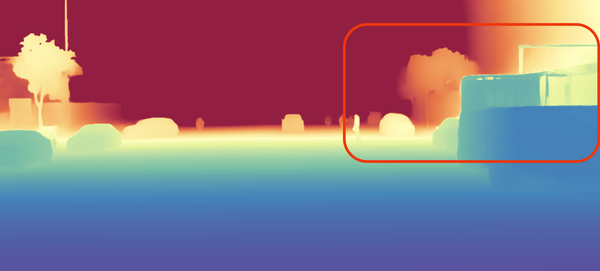} & 
    \includegraphics[width=.20\textwidth, height=2.1cm]{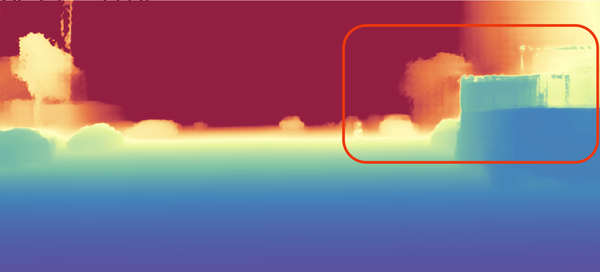} &
    \includegraphics[width=.20\textwidth, height=2.1cm]{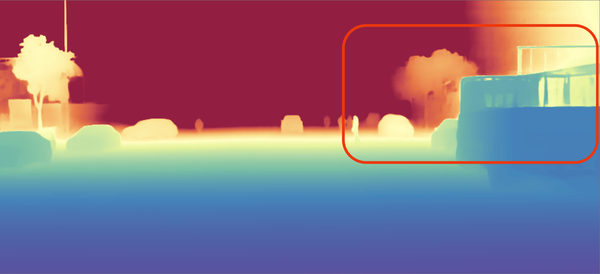} \\    
        \multicolumn{5}{c}{\vspace{-14pt}} \\
    
    \includegraphics[width=.20\textwidth, height=2.1cm]{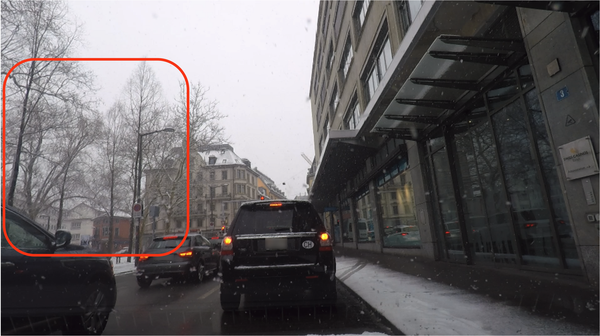} &
    \includegraphics[width=.20\textwidth, height=2.1cm]{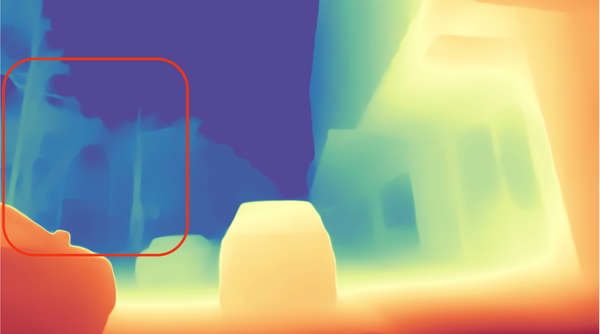} & 
    \includegraphics[width=.20\textwidth, height=2.1cm]{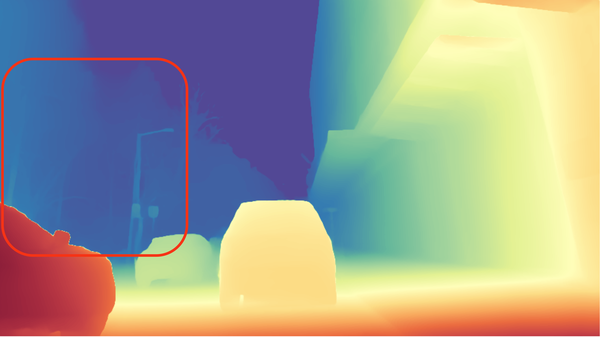} &
    \includegraphics[width=.20\textwidth, height=2.1cm]{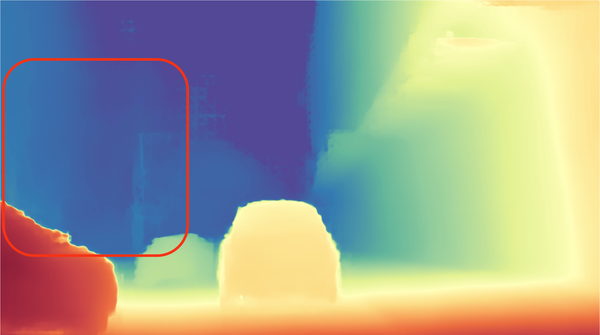} & 
    \includegraphics[width=.20\textwidth, height=2.1cm]{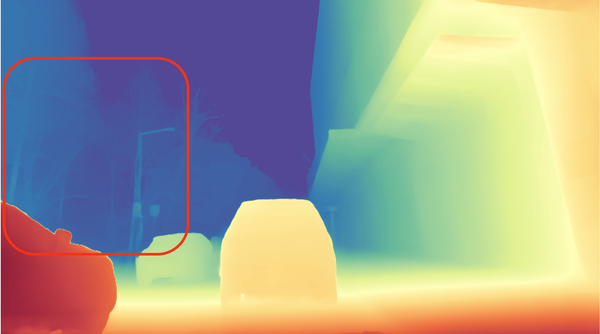} \\
        \multicolumn{5}{c}{\vspace{-14pt}} \\

    \includegraphics[width=.20\textwidth, height=2.1cm]{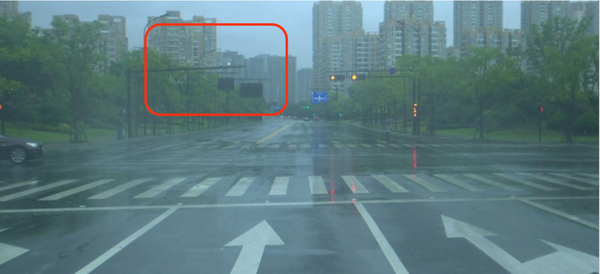} &
    \includegraphics[width=.20\textwidth, height=2.1cm]{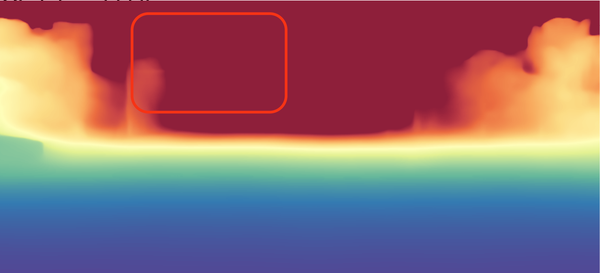} & 
    \includegraphics[width=.20\textwidth, height=2.1cm]{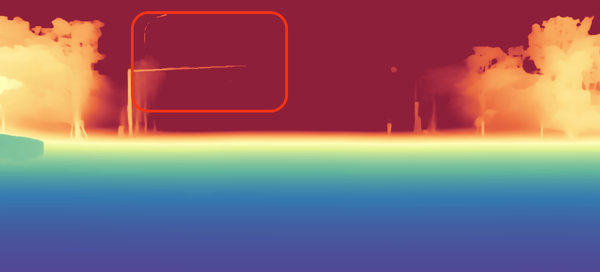} &
    \includegraphics[width=.20\textwidth, height=2.1cm]{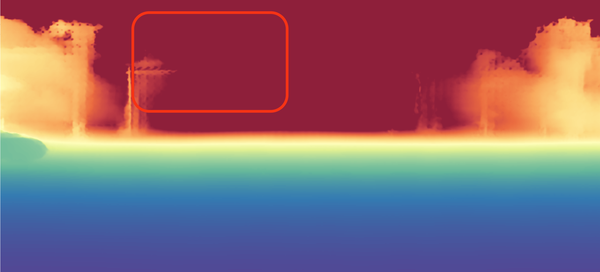} & 
    \includegraphics[width=.20\textwidth, height=2.1cm]{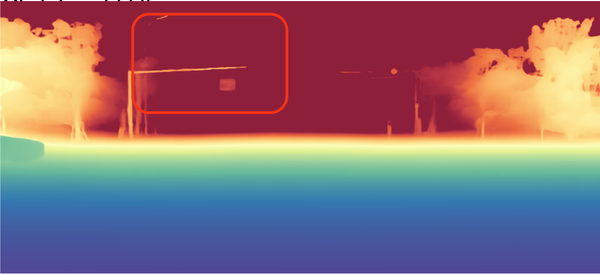} \\
        \multicolumn{5}{c}{\vspace{-14pt}} \\

    \includegraphics[width=.20\textwidth, height=2.1cm]{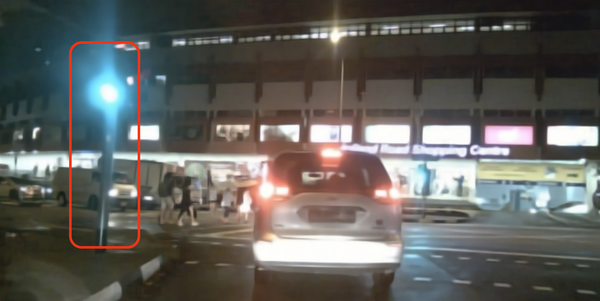} &
    \includegraphics[width=.20\textwidth, height=2.1cm]{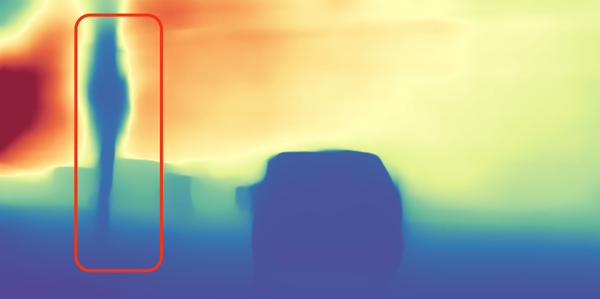} & 
    \includegraphics[width=.20\textwidth, height=2.1cm]{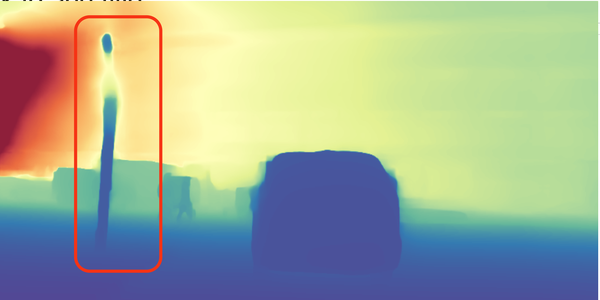} &
    \includegraphics[width=.20\textwidth, height=2.1cm]{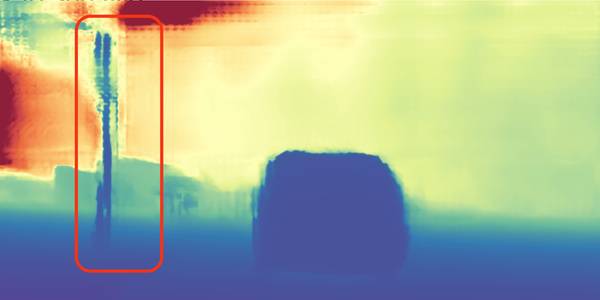} & 
    \includegraphics[width=.20\textwidth, height=2.1cm]{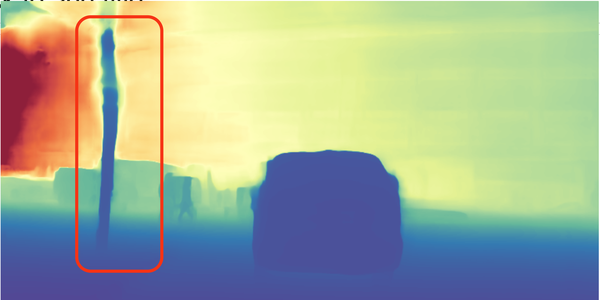} \\
        \multicolumn{5}{c}{\vspace{-14pt}} \\

    \includegraphics[width=.20\textwidth, height=2.1cm]{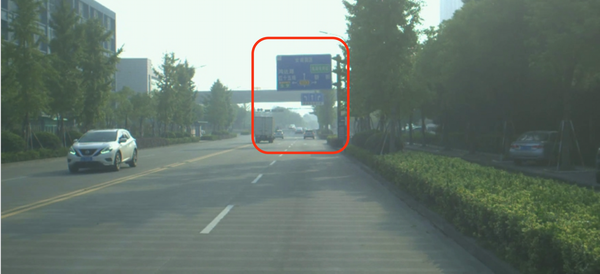} &
    \includegraphics[width=.20\textwidth, height=2.1cm]{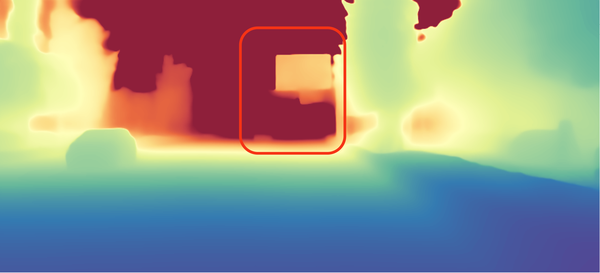} & 
    \includegraphics[width=.20\textwidth, height=2.1cm]{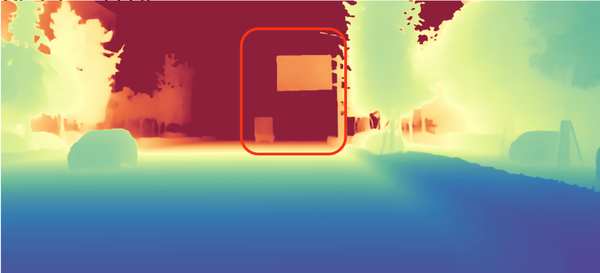} &
    \includegraphics[width=.20\textwidth, height=2.1cm]{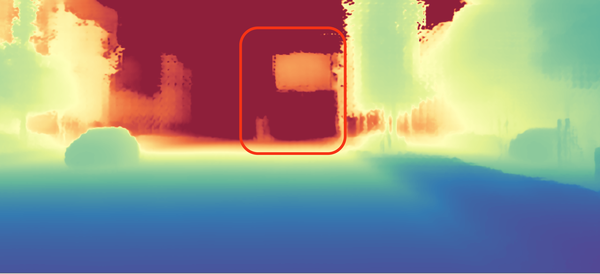} & 
    \includegraphics[width=.20\textwidth, height=2.1cm]{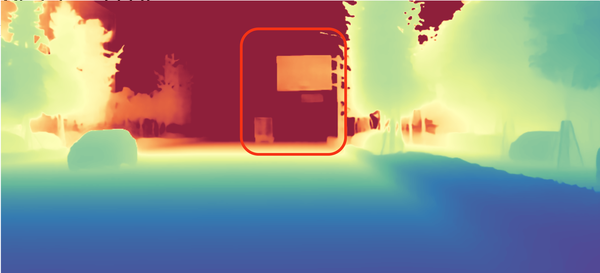} \\
        \multicolumn{5}{c}{\vspace{-14pt}} \\

    \includegraphics[width=.20\textwidth, height=2.1cm]{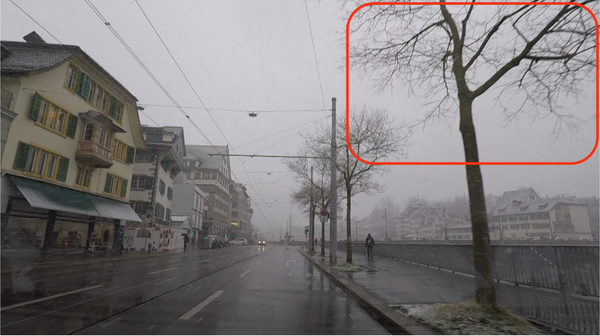} &
    \includegraphics[width=.20\textwidth, height=2.1cm]{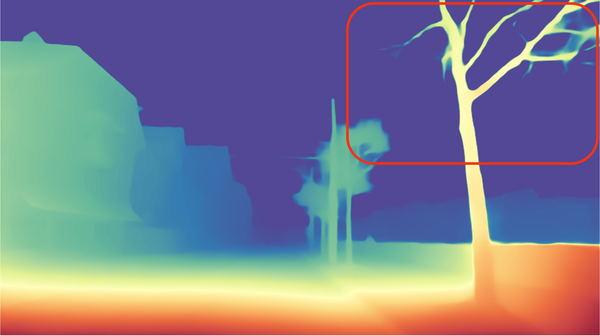} & 
    \includegraphics[width=.20\textwidth, height=2.1cm]{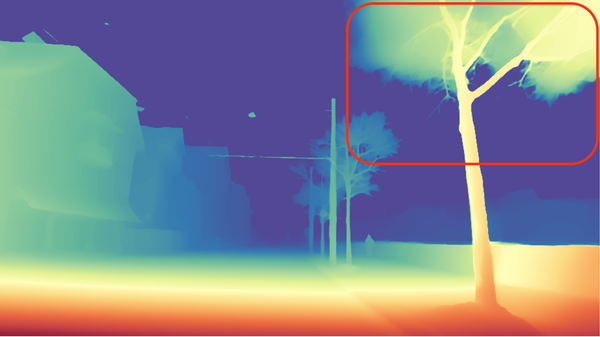} &
    \includegraphics[width=.20\textwidth, height=2.1cm]{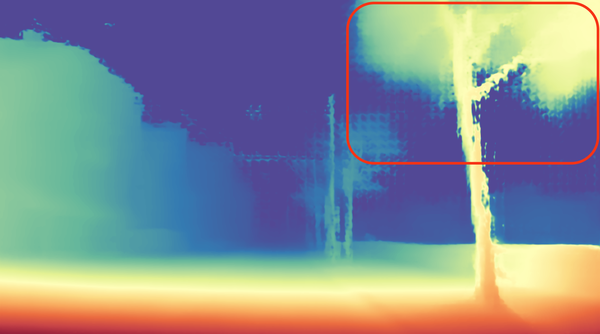} & 
    \includegraphics[width=.20\textwidth, height=2.1cm]{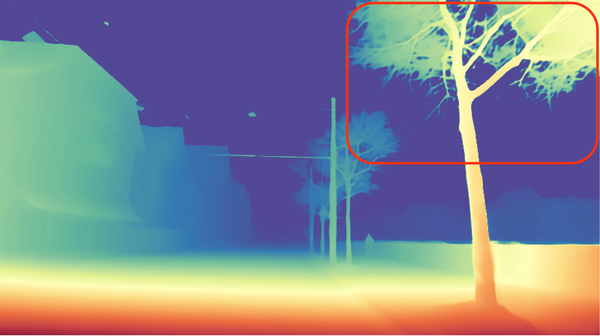} \\
        \multicolumn{5}{c}{\vspace{-14pt}} \\

  \end{tabular}
  }
\caption{\textbf{Qualitative Depth Map Comparison Under Real-World Weather Degradations.}}
  \label{fig:supp_real_qualitative}
\end{figure*}

\begin{figure*}[t!]
  \centering
  \setlength{\tabcolsep}{1pt} 
  \resizebox{\textwidth}{!}{
  \begin{tabular}{ccccc}
    \textbf{Input} & \textbf{DA v1} & \textbf{DA v2} & \textbf{DA-AC} & \textbf{DA-W(ours)}\\

    \includegraphics[width=.20\textwidth, height=1.9cm]{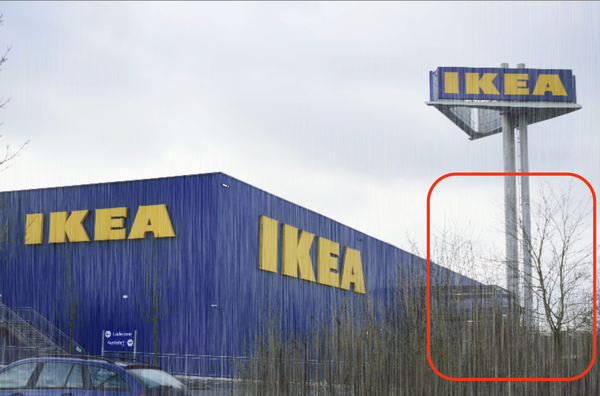} & 
    \includegraphics[width=.20\textwidth, height=1.9cm]{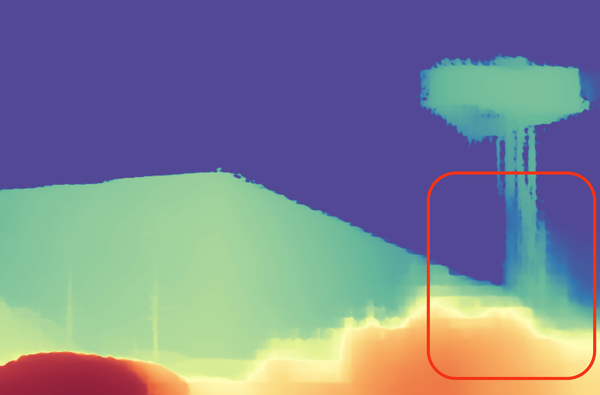} & 
    \includegraphics[width=.20\textwidth, height=1.9cm]{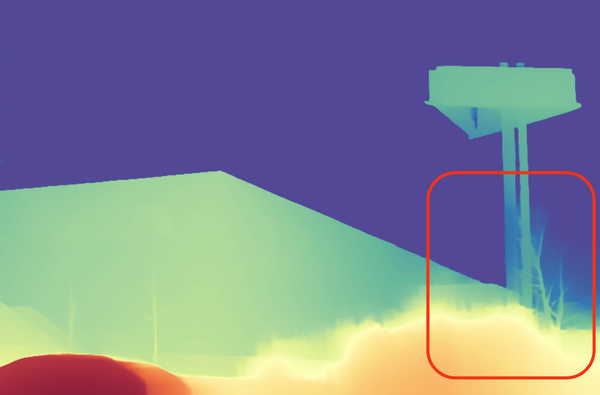} & 
    \includegraphics[width=.20\textwidth, height=1.9cm]{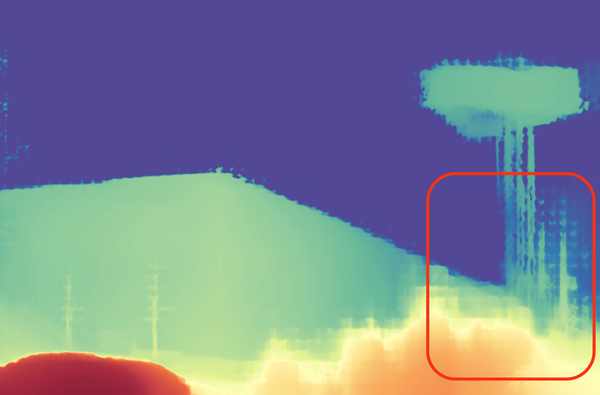} &
    \includegraphics[width=.20\textwidth, height=1.9cm]{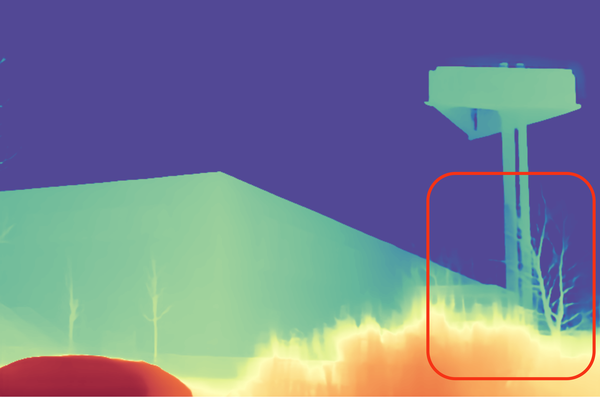} \\
        \multicolumn{5}{c}{\vspace{-14pt}} \\
    
    \includegraphics[width=.20\textwidth, height=1.8cm]{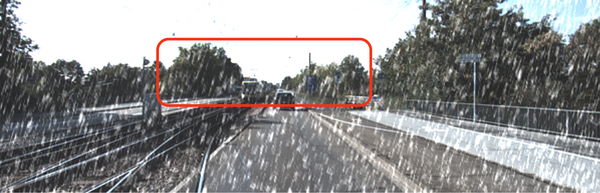} &
    \includegraphics[width=.20\textwidth, height=1.8cm]{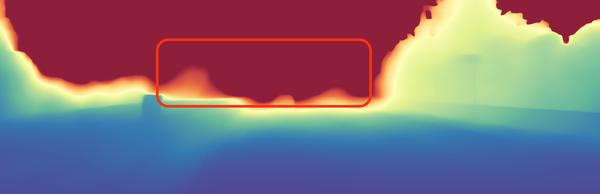} &
    \includegraphics[width=.20\textwidth, height=1.8cm]{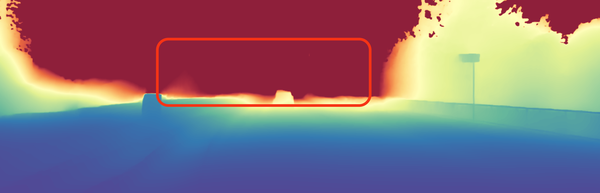} & 
    \includegraphics[width=.20\textwidth, height=1.8cm]{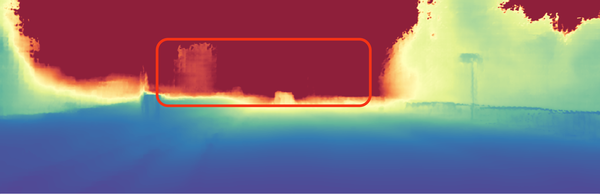} &
    \includegraphics[width=.20\textwidth, height=1.8cm]{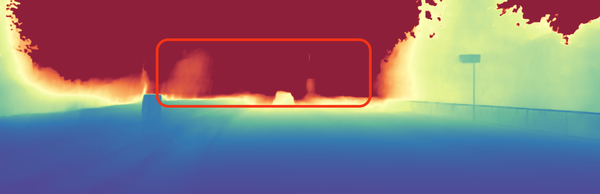} \\
        \multicolumn{5}{c}{\vspace{-14pt}} \\

    \includegraphics[width=.20\textwidth, height=1.8cm]{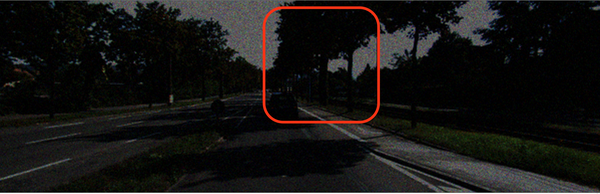} &
    \includegraphics[width=.20\textwidth, height=1.8cm]{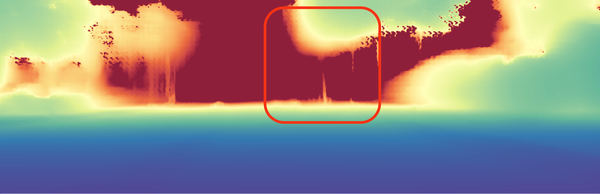} &
    \includegraphics[width=.20\textwidth, height=1.8cm]{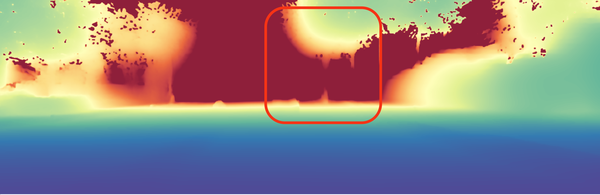} & 
    \includegraphics[width=.20\textwidth, height=1.8cm]{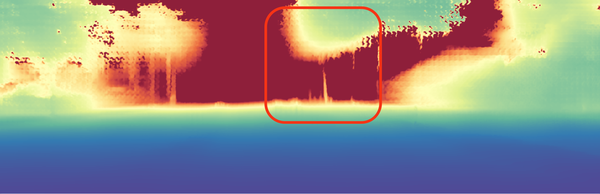} &
    \includegraphics[width=.20\textwidth, height=1.8cm]{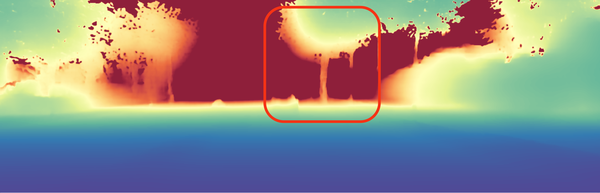} \\    
        \multicolumn{5}{c}{\vspace{-14pt}} \\
    
    \includegraphics[width=.20\textwidth, height=1.8cm]{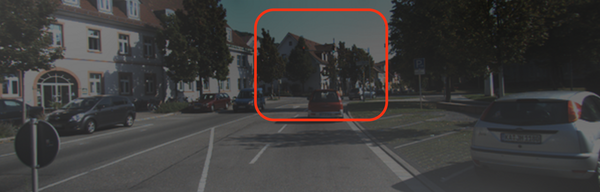} &
    \includegraphics[width=.20\textwidth, height=1.8cm]{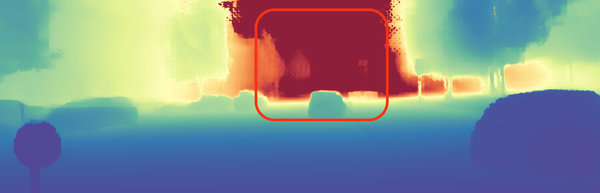} & 
    \includegraphics[width=.20\textwidth, height=1.8cm]{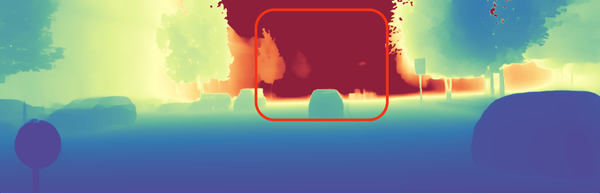} &
    \includegraphics[width=.20\textwidth, height=1.8cm]{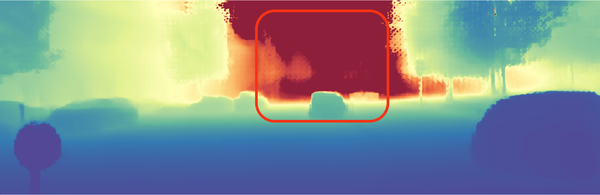} & 
    \includegraphics[width=.20\textwidth, height=1.8cm]{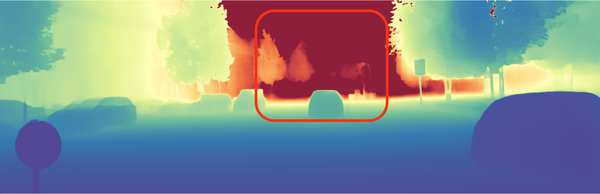} \\
        \multicolumn{5}{c}{\vspace{-14pt}} \\

    \includegraphics[width=.20\textwidth, height=1.8cm]{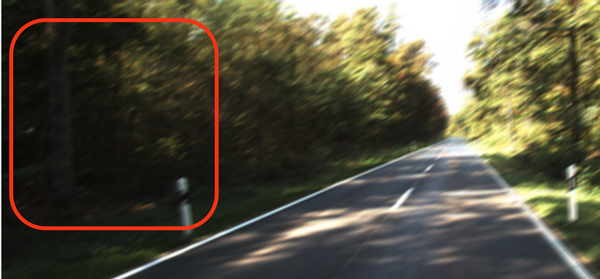} &
    \includegraphics[width=.20\textwidth, height=1.8cm]{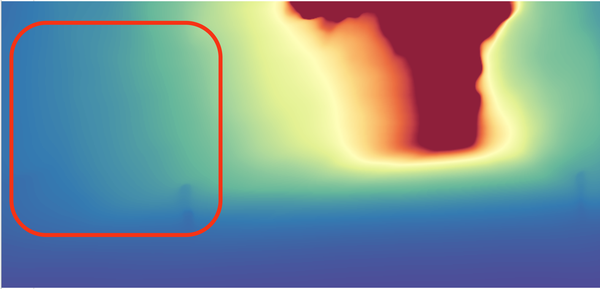} & 
    \includegraphics[width=.20\textwidth, height=1.8cm]{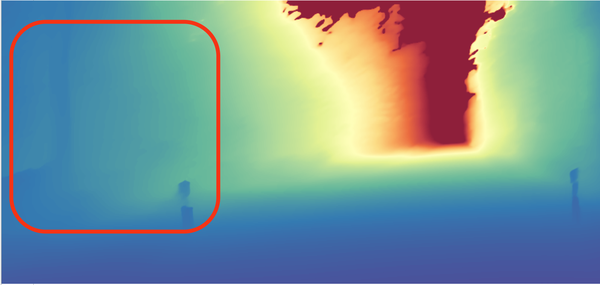} &
    \includegraphics[width=.20\textwidth, height=1.8cm]{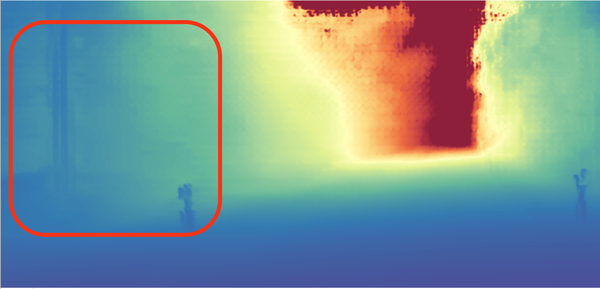} & 
    \includegraphics[width=.20\textwidth, height=1.8cm]{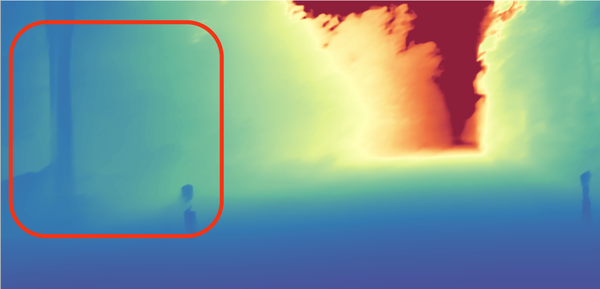} \\
        \multicolumn{5}{c}{\vspace{-14pt}} \\

    \includegraphics[width=.20\textwidth, height=2cm]{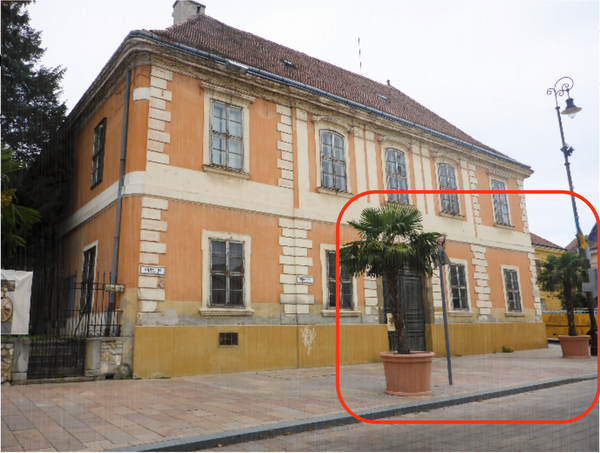} &
    \includegraphics[width=.20\textwidth, height=2cm]{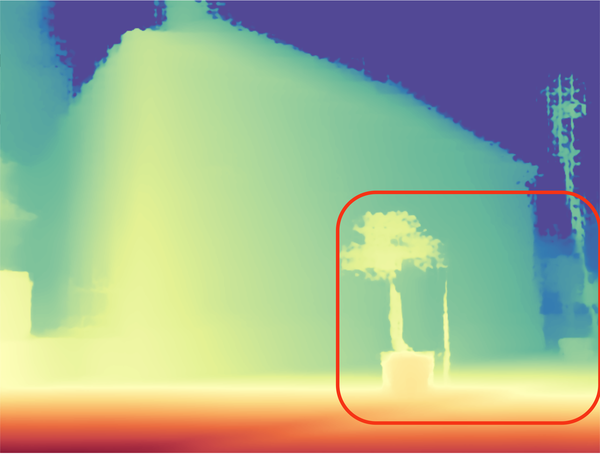} &
    \includegraphics[width=.20\textwidth, height=2cm]{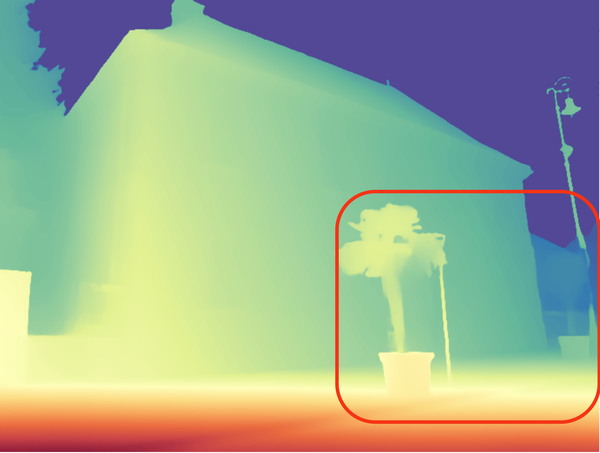} & 
    \includegraphics[width=.20\textwidth, height=2cm]{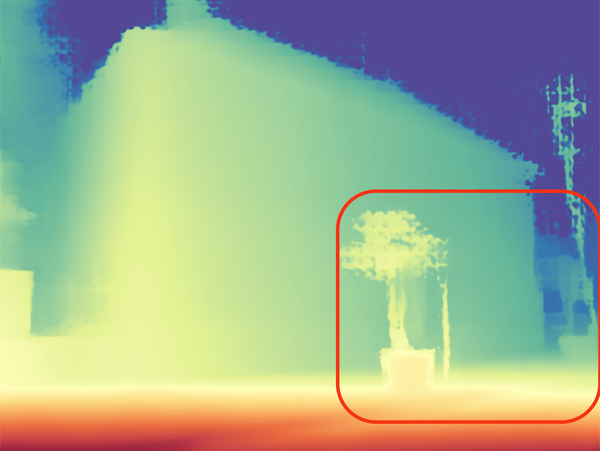} &
    \includegraphics[width=.20\textwidth, height=2cm]{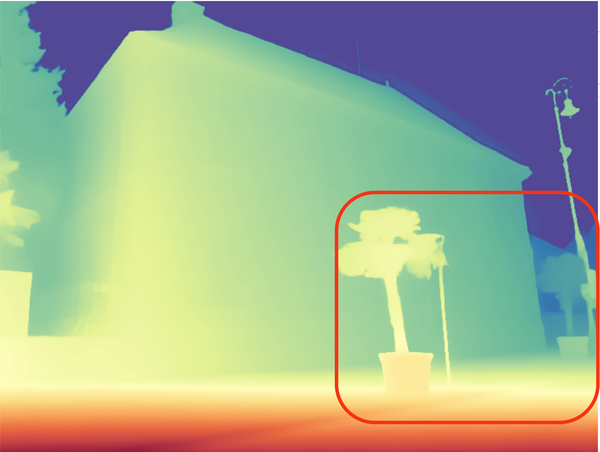} \\ 
        \multicolumn{5}{c}{\vspace{-14pt}} \\
        
    \includegraphics[width=.20\textwidth, height=2.cm]{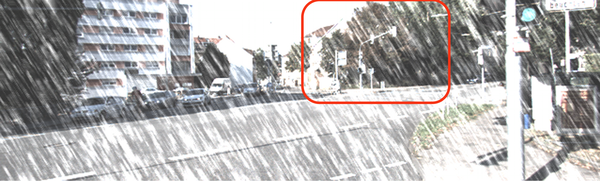} &
    \includegraphics[width=.20\textwidth, height=2.cm]{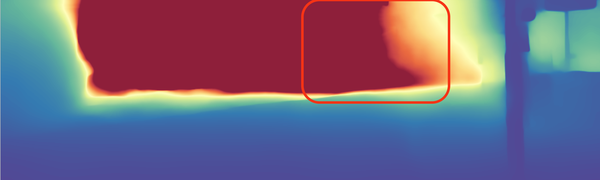} &
    \includegraphics[width=.20\textwidth, height=2.cm]{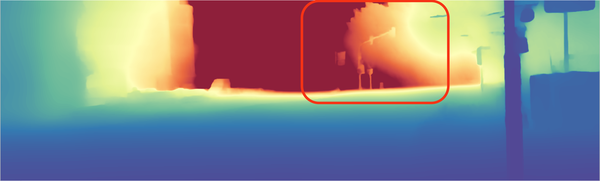} & 
    \includegraphics[width=.20\textwidth, height=2.cm]{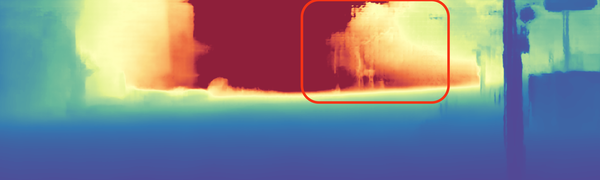} &
    \includegraphics[width=.20\textwidth, height=2.cm]{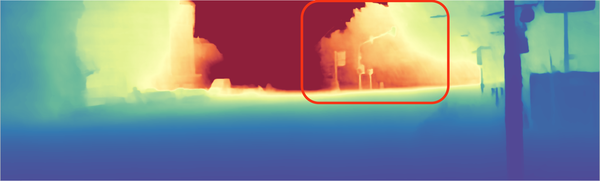} \\    
        \multicolumn{5}{c}{\vspace{-14pt}} \\

    \includegraphics[width=.20\textwidth, height=1.9cm]{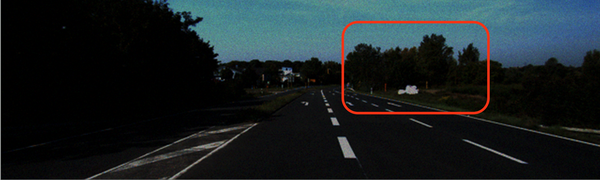} &
    \includegraphics[width=.20\textwidth, height=1.9cm]{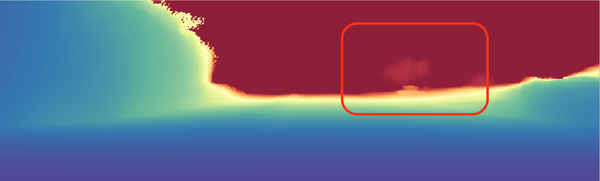} &
    \includegraphics[width=.20\textwidth, height=1.9cm]{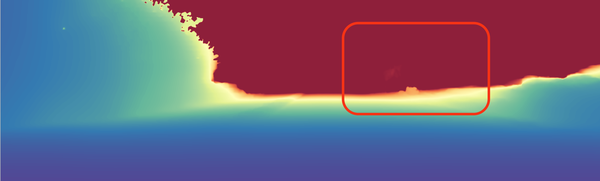} & 
    \includegraphics[width=.20\textwidth, height=1.9cm]{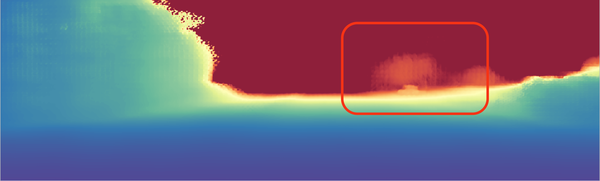} &
    \includegraphics[width=.20\textwidth, height=1.9cm]{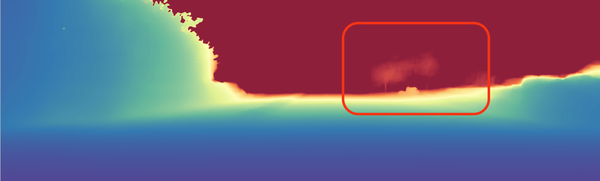} \\     
        \multicolumn{5}{c}{\vspace{-14pt}} \\

    \includegraphics[width=.20\textwidth, height=1.9cm]{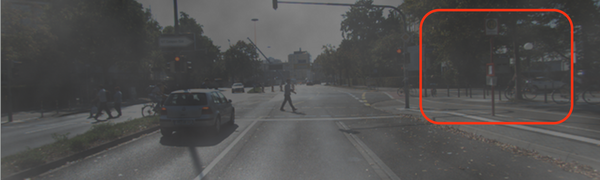} &
    \includegraphics[width=.20\textwidth, height=1.9cm]{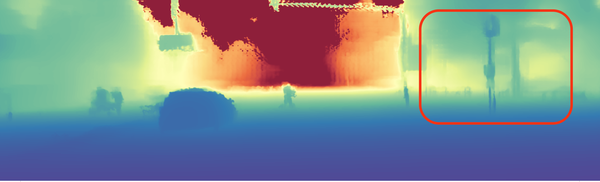} &
    \includegraphics[width=.20\textwidth, height=1.9cm]{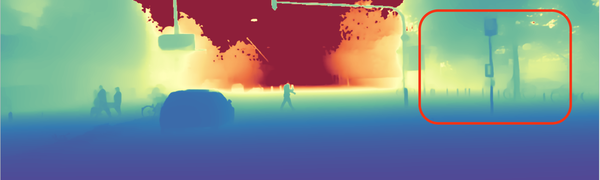} & 
    \includegraphics[width=.20\textwidth, height=1.9cm]{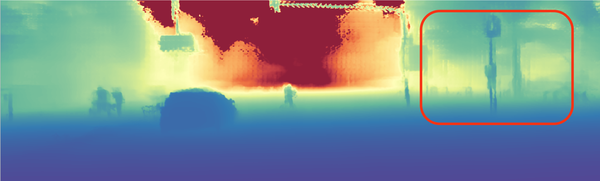} &
    \includegraphics[width=.20\textwidth, height=1.9cm]{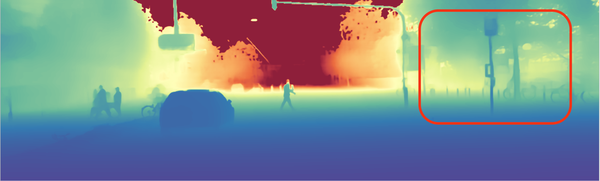} \\    
        \multicolumn{5}{c}{\vspace{-14pt}} \\

    \includegraphics[width=.20\textwidth, height=1.9cm]{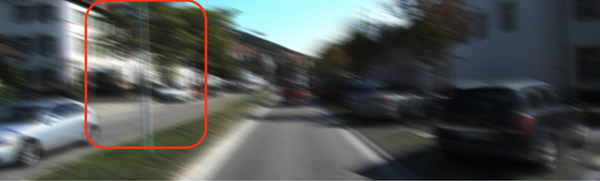} &
    \includegraphics[width=.20\textwidth, height=1.9cm]{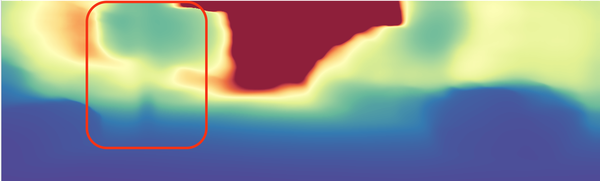} &
    \includegraphics[width=.20\textwidth, height=1.9cm]{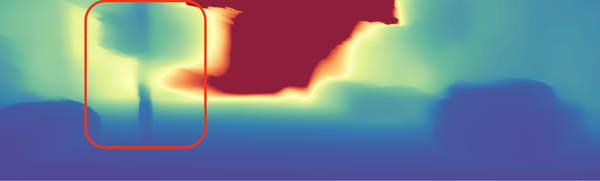} & 
    \includegraphics[width=.20\textwidth, height=1.9cm]{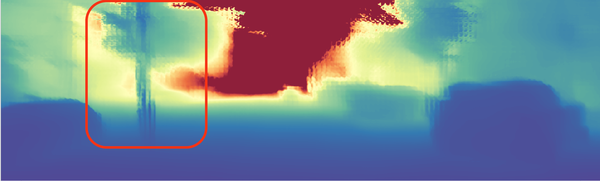} &
    \includegraphics[width=.20\textwidth, height=1.9cm]{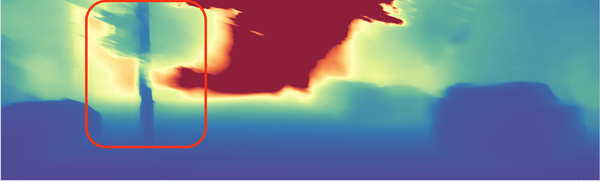} \\   
        \multicolumn{5}{c}{\vspace{-14pt}} \\

  \end{tabular}
  }
\caption{\textbf{Qualitative Depth Map Comparison Under Synthetic Degradations.}}
  \label{fig:supp_sync_qualitative}
\end{figure*}

Figure~\ref{fig:supp_real_qualitative} and Figure~\ref{fig:supp_sync_qualitative} show more qualitative results compared with three state-of-the-art methods: Depth Anything v1~\cite{depthanything}, v2~\cite{yang2024depth} and DepthAnything-AC~\cite{sun2025depth}. The qualitative results suggest that Depth Anything v1 directly loses fine-grained details under extreme weather conditions; Depth Anything v2 achieves better overall performance but still produces discontinuous edges; and the recent DepthAnything-AC model tends to generate diffuse depth boundaries. 

We attribute DepthAnything-AC's diffuse predictions to its physics-based degradation modeling strategy, which applies synthetic weather augmentations uniformly across the entire image during training. While this approach improves robustness, it inherently assumes spatially homogeneous degradation patterns and may over-smooth depth boundaries to maintain consistency under such synthetic corruptions. Moreover, the knowledge distillation from clean-image predictions to degraded inputs encourages the model to produce conservative, smoothed estimates rather than preserving sharp geometric transitions, leading to the observed diffuse boundaries particularly in regions with high-frequency texture or depth discontinuities.

In contrast, Depth Anything v1's severe detail loss under adverse weather stems from its training paradigm, which primarily focuses on maximizing generalization across diverse clean domains without explicit weather-aware conditioning. When confronted with extreme weather degradations—such as dense fog, heavy snow, or low-light conditions—that introduce significant domain shifts from its training distribution, the model's feature extractor fails to capture reliable monocular depth cues. The entangled representation of weather appearance and scene geometry causes the encoder to suppress high-frequency details as noise, resulting in over-smoothed or even collapsed depth predictions in severely degraded regions.

Our proposed DA-W addresses these limitations by explicitly disentangling style and content via the Style Filter and weather-conditioned modulation. By learning domain-aligned weather embeddings that remain invariant to scene content, our model adaptively modulates decoder features based on the estimated degradation type, thereby preserving sharp geometric boundaries and fine-grained details even under challenging weather conditions, while avoiding the over-smoothing artifacts observed in prior methods.

To further validate the effectiveness of our weather-aware conditioning, we compare feature modulation patterns across scenes with varying levels of synthetic degradation. 

\begin{figure*}
    \centering
    \setlength{\tabcolsep}{0pt}
    \renewcommand{\arraystretch}{1.0}
    \resizebox{\textwidth}{!}{
    \begin{tabular}{@{}m{0.2cm}c@{}}
        & \resizebox{\textwidth}{!}{%
            \begin{tabular}{@{}p{0.25\textwidth}p{0.25\textwidth}p{0.25\textwidth}p{0.25\textwidth}@{}}
                \centering\textbf{Input} & 
                \centering\textbf{Before Cond.} & 
                \centering\textbf{After Cond.} & 
                \centering\arraybackslash\textbf{Difference}
            \end{tabular}
        } \\[2pt]
        
        \smash{\raisebox{0.3cm}{\rotatebox[origin=c]{90}{{Clean}}}} &
        \includegraphics[width=0.97\textwidth]{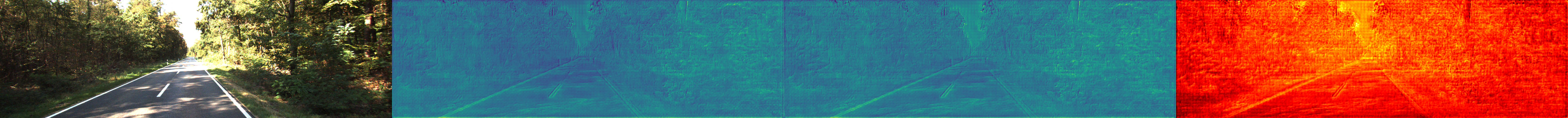} \\[-1pt]
        
        \smash{\raisebox{0.3cm}{\rotatebox[origin=c]{90}{{Fog}}}} &
        \includegraphics[width=0.97\textwidth]{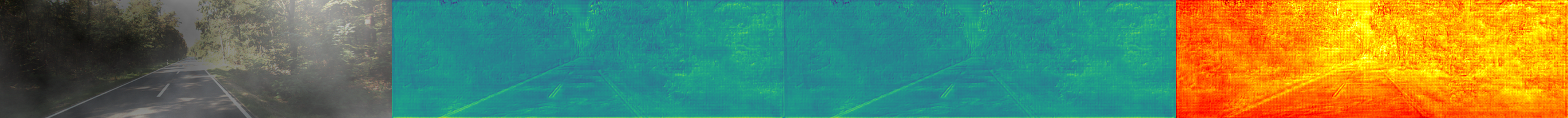} \\[-1pt]
        
        \smash{\raisebox{0.3cm}{\rotatebox[origin=c]{90}{{Snow}}}} &
        \includegraphics[width=0.97\textwidth]{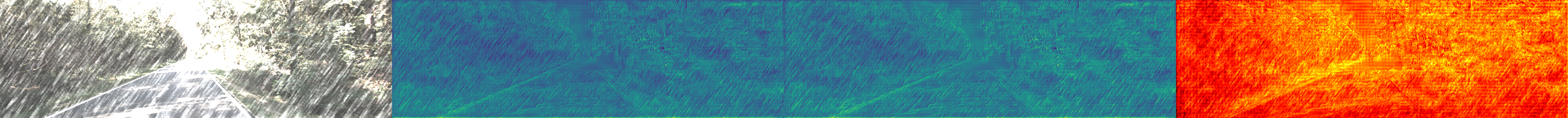} \\
    \end{tabular}
    }
    \vspace{-2mm}
    \caption{\textbf{Visualization of decoder feature modulation by weather-aware conditioning.} 
    The same scene is shown under clean (top), foggy (middle), and snowy (bottom) conditions, where degradations are synthetically applied to enable controlled comparison. 
    From left to right: input image, decoder features before conditioning (zero condition baseline), decoder features after conditioning (with learned weather condition), and absolute difference (hot colormap: darker $\rightarrow$ smaller, brighter $\rightarrow$ larger).}
    \label{fig:supp_feature_map}
\end{figure*}

Figure~\ref{fig:supp_feature_map} shows three representative examples: a clear sunny scene (top), a foggy scene with low visibility (middle), and a snowy scene with heavy snowfall (bottom).
The difference maps reveal two key characteristics of our conditioning mechanism. 
First, at the scene level, severely degraded scenes (snow and fog) exhibit substantially stronger overall modulation compared to the clear scene, demonstrating global adaptiveness to degradation severity.
Second, the modulation exhibits \textit{spatial selectivity} within degraded scenes: in the snowy scene, regions with dense snowflakes (left side of the image) appear significantly brighter yellow-white in the difference map, whereas less-affected regions exhibit relatively lower modulation.
This spatial variation demonstrates that our weather-aware conditioning not only detects weather degradation but also adaptively allocates computational resources to the most severely affected regions.

\section{Optional Design of \textit{Style Filter} Network}

Motivated by the success of CLIP-style vision-language alignment in learning domain-invariant representations, we explored two alternative approaches for learning weather embeddings before settling on our \textit{Style Filter} design.

\clearpage

\clearpage

\subsection{Visualization Results of Decoder Feature Modulation}

\begin{figure}[t]
    \centering
    \begin{minipage}{0.48\linewidth}
        \centering
        \includegraphics[width=\linewidth]{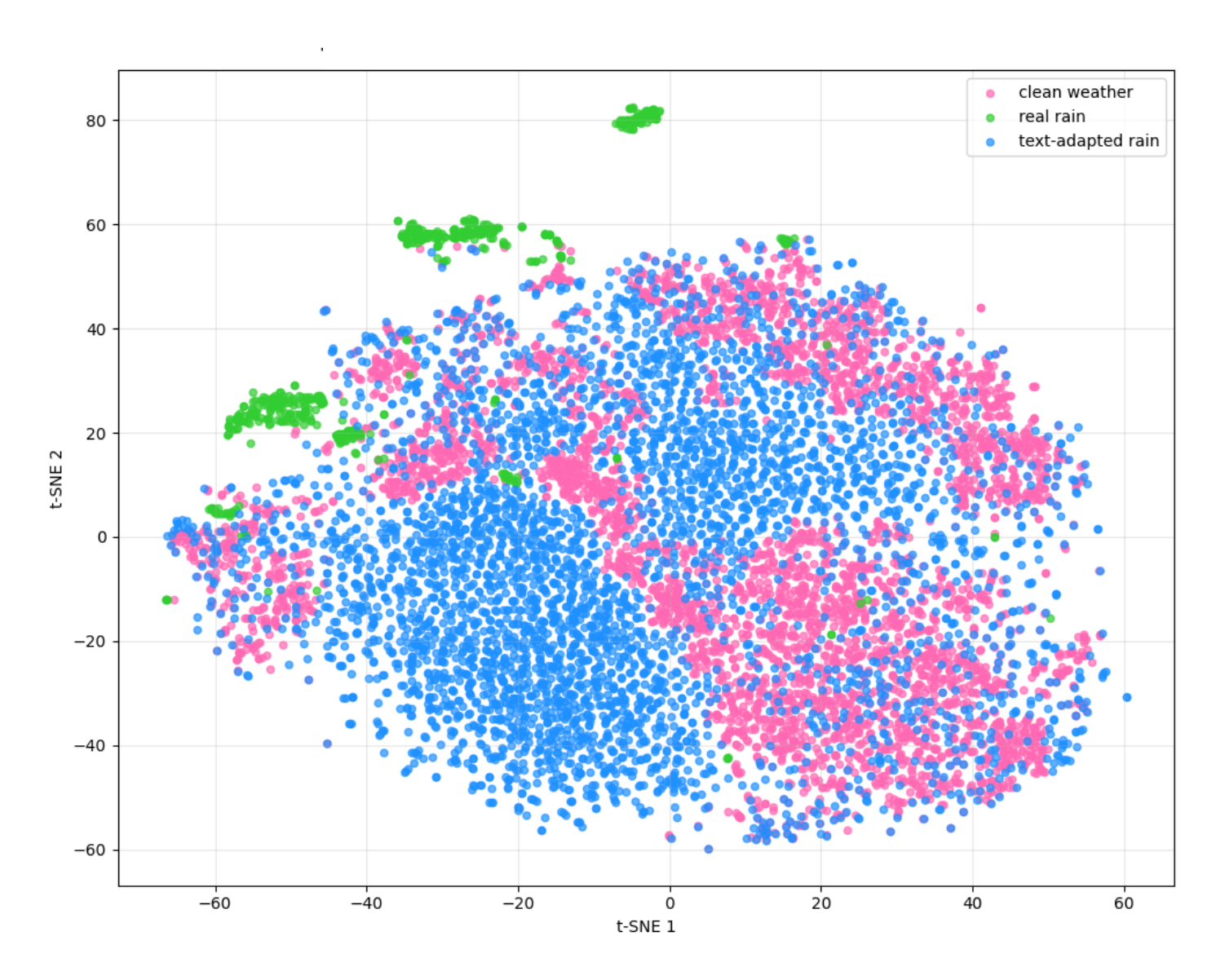}
    \end{minipage}
    \hfill
    \begin{minipage}{0.48\linewidth}
        \centering
        \includegraphics[width=\linewidth]{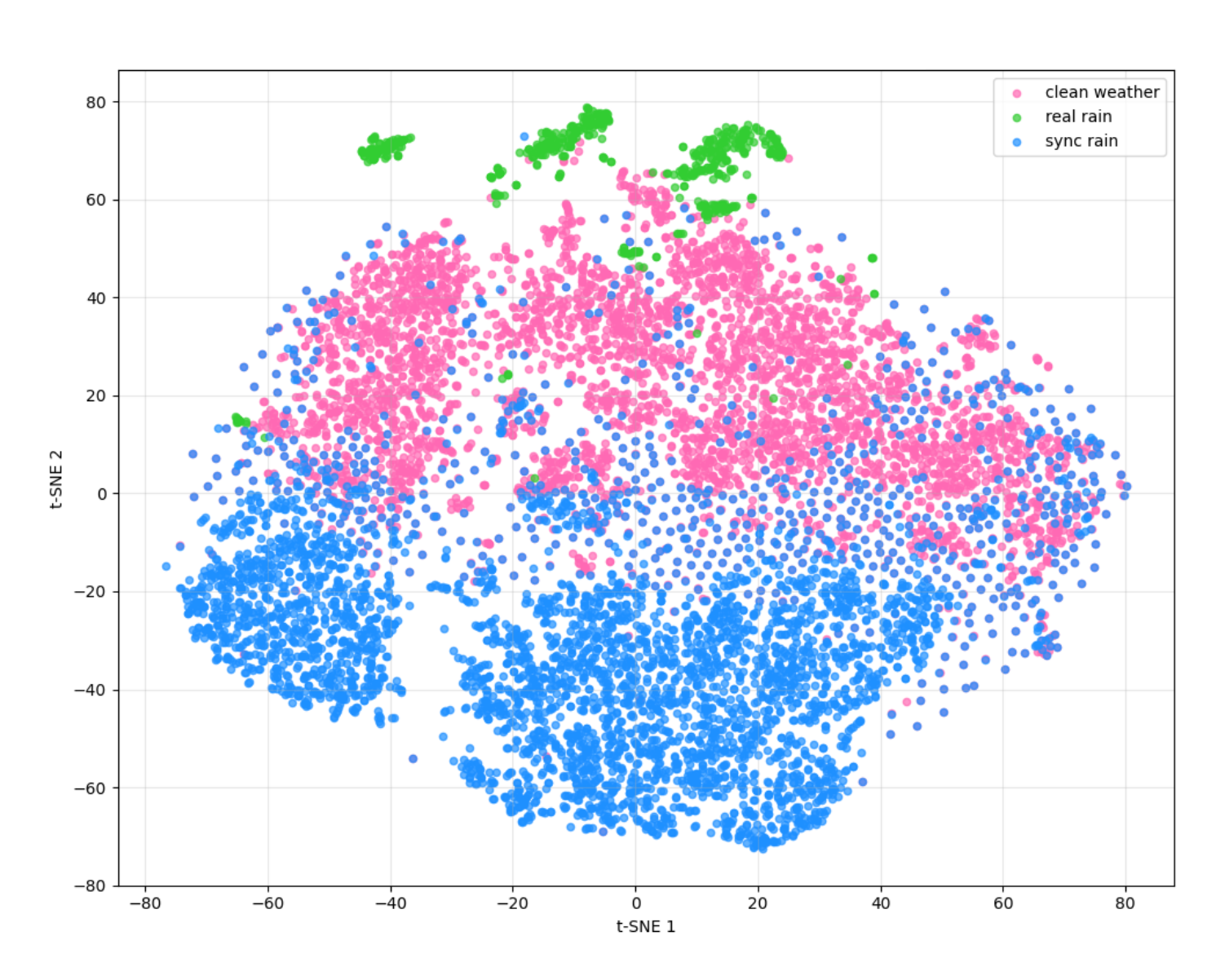}
    \end{minipage}
    
    \caption{\textbf{Feature space visualization for CLIP-style alignment variants.} 
    \textit{Left}: \textbf{Variant (i)} with prompt-driven image-text alignment shows severe overlap between text-adapted rain (blue) and clean weather (pink) embeddings, while real rain (green) forms separate clusters. 
    \textit{Right}: \textbf{Variant (ii)} with direct image-image alignment shows that synthetic rain (blue) and clean weather (pink) form relatively distinct clusters, but real rain (green) occupies entirely different regions in the embedding space.}
    \label{fig:supp_optional}
\end{figure}

\textbf{Variant (i): Prompt-driven Image-Text Alignment (P{\O}DA).~\cite{fahes2022p}} In this variant, we leverage CLIP's image-text alignment capability by constructing handcrafted prompt templates to guide the image encoder toward weather-aware features. Specifically, we use templates such as ``a \{weather\} street scene'' with synonym ensembling for each weather category (\eg, \textit{foggy}/\textit{misty}, \textit{night}/\textit{dark}, \textit{snowy}/\textit{snow-covered}, \textit{rainy}/\textit{wet}). We experiment with extracting Gram matrices from several CLIP visual encoder layers (including the stem, early ViT blocks, and mid-level ViT blocks) and training MLP heads to project these statistics into a compact weather embedding space. The objective encourages images matching the prompted weather descriptions to cluster together while pushing apart images from different weather categories.

\textbf{Variant (ii): Direct Image-Image Alignment.~\cite{fahes2024domain}} To eliminate potential ambiguities introduced by text prompts, we also evaluated a purely visual approach that removes text guidance entirely. This variant aligns CLIP image features directly across domains by treating real-synthetic image pairs of the same weather condition as positive samples. Similar to variant (i), we extract multi-layer Gram matrices from the CLIP visual encoder, vectorize them, and learn weather embeddings via contrastive learning. The key difference is that the supervision signal comes exclusively from image-image similarity rather than image-text correspondence.

\textbf{Negative Results and Analysis.} Despite extensive hyperparameter tuning—including layer selection sweeps (testing features from different CLIP encoder depths), prompt template variations, margin adjustments, and both image-text and image-image alignment strategies—both variants failed to achieve satisfactory domain alignment. 

As visualized in t-SNE projections in Figure~\ref{fig:supp_optional}, \textbf{Variant (i)} shows severe overlap where text-adapted rain embeddings (blue) are heavily intermingled with clean weather samples (pink), indicating that text-guided CLIP features conflate rainy and clean conditions due to their semantic similarity. For \textbf{Variant (ii)}, while synthetic rain (blue) forms a concentrated cluster separated from clean weather (pink), real rain samples (green) occupy entirely different regions, revealing that direct image-image alignment cannot bridge the real-synthetic domain gap.

We conjecture that CLIP features are predominantly optimized for high-level semantic understanding, making them inherently invariant to low-level, physics-driven appearance variations such as atmospheric scattering and noise patterns that distinguish real weather degradations from synthetic counterparts, rendering them insufficient for capturing the distributional shifts critical for robust domain alignment.